\documentclass{article} %
\usepackage[T1]{fontenc}
\usepackage{iftex}
\usepackage{iclr2027_conference,times}
\usepackage{natbib}

\usepackage{amsmath,amsfonts,bm}

\def\eqref#1{equation~\ref{#1}}

\def\1{\bm{1}}

\DeclareMathAlphabet{\mathsfit}{\encodingdefault}{\sfdefault}{m}{sl}
\SetMathAlphabet{\mathsfit}{bold}{\encodingdefault}{\sfdefault}{bx}{n}

\usepackage{hyperref}
\usepackage{url}
\usepackage{algorithm}  
\usepackage{algpseudocode}
\usepackage{xcolor}

\usepackage{url}           
\usepackage{booktabs}       
\usepackage{amsfonts}  
\usepackage{nicefrac}      
\usepackage{microtype} 
\usepackage{xcolor}         
\usepackage{enumitem}
\usepackage{multirow}
\usepackage{mathtools}
\usepackage{amssymb}
\usepackage{colortbl}
\usepackage{graphicx}
\usepackage{pgfplots}  
\usepackage{caption}    

\usepackage{subcaption} 
\usepackage{amsmath}
\usepackage{booktabs}
\usepackage{wrapfig}
\usepackage{float}
\usepackage{enumitem}
\usepackage{hyperref}
\usepackage{tcolorbox}
\usepackage{siunitx}         %
\usepackage{pgfmath}  
\definecolor{RowHighlight}{gray}{0.90}

\usepackage{microtype}
\usepackage{xspace}
\usepackage{xcolor}
\definecolor{MethodInk}{HTML}{7A3410}

\newcommand{\methodmark}{%
  \ifXeTeX A\kern0.07em S\kern0.07em K\else\textls[70]{ASK}\fi}
\newcommand{\method}{%
  {\fontfamily{bch}\selectfont\bfseries
   \textcolor{MethodInk}{\methodmark}}\xspace}
\newcommand{\methodlong}{Alignment from Solicited Kernels\xspace}

\title{Kernel-Based Steering of CLIP with Vision-Language Model Preferences}

\author{\begin{minipage}{\dimexpr\textwidth-2\tabcolsep\relax}
\centering
Sajjad Ghiasvand\textsuperscript{1}\quad
Haniyeh Ehsani Oskouie\textsuperscript{2}\quad
Sina Mansouri\textsuperscript{3}\\[3pt]
Mahnoosh Alizadeh\textsuperscript{1}\quad
Farzan Farnia\textsuperscript{4}\quad
Ramtin Pedarsani\textsuperscript{1}\\[7pt]
{\normalfont\small
\textsuperscript{1}University of California, Santa Barbara (\href{https://ucsb.edu}{ucsb.edu})\\
\textsuperscript{2}University of California, Los Angeles (\href{https://ucla.edu}{ucla.edu})\\
\textsuperscript{3}George Mason University (\href{https://gmu.edu}{gmu.edu})\\
\textsuperscript{4}The Chinese University of Hong Kong (\href{https://cuhk.edu.hk}{cuhk.edu.hk})}\\[6pt]
{\normalfont\footnotesize
\href{mailto:sajjad@ucsb.edu}{\texttt{sajjad@ucsb.edu}}\quad
\href{mailto:haniyeh@cs.ucla.edu}{\texttt{haniyeh@cs.ucla.edu}}\quad
\href{mailto:smansou3@gmu.edu}{\texttt{smansou3@gmu.edu}}\\[2pt]
\href{mailto:alizadeh@ucsb.edu}{\texttt{alizadeh@ucsb.edu}}\quad
\href{mailto:farnia@cse.cuhk.edu.hk}{\texttt{farnia@cse.cuhk.edu.hk}}\quad
\href{mailto:ramtin@ucsb.edu}{\texttt{ramtin@ucsb.edu}}}
\end{minipage}}

\iclrfinalcopy
\hypersetup{
  pdftitle={Kernel-Based Steering of CLIP with Vision-Language Model Preferences},
  pdfauthor={Sajjad Ghiasvand; Haniyeh Ehsani Oskouie; Sina Mansouri; Mahnoosh Alizadeh; Farzan Farnia; Ramtin Pedarsani},
  pdfsubject={Preprint}
}
\begin{document}
\addtocontents{toc}{\protect\setcounter{tocdepth}{-1}}

\maketitle

\begin{abstract}
Large vision-language models (VLMs) can judge visual similarity, but their judgments are not directly available as compact image embeddings for efficient comparison. We study how to transfer these preferences into CLIP while retaining its image--text capabilities. We introduce \method{}, a kernel-based steering method that learns from elicited pairwise judgments without accessing teacher embeddings or collecting new human similarity annotations. \method{} constructs positive semidefinite target kernels within small image groups and combines visual kernel matching with an image--text distributional anchor. Low-rank adapters jointly update the visual and text encoders while regularizing predictions toward frozen CLIP. After adaptation, retrieval uses CLIP image embeddings and cosine similarity, with no VLM calls. Experiments across five image domains, four CLIP backbones, and six judges evaluate teacher agreement, retrieval, and recognition retention. For ViT-B/16, mean retrieval mAP on classes excluded from adaptation increases from 53.8 to 75.0, compared with 71.7 for DINOv2 targets with KL anchoring. Mean zero-shot accuracy with jointly adapted encoders increases from 61.8\% to 62.4\%, averaged over 12 benchmarks and the five adaptation domains. Prompting provides an additional capability: selecting which visual distinctions the student learns. Human-annotated evaluations across four datasets support this criterion-specific control.

\end{abstract}

\section{Introduction}
\label{sec:intro}
Large vision-language models (VLMs), such as GPT~\citep{openai2026gpt56sol} and Qwen~\citep{qwen2026qwen35}, can judge visual similarity through natural-language interactions \citep{wu2024qalign,xiong2025llavacritic}. Conversational interfaces return responses to image pairs rather than reusable embeddings for cosine comparison. CLIP provides compact embeddings and a shared image--text space \citep{radford2021learning}, but its pretrained similarity need not agree with a VLM's judgments \citep{tong2024eyes,vaze2023genecis}. We ask: \emph{How can a compact CLIP model learn a large VLM's visual similarity preferences while retaining its pretrained image--text capabilities?}

\begin{figure}[t]
    \centering
    \includegraphics[width=0.85\columnwidth,height=0.45\textheight,keepaspectratio]{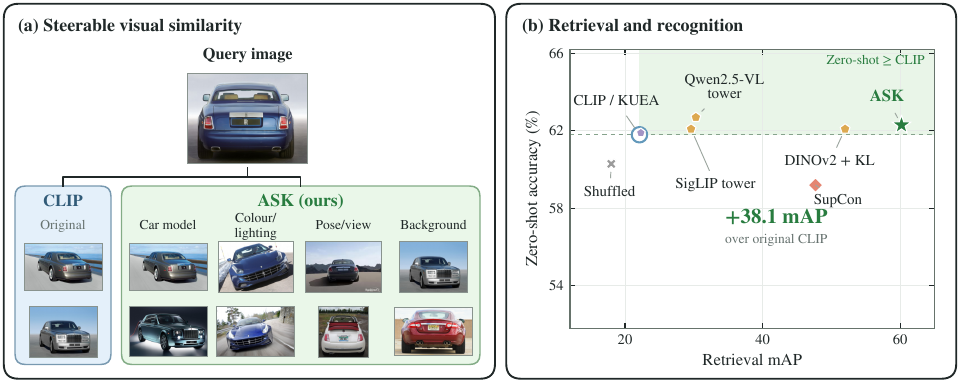}
    \caption{\textbf{Transferring VLM similarity preferences into CLIP while retaining recognition.} \textbf{(a) Criterion-specific retrieval.} For one selected car query, columns show the exact top-two cosine matches from CLIP and \method{}, ranked top to bottom. Separate \method{} encoders emphasize car model; colour and lighting; pose, viewpoint, and framing; or background. Retrieval uses image embeddings alone, with no VLM calls. Additional examples and selection details are in Appendix~\ref{app:visual-examples}. \textbf{(b) Retrieval and recognition together.} Eight CLIP ViT-B/16 methods on Dogs (Table~\ref{tab:backbones}) compare unseen-breed retrieval mAP with mean zero-shot accuracy over 12 benchmarks. The green region shows retrieval gains while matching or exceeding CLIP's recognition accuracy. Feature labels identify teachers, including SigLIP \citep{zhai2023sigmoid}; CLIP and KUEA nearly coincide.}
    \label{fig:intro-teaser}
  \end{figure}

Our starting point is relational transfer, which learns relationships among examples without matching the teacher's feature coordinates \citep{park2019relational,tung2019similarity}. KUEA~\citep{gong2025kernel} aligns CLIP with a vision teacher such as DINOv2~\citep{oquab2024dinov2}, using an L2 anchor on CLIP's original visual features. We instead obtain the target relationships by asking a VLM to compare images. This requires no internal representations, making hosted VLMs usable as teachers.

We introduce \methodlong{} (\method{}), a method that transfers these judgments into CLIP's embedding geometry. \method{} scores image pairs within small groups and converts their scores into positive semidefinite target kernels. Projection ensures that independently elicited scores form a valid kernel. Low-rank adaptation \citep{hu2022lora} then fits the visual encoder to the resulting similarity structure. Because matching image--image relationships alone does not constrain their correspondence with text, we combine kernel alignment with an image--text distributional anchor. The anchor jointly updates adapters in both encoders while regularizing predictions over shared text prompts toward frozen CLIP. After adaptation, images can be encoded independently and gallery embeddings cached for cosine retrieval. A new query therefore requires one CLIP image encoding and comparisons with the stored embeddings, avoiding repeated VLM calls for candidate pairs. The jointly adapted encoders also support zero-shot recognition.

We measure agreement with held-out VLM judgments, retrieval on classes excluded from adaptation, and retention of zero-shot recognition across five domains, four CLIP backbones, and six judges (Tables~\ref{tab:main}--\ref{tab:judges}). Class labels define retrieval relevance independently of the judge. For ViT-B/16, mean retrieval mAP increases from 53.8 to 75.0, compared with 68.3 for supervised contrastive learning and 71.7 for DINOv2+KL. Mean zero-shot accuracy with jointly adapted encoders increases from 61.8\% to 62.4\%, averaged over 12 benchmarks and the five adaptation domains. Figure~\ref{fig:intro-teaser}(b) illustrates the combination of improved retrieval and retained recognition.

Language provides an additional advantage: prompts can specify which distinctions the student should learn. Asking the same teacher to emphasize different properties yields target relationships, which \method{} transfers into separate CLIP encoders. For the same car query in Figure~\ref{fig:intro-teaser}(a), colour-aligned retrieval favours blue cars, whereas viewpoint-aligned retrieval favours rear views. We test criterion specificity on shared images and evaluate attribute, viewpoint, pose, and background retrieval against human annotations across four datasets. Annotations are reserved for evaluation.

Conditional similarity networks and instruction-conditioned embedders support different notions of similarity \citep{veit2017conditional,hsieh2025focallens,zhang2024magiclens,jiang2025vlm2vec}. \method{} incorporates the criterion during adaptation, so deployment uses CLIP embeddings without an instruction or judge call. New criteria require supervision and adaptation (Appendix~\ref{app:related-work}).

Our contributions are: \textbf{(I) Transferring VLM preferences into CLIP.} We learn visual similarity from elicited pairwise judgments through batchwise kernel alignment, without requiring teacher embeddings or new human similarity annotations. \textbf{(II) Joint visual and textual adaptation.} We combine visual kernel matching with an image--text distributional anchor, adapting both encoders to support preference alignment and retention of pretrained recognition capabilities. \textbf{(III) Evaluating alignment, utility, and control.} We assess teacher fidelity, unseen-class retrieval, and zero-shot recognition across domains, backbones, and judges, and demonstrate prompt-selected criterion control through independent human-annotated evaluations.

\section{Method}
\label{sec:method}
\begin{figure*}[t]
    \centering
    \includegraphics[width=1.0\textwidth]{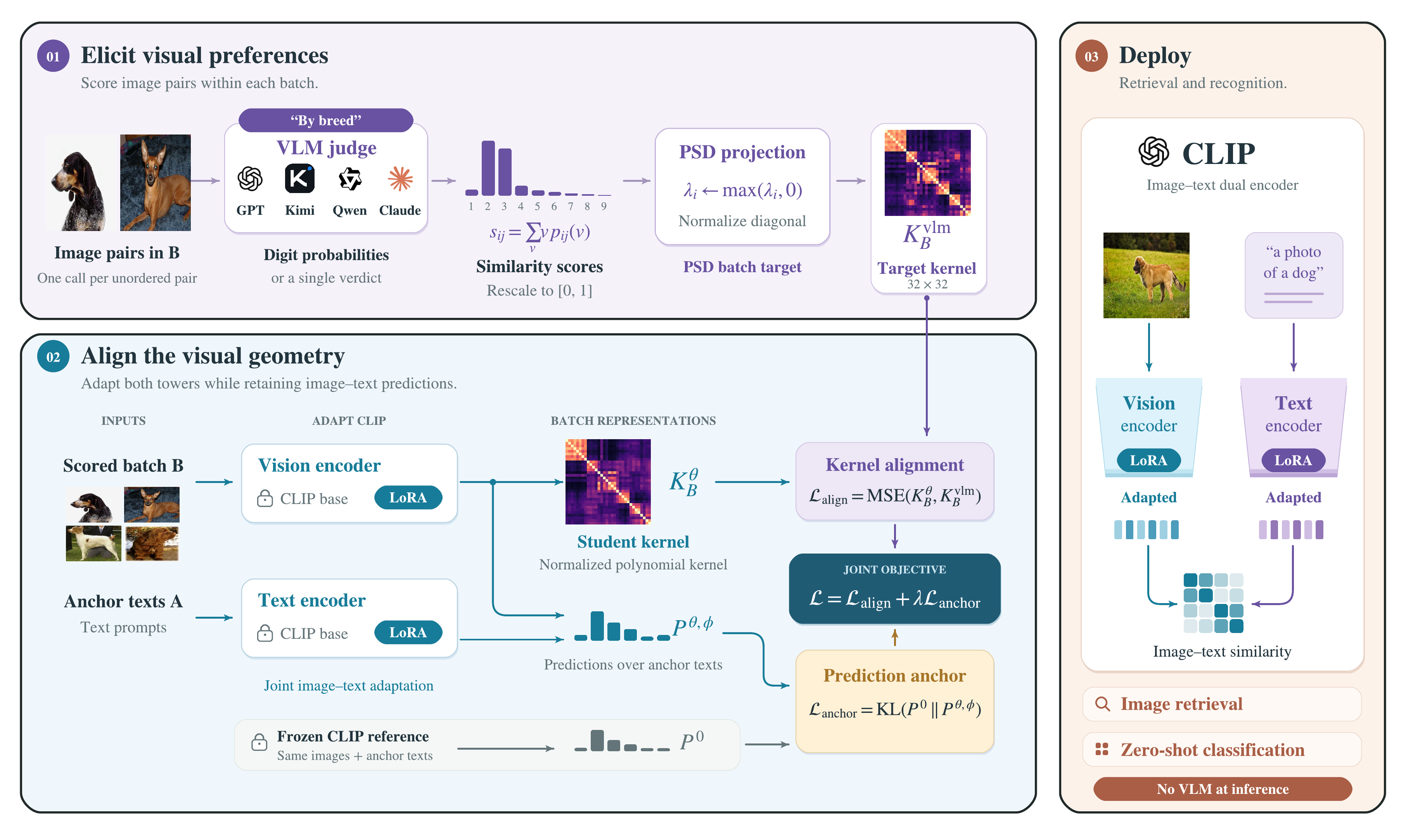}
    \caption{\textbf{\method{} transfers a VLM's similarity preferences into CLIP for retrieval and recognition.} \textbf{(1) Elicit supervision.} A prompted VLM scores image pairs within small groups. Digit probabilities, when available, yield expected scores; symmetrization, positive-semidefinite (PSD) projection, and diagonal normalization produce a target kernel for each group. Neither teacher embeddings nor new human similarity annotations are required. \textbf{(2) Adapt both encoders.} Visual kernel matching teaches the image encoder the requested similarity structure. A KL anchor keeps image--text predictions over shared text prompts close to frozen CLIP's predictions, updating LoRA adapters in both towers while pretrained weights remain fixed. \textbf{(3) Deploy CLIP.} Both adapted encoders are retained: image--image cosine similarity supports retrieval, and image--text similarity supports zero-shot classification. Deployment requires no VLM calls; a new criterion requires new judgments and adaptation. Heatmaps are measured; probability bars are illustrative.}
    \label{fig:workflow}
\end{figure*}

We introduce \method{}, a framework for adapting CLIP's visual similarity to pairwise judgments from a vision-language model. \method{} elicits pairwise scores within image batches, converts these scores into positive semidefinite target kernels, and aligns the visual encoder to their similarity structure. An image--text distributional regularizer supports retention of zero-shot classification performance. Figure~\ref{fig:workflow} summarizes the framework, and Algorithm~\ref{alg:ask} gives the training procedure.

\begin{algorithm}[t]
\caption{\method{}: \methodlong}
\label{alg:ask}
\small
\algrenewcommand\algorithmicindent{1em}
\algrenewcommand\alglinenumber[1]{\scriptsize\textcolor{black!45}{#1}}
\newcommand{\askstage}[1]{%
  \Statex\colorbox{black!6}{%
    \makebox[\dimexpr\linewidth-2\fboxsep\relax][l]{\strut\textbf{#1}}}}
\begin{algorithmic}[1]
\Statex \textbf{Data:} Image groups $\{\mathcal{G}_g\}_{g=1}^{G}$ of size $b$; comparison prompt $q$; anchor texts $\mathcal{A}$.
\Statex \textbf{Models:} VLM judge $\mathcal{J}$; frozen CLIP encoders $(f_{\theta_0},g_{\phi_0})$.
\Statex \textbf{Settings:} Score scale $V$; kernel $(\gamma,c,p)$; LoRA rank $r$; steps $T$; anchor weight $\lambda$; logit scale $\tau$.
\askstage{1. Construct teacher targets}
\State Query $\mathcal{J}$ under $q$ once per distinct unordered within-group pair; store scores $s_{ij}$.
\For{each group $\mathcal{G}_g$}
  \State Form symmetric $\mathbf{S}_g$ with unit diagonal and off-diagonal entries $(s_{ij}-1)/(V-1)$.
  \State $\mathbf{A}_g\gets\Pi_+(\mathbf{S}_g)$; \quad $D_g\gets\operatorname{diag}\!\bigl((\mathbf{A}_g)_{11},\dots,(\mathbf{A}_g)_{bb}\bigr)$
  \State Store $\mathbf{K}^{\mathrm{vlm}}_{\mathcal{G}_g}\gets D_g^{-1/2}\mathbf{A}_gD_g^{-1/2}$.
\EndFor
\askstage{2. Adapt CLIP}
\State Initialize $(f_\theta,g_\phi)$ from $(f_{\theta_0},g_{\phi_0})$ with rank-$r$ LoRA in both towers.
\State Keep pretrained weights and kernel parameters fixed; shuffle the group order.
\For{$t = 1,\dots,T$}
  \State Take the next group as $B$; reshuffle after each pass through the groups.
  \State $\mathbf{K}^{\theta}_B\gets[\tilde{k}_{\gamma,c}(f_\theta(I_i),f_\theta(I_j))]_{i,j=1}^{b}$
  \State Compute $P^0_i$ and $P^{\theta,\phi}_i$ over $\mathcal{A}$ for each $I_i\in B$ using Equation~\ref{eq:anchordist}.
  \State $\mathcal{L}\gets\mathcal{L}_{\mathrm{align}}(\theta;B)+\lambda\,\mathcal{L}_{\mathrm{anchor}}(\theta,\phi;B)$ \hfill (Equation~\ref{eq:full})
  \State Update only the LoRA parameters in $(\theta,\phi)$ using gradients of $\mathcal{L}$.
\EndFor
\State \Return $f_\theta$ and $g_\phi$
\end{algorithmic}
\end{algorithm}

\subsection{Preliminaries: kernels over embeddings}
\label{sec:prelim}

Let $f_\theta$ denote CLIP's visual encoder \citep{radford2021learning}, mapping an image $I$ to an embedding $x=f_\theta(I)\in\mathbb{R}^d$. A positive semidefinite (PSD) kernel $k$ defines a symmetric matrix $\mathbf{K}=[k(x_i,x_j)]_{ij}\succeq 0$ for any finite collection of embeddings. This matrix represents their pairwise similarity structure. We use the polynomial kernel
\begin{equation}
\label{eq:polykernel}
k_{\gamma,c}(x,y)=(\gamma\,x^\top y+c)^p,
\qquad \gamma>0,\quad c\geq 0,
\end{equation}
where $\gamma$ scales the inner product, $c$ is an additive offset, and $p\geq 1$ is an integer degree. We normalize the kernel to unit diagonal: $\tilde{k}_{\gamma,c}(x,y)=k_{\gamma,c}(x,y)/\sqrt{k_{\gamma,c}(x,x)\,k_{\gamma,c}(y,y)}$.
Here, $x$ and $y$ are unnormalized encoder outputs; the normalization acts on the kernel values. Following kernel alignment \citep{gong2025kernel}, we match pairwise similarity structure without requiring correspondence between teacher and student feature coordinates. We separately regularize image--text predictions to preserve alignment between visual and textual representations (Section~\ref{sec:anchor}).

\subsection{Eliciting a preference kernel from a VLM}
\label{sec:elicit}
\label{sec:groups}

The teacher in our framework is not an embedding model but a conversation. Given a pair of images $(I_i, I_j)$ and a natural-language comparison prompt $q$ (for instance, ``judge similarity by breed, build, coat, and markings''), we present both images to an instruction-tuned VLM and ask it to answer with a single digit on a $V$-point scale.

\paragraph{Expected-digit scoring.}
Selecting the most probable digit discards the probability assigned to alternative ratings. When token probabilities are available, we instead summarize the model's next-token distribution over the rating digits by its expectation,
\begin{equation}
\label{eq:expectation}
s_{ij} \;=\; \sum_{v=1}^{V} v \cdot p_{ij}(v),
\qquad
p_{ij}(v) = \frac{\exp z_{ij}(v)}{\sum_{u=1}^{V} \exp z_{ij}(u)},
\end{equation}
where $z_{ij}(v)$ is the logit of digit token $v$. If the API returns only the top-$k$ token log-probabilities, we retain the returned rating digits and divide their probabilities by their sum before computing the expectation. No additional teacher query is needed. Each unordered image pair is scored once. We map its score to $[0,1]$ and assign it to both corresponding off-diagonal entries. Setting the diagonal to one yields the symmetric elicited matrix $\mathbf{S}$.

\textbf{Prompt-dependent similarity.}
The comparison prompt $q$ can name a category, such as dogs or cars, to elicit within-category similarity, or specify an attribute such as colour or pose. This covers both domain-focused alignment and attribute-focused steering. For $C$ prompts, we construct $C$ sets of groupwise target kernels, each used to train a separate encoder. Holding the images and groups fixed isolates how the requested comparison changes the learned geometry.

\paragraph{Batchwise scoring.}
The alignment loss requires only within-group similarities. For each group of $b$ images, we elicit at most $\binom{b}{2}$ judgments and construct a $b\times b$ target, avoiding a full dataset-wide similarity matrix. Our experiments cache scores for $G$ predefined groups and reuse them across training steps without additional teacher queries (Algorithm~\ref{alg:ask}).

\subsection{From elicited similarities to PSD kernels}
\label{sec:psd}

For a batch $B$, the elicited matrix $\mathbf{S}_B$ is symmetric with unit diagonal, but its pairwise judgments need not be consistent with a PSD kernel. This differs from a Gram matrix constructed from teacher embeddings, which is PSD by construction. The student's polynomial kernel is also PSD under the constraints in Equation~\ref{eq:polykernel}, and diagonal normalization preserves this property. An indefinite target therefore cannot be matched exactly by the student.

We construct a PSD target for each batch by clipping negative eigenvalues. Let $\mathbf{S}_B=U_B\Lambda_B U_B^\top$, with eigenvalues $\lambda_1,\ldots,\lambda_b$. Its nearest PSD matrix in Frobenius norm is \citep{higham1988computing}
\begin{equation}
\label{eq:psdproj}
\mathbf{A}_B=\Pi_+(\mathbf{S}_B)
=U_B\operatorname{diag}\!\left(\max\{\lambda_1,0\},\ldots,\max\{\lambda_b,0\}\right)U_B^\top.
\end{equation}
We then restore unit self-similarity by diagonal normalization. Writing $D_B$ for the diagonal matrix with $(D_B)_{ii}=(\mathbf{A}_B)_{ii}$, the training target is
\begin{equation}
\label{eq:targetnormalize}
\mathbf{K}^{\mathrm{vlm}}_B=D_B^{-1/2}\mathbf{A}_B D_B^{-1/2}.
\end{equation}
This normalization preserves PSD and gives unit diagonal. The nearest-PSD property applies to $\mathbf{A}_B$ before normalization; the final target is not necessarily the nearest unit-diagonal PSD matrix. Both operations are applied independently within each scored group.

\subsection{Alignment objective}
\label{sec:objective}

For a batch $B = \{I_1,\dots,I_b\}$, let $\mathbf{K}^{\theta}_B = [\tilde{k}_{\gamma,c}(f_\theta(I_i), f_\theta(I_j))]_{ij}$ be the normalized kernel matrix of the adapted visual encoder. We align it with the PSD teacher target $\mathbf{K}^{\mathrm{vlm}}_B$ by minimizing the mean squared difference between their entries:
\begin{equation}
\label{eq:align}
\mathcal{L}_{\mathrm{align}}(\theta;B)
\;=\;
\frac{1}{b^2}\,\bigl\Vert \mathbf{K}^{\theta}_B - \mathbf{K}^{\mathrm{vlm}}_B \bigr\Vert_F^2 .
\end{equation}
In our experiments, the alignment objective averages this loss over the scored groups (Section~\ref{sec:elicit}). We train rank-$r$ LoRA adapters \citep{hu2022lora} in the attention projections of both the visual and text encoders, keeping all pretrained weights frozen. The alignment loss updates the visual adapters, while the image--text regularizer in Section~\ref{sec:anchor} updates both visual and text adapters.

\subsection{Distributional anchoring}
\label{sec:anchor}

Matching image--image similarities alone does not constrain the visual embeddings' relationship to CLIP's text embeddings. To support retention of zero-shot predictions, we regularize image--text distributions against the frozen model. Empirically, \method{} can also improve zero-shot accuracy over untrained CLIP (Section~\ref{sec:main-results}). Let $g_\phi$ denote the adapted text encoder and $\mathcal{A} = \{T_1,\dots,T_M\}$ a set of anchor texts. For image $I_i$, the adapted model defines a distribution over these texts:
\begin{equation}
\label{eq:anchordist}
P^{\theta,\phi}_i
=\mathrm{softmax}\!\left(\tau\,\bigl[\hat{f}_\theta(I_i)^\top\hat{g}_\phi(T_m)\bigr]_{m=1}^{M}\right),
\end{equation}
where hats denote $\ell_2$ normalization and $\tau>0$ is a fixed logit scale. Let $P^0_i$ be the same distribution computed using the frozen encoders $f_{\theta_0}$ and $g_{\phi_0}$. For a batch $B$ of $b$ images, the anchor loss is
\begin{equation}
\label{eq:kl}
\mathcal{L}_{\mathrm{anchor}}(\theta,\phi;B)
=\frac{1}{b}\sum_{i=1}^{b}\mathrm{KL}\!\left(P^0_i\,\Vert\,P^{\theta,\phi}_i\right),
\end{equation}
and the full objective is
\begin{equation}
\label{eq:full}
\mathcal{L}(\theta,\phi;B)
=\mathcal{L}_{\mathrm{align}}(\theta;B)
+\lambda\,\mathcal{L}_{\mathrm{anchor}}(\theta,\phi;B).
\end{equation}

Here, $\lambda\geq 0$ controls the strength of the anchor. The KL term couples the two encoders through their predictions on $\mathcal{A}$, with gradients flowing through both adapted towers. At inference, zero-shot classification compares normalized image and prompt embeddings from the jointly adapted encoders $f_\theta$ and $g_\phi$, while image retrieval uses cosine similarity between visual embeddings. The anchor texts and training settings are specified in Section~\ref{sec:setup}; comparisons of regularizers and KL directions are given in Section~\ref{sec:regablation}.

\section{Experimental Results}

\subsection{Setup}
\label{sec:setup}

\textbf{Data.}
We evaluate five fine-grained domains: ImageNet dog breeds \citep{deng2009imagenet}, Oxford-IIIT Pets \citep{parkhi2012cats}, Describable Textures \citep{cimpoi2014describing}, Stanford Cars \citep{krause2013collecting}, and Oxford Flowers \citep{nilsback2008automated}. Each domain contains 25 adaptation classes with 40 training and 10 held-out fidelity-test images per class. We elicit domain-specific similarities and additionally score dogs and cars under multiple comparison criteria. For \method{}, SupCon, and DINOv2+KL, we select hyperparameters by retrieval mAP on a separate Dogs validation set and reuse the selected settings across domains. Dataset splits, scoring budgets, and evaluation protocols appear in Appendix~\ref{app:experimental-setup}. We also evaluate attribute, viewpoint, pose, and background retrieval using human annotations from four datasets: CUB-200-2011~\citep{wah2011cub}, PASCAL3D+~\citep{xiang2014beyond}, COCO-Stuff~\citep{caesar2018cocostuff}, and AP-10K~\citep{yu2021ap10k} (Appendix~\ref{app:human-attributes}).

\textbf{Baselines.}
We compare untrained CLIP \citep{radford2021learning} with label supervision (SupCon; \citealp{khosla2020supervised}), class-label kernels (Appendix~\ref{app:label-kernels}), and feature-based kernel targets from frozen encoders, including DINOv2 \citep{oquab2024dinov2}, SigLIP \citep{zhai2023sigmoid}, and the Qwen2.5-VL vision tower \citep{bai2025qwen25vl,bai2025qwen25report}. Feature-target comparisons retain our KL anchor; KUEA \citep{gong2025kernel} uses its L2 feature anchor. We also evaluate shuffled VLM scores and linear or MLP heads trained on frozen CLIP features. Adaptation comparisons use the same training images, with changes to supervision, anchoring, or adaptation stated in each table. SupCon combines a supervised contrastive image loss with KL anchoring (Appendix~\ref{app:supcon-repair}).

\textbf{Training.}
Unless stated otherwise, we align CLIP ViT-B/16 to Qwen3.5-397B-A17B judgments \citep{qwen2026qwen35} using rank-32 LoRA in both towers for 3{,}000 steps, keeping pretrained weights frozen. The default uses expected-digit scores, PSD-projected targets, a normalized cubic kernel, and a KL anchor over ImageNet class-name prompts with weight selected on the Dogs validation set. At evaluation, we retain both the adapted image and text encoders. We also compare GPT-5.6 Sol \citep{openai2026gpt56sol}, Kimi-K3 \citep{kimi2026k3}, Claude Sonnet 5 \citep{anthropic2026sonnet5}, Qwen3.8-27B \citep{qwen2026qwen38}, and Gemma 4 12B \citep{gemma2026gemma4,gemmateam2026gemma4} as judges under a matched 31-group budget. Full settings are in Appendix~\ref{app:experimental-setup}.

\textbf{Metrics.}
\emph{Fidelity} (Fid.) is Spearman correlation between the student's normalized kernel similarities and raw VLM scores on held-out image pairs from the adaptation classes. \emph{Retrieval} (mAP) ranks images by cosine similarity on classes excluded from adaptation, using class labels to define relevance independently of the judge. CUB attribute retrieval uses nDCG@10 ($\times100$), with human annotations defining graded relevance. \emph{Zero-shot accuracy} (ZS) averages top-1 accuracy over 12 benchmarks using the jointly adapted encoders, measuring retention of CLIP's classification ability. Unless stated otherwise, results are averaged over three random seeds.

\subsection{Main results}
\label{sec:main-results}

\begin{table}[t]
\caption{\textbf{Supervision across five adaptation domains.} CLIP ViT-B/16 with label, feature, or judgment supervision. KUEA uses an L2 feature anchor; other adapted methods use KL. Mean averages the five domains. Blue/red indicates improvement/decline relative to untrained CLIP; intensity is scaled separately per metric. Bold marks the best score.}
\label{tab:main}
\centering
\begingroup
\footnotesize
\setlength{\tabcolsep}{1.6pt}
\setlength{\arrayrulewidth}{0.4pt}
\arrayrulecolor{black}
\renewcommand{\arraystretch}{1.16}
\resizebox{0.9\linewidth}{!}{%
\begin{tabular}{|l|ccc|ccc|ccc|ccc|ccc|ccc|}
\hline
 & \multicolumn{3}{c|}{Untrained CLIP} & \multicolumn{3}{c|}{SupCon} & \multicolumn{3}{c|}{KUEA} & \multicolumn{3}{c|}{Qwen2.5-VL tower} & \multicolumn{3}{c|}{DINOv2 $+$ KL} & \multicolumn{3}{c|}{\textbf{\method{} (ours)}} \\
\cline{2-19}
Domain & Fid. & mAP & ZS & Fid. & mAP & ZS & Fid. & mAP & ZS & Fid. & mAP & ZS & Fid. & mAP & ZS & Fid. & mAP & ZS \\
\hline
Dogs & 0.281 & 22.1 & 61.8 & \cellcolor{blue!12}0.466 & \cellcolor{blue!18}47.7 & \cellcolor{red!21}59.2 & \cellcolor{blue!5}0.286 & \cellcolor{blue!5}22.3 & \cellcolor{blue!6}61.9 & \cellcolor{blue!10}0.397 & \cellcolor{blue!9}30.3 & \cellcolor{blue!11}\textbf{62.7} & \cellcolor{blue!15}0.539 & \cellcolor{blue!21}52.0 & \cellcolor{blue!7}62.1 & \cellcolor{blue!25}\textbf{0.794} & \cellcolor{blue!25}\textbf{60.2} & \cellcolor{blue!8}62.3 \\
Textures & 0.371 & 44.1 & 61.8 & \cellcolor{red!5}0.366 & \cellcolor{red!6}42.6 & \cellcolor{red!25}58.6 & \cellcolor{blue!5}0.373 & \cellcolor{blue!5}44.5 & \cellcolor{blue!6}61.9 & \cellcolor{blue!8}0.447 & \cellcolor{blue!7}48.7 & \cellcolor{blue!8}62.2 & \cellcolor{red!7}0.323 & \cellcolor{blue!8}49.1 & \cellcolor{blue!7}62.1 & \cellcolor{blue!18}\textbf{0.706} & \cellcolor{blue!10}\textbf{54.5} & \cellcolor{blue!12}\textbf{62.9} \\
Flowers & 0.679 & 77.4 & 61.8 & \cellcolor{red!14}0.440 & \cellcolor{blue!8}83.3 & \cellcolor{red!12}60.7 & \cellcolor{blue!5}0.681 & \cellcolor{blue!5}77.8 & 61.8 & \cellcolor{blue!5}0.686 & \cellcolor{blue!8}83.5 & \cellcolor{red!6}61.7 & \cellcolor{red!13}0.486 & \cellcolor{blue!10}\textbf{86.8} & \cellcolor{red!6}61.6 & \cellcolor{blue!10}\textbf{0.798} & \cellcolor{blue!9}85.8 & \cellcolor{blue!7}\textbf{62.1} \\
Cars & 0.616 & 67.2 & 61.8 & \cellcolor{red!12}0.434 & \cellcolor{blue!12}81.1 & \cellcolor{red!8}61.3 & \cellcolor{blue!5}0.618 & \cellcolor{blue!5}67.4 & \cellcolor{blue!6}61.9 & \cellcolor{blue!6}0.651 & \cellcolor{blue!6}68.3 & \cellcolor{red!6}61.7 & \cellcolor{red!6}0.581 & \cellcolor{blue!14}83.9 & \cellcolor{blue!6}61.9 & \cellcolor{blue!14}\textbf{0.859} & \cellcolor{blue!14}\textbf{85.1} & \cellcolor{blue!7}\textbf{62.1} \\
Pets & 0.582 & 58.2 & 61.8 & \cellcolor{red!9}0.471 & \cellcolor{blue!20}87.0 & \cellcolor{red!12}60.7 & \cellcolor{blue!5}0.589 & \cellcolor{blue!5}58.7 & \cellcolor{blue!6}61.9 & \cellcolor{red!6}0.556 & \cellcolor{blue!10}67.0 & \cellcolor{blue!8}62.2 & \cellcolor{red!6}0.548 & \cellcolor{blue!20}86.9 & \cellcolor{blue!8}62.2 & \cellcolor{blue!18}\textbf{0.917} & \cellcolor{blue!21}\textbf{89.3} & \cellcolor{blue!10}\textbf{62.6} \\
\hline
Mean & 0.506 & 53.8 & 61.8 & \cellcolor{red!8}0.435 & \cellcolor{blue!13}68.3 & \cellcolor{red!16}60.1 & \cellcolor{blue!5}0.509 & \cellcolor{blue!5}54.1 & \cellcolor{blue!6}61.9 & \cellcolor{blue!7}0.548 & \cellcolor{blue!8}59.5 & \cellcolor{blue!7}62.1 & \cellcolor{red!5}0.495 & \cellcolor{blue!14}71.7 & \cellcolor{blue!6}62.0 & \cellcolor{blue!17}\textbf{0.815} & \cellcolor{blue!16}\textbf{75.0} & \cellcolor{blue!9}\textbf{62.4} \\
\hline
\end{tabular}
}
\endgroup
\end{table}

\textbf{Judgments versus labels and features.}
Table~\ref{tab:main} compares supervision sources using CLIP ViT-B/16 and the same training images. SupCon, Qwen2.5-VL tower, DINOv2+KL, and \method{} use the same LoRA configuration and KL anchoring objective; KUEA uses DINOv2 targets with its L2 feature anchor. Feature-based targets use the Qwen2.5-VL-7B vision tower or DINOv2, while \method{} uses elicited judgments. KUEA remains close to untrained CLIP on fidelity and retrieval. \method{} achieves the highest fidelity in all five domains and the highest retrieval mAP in four, with DINOv2+KL leading on Flowers. Across domains, mean mAP reaches 75.0, compared with 71.7 for DINOv2+KL and 68.3 for SupCon. Mean zero-shot accuracy is 62.4 for \method{}, 60.1 for SupCon, and 61.8 for untrained CLIP. \method{} achieves higher average retrieval mAP than the evaluated feature-target baselines, while retaining zero-shot performance.

\begin{table}[t]
\caption{\textbf{Supervision across four CLIP backbones.} Models are adapted on Dogs. Blue/red indicates improvement/decline relative to each backbone's untrained CLIP; intensity is scaled separately per metric. Bold marks the best score per column.}
\label{tab:backbones}
\centering
\begingroup
\footnotesize
\setlength{\tabcolsep}{1.6pt}
\setlength{\arrayrulewidth}{0.4pt}
\arrayrulecolor{black}
\renewcommand{\arraystretch}{1.16}
\resizebox{0.8\linewidth}{!}{%
\begin{tabular}{|l|ccc|ccc|ccc|ccc|}
\hline
& \multicolumn{3}{c|}{ViT-B/32} & \multicolumn{3}{c|}{ViT-B/16} & \multicolumn{3}{c|}{ViT-L/14} & \multicolumn{3}{c|}{ViT-L/14-336} \\
\cline{2-13}
Method & Fid. & mAP & ZS & Fid. & mAP & ZS & Fid. & mAP & ZS & Fid. & mAP & ZS \\
\hline
Untrained CLIP & 0.288 & 18.7 & \textbf{60.6} & 0.281 & 22.1 & 61.8 & 0.283 & 30.5 & 66.1 & 0.264 & 30.0 & 66.9 \\
Shuffled teacher & \cellcolor{red!7}0.221 & \cellcolor{red!6}16.5 & \cellcolor{red!16}58.2 & \cellcolor{red!8}0.203 & \cellcolor{red!7}18.0 & \cellcolor{red!12}60.3 & \cellcolor{red!6}0.253 & \cellcolor{red!8}24.3 & \cellcolor{red!9}65.2 & \cellcolor{red!7}0.206 & \cellcolor{red!8}22.9 & \cellcolor{red!11}65.5 \\
SupCon~\citep{khosla2020supervised} & \cellcolor{blue!13}0.513 & \cellcolor{blue!15}39.3 & \cellcolor{red!25}56.2 & \cellcolor{blue!12}0.466 & \cellcolor{blue!17}47.7 & \cellcolor{red!17}59.2 & \cellcolor{blue!13}0.503 & \cellcolor{blue!22}65.6 & \cellcolor{blue!6}66.4 & \cellcolor{blue!11}0.435 & \cellcolor{blue!21}63.4 & \cellcolor{red!11}65.5 \\
KUEA~\citep{gong2025kernel} & \cellcolor{blue!5}0.291 & \cellcolor{blue!5}18.8 & \textbf{60.6} & \cellcolor{blue!5}0.286 & \cellcolor{blue!5}22.3 & \cellcolor{blue!5}61.9 & \cellcolor{blue!5}0.286 & \cellcolor{blue!5}30.9 & 66.1 & \cellcolor{blue!5}0.269 & \cellcolor{blue!5}30.4 & \cellcolor{blue!5}67.0 \\
DINOv2 $+$ KL & \cellcolor{blue!14}0.533 & \cellcolor{blue!15}40.7 & \cellcolor{red!11}59.3 & \cellcolor{blue!14}0.539 & \cellcolor{blue!19}52.0 & \cellcolor{blue!6}62.1 & \cellcolor{blue!14}0.521 & \cellcolor{blue!21}64.1 & \cellcolor{blue!9}67.0 & \cellcolor{blue!14}0.518 & \cellcolor{blue!22}65.8 & \cellcolor{blue!9}67.8 \\
SigLIP tower & \cellcolor{blue!8}0.378 & \cellcolor{blue!7}23.6 & \cellcolor{red!5}60.5 & \cellcolor{blue!9}0.377 & \cellcolor{blue!9}29.6 & \cellcolor{blue!6}62.1 & \cellcolor{blue!9}0.380 & \cellcolor{blue!9}38.1 & \cellcolor{blue!8}66.8 & \cellcolor{blue!9}0.370 & \cellcolor{blue!9}38.8 & \cellcolor{blue!7}67.4 \\
Qwen2.5-VL tower & \cellcolor{blue!9}0.397 & \cellcolor{blue!8}24.9 & \cellcolor{red!6}60.4 & \cellcolor{blue!9}0.397 & \cellcolor{blue!9}30.3 & \cellcolor{blue!9}\textbf{62.7} & \cellcolor{blue!9}0.398 & \cellcolor{blue!9}39.1 & \cellcolor{blue!8}66.8 & \cellcolor{blue!10}0.390 & \cellcolor{blue!10}40.4 & \cellcolor{blue!10}\textbf{68.0} \\
\method{} (ours) & \cellcolor{blue!23}\textbf{0.779} & \cellcolor{blue!20}\textbf{51.2} & \cellcolor{red!11}59.2 & \cellcolor{blue!24}\textbf{0.794} & \cellcolor{blue!23}\textbf{60.2} & \cellcolor{blue!7}62.3 & \cellcolor{blue!24}\textbf{0.805} & \cellcolor{blue!24}\textbf{70.9} & \cellcolor{blue!10}\textbf{67.2} & \cellcolor{blue!25}\textbf{0.810} & \cellcolor{blue!25}\textbf{72.3} & \cellcolor{blue!8}67.6 \\
\hline
\end{tabular}
}
\endgroup
\end{table}

\textbf{Consistency across CLIP backbones.}
Table~\ref{tab:backbones} repeats the comparison on four CLIP vision-transformer backbones \citep{radford2021learning,dosovitskiy2021image} using the same dog training set. \method{} leads fidelity and retrieval at every scale, exceeding DINOv2+KL by 6.5--10.5 mAP points. On ViT-L/14, SupCon reaches 65.6 mAP, exceeding DINOv2+KL at 64.1. Zero-shot accuracy improves over untrained CLIP on three backbones, with a 1.4-point decrease on ViT-B/32. Shuffling the judgments reduces all three metrics below untrained CLIP at every scale, showing that the structure of the elicited similarities matters. The contrast between KUEA and DINOv2+KL also highlights the role of the anchor, examined in the ablations.

\begin{table}[t]
\caption{\textbf{Effect of the VLM judge.} Rows identify the training judge; fidelity columns identify the evaluation judge. All use 31 training groups. Blue/red indicates improvement/decline relative to untrained CLIP, scaled per metric. Bold marks column bests.}
\label{tab:judges}
\centering
\begingroup
\footnotesize
\setlength{\tabcolsep}{1.6pt}
\setlength{\arrayrulewidth}{0.4pt}
\arrayrulecolor{black}
\renewcommand{\arraystretch}{1.10}
\resizebox{0.75\linewidth}{!}{%
\begin{tabular}{|l|cc|cccccc|cc|}
\hline
\multirow{2}{*}{Judge} & \multicolumn{2}{c|}{Verdicts} & \multicolumn{6}{c|}{Student fidelity to judge} & \multicolumn{2}{c|}{Student} \\
\cline{2-11}
 & Logp. & Bits & Q3.5-397B & GPT & Kimi & Sonnet & Q3.8-27B & G4-12B & mAP & ZS \\
\hline
Qwen3.5-397B & \checkmark & 2.22 & \cellcolor{blue!22}\textbf{0.79} & \cellcolor{blue!20}0.76 & \cellcolor{blue!24}0.80 & \cellcolor{blue!20}0.74 & \cellcolor{blue!24}\textbf{0.78} & \cellcolor{blue!19}0.68 & \cellcolor{blue!25}\textbf{59.66} & \cellcolor{blue!25}\textbf{62.00} \\
GPT-5.6 Sol & \checkmark & 2.20 & \cellcolor{blue!21}0.75 & \cellcolor{blue!21}\textbf{0.79} & \cellcolor{blue!23}0.79 & \cellcolor{blue!20}0.73 & \cellcolor{blue!24}0.77 & \cellcolor{blue!19}0.68 & \cellcolor{blue!25}58.94 & \cellcolor{blue!20}61.95 \\
Kimi-K3 & \checkmark & 2.34 & \cellcolor{blue!22}0.77 & \cellcolor{blue!20}0.77 & \cellcolor{blue!25}\textbf{0.84} & \cellcolor{blue!20}0.74 & \cellcolor{blue!24}\textbf{0.78} & \cellcolor{blue!19}0.67 & \cellcolor{blue!25}58.96 & \cellcolor{blue!9}61.85 \\
Claude Sonnet 5 & \texttimes & 2.32 & \cellcolor{blue!22}0.77 & \cellcolor{blue!20}0.76 & \cellcolor{blue!23}0.79 & \cellcolor{blue!21}\textbf{0.78} & \cellcolor{blue!23}0.75 & \cellcolor{blue!18}0.65 & \cellcolor{blue!24}58.24 & \cellcolor{blue!16}61.91 \\
Qwen3.8-27B & \checkmark & 1.68 & \cellcolor{blue!21}0.74 & \cellcolor{blue!20}0.75 & \cellcolor{blue!23}0.78 & \cellcolor{blue!18}0.69 & \cellcolor{blue!24}0.76 & \cellcolor{blue!19}0.67 & \cellcolor{blue!25}58.94 & \cellcolor{blue!25}\textbf{62.00} \\
Gemma 4 12B & \checkmark & 2.26 & \cellcolor{blue!20}0.72 & \cellcolor{blue!19}0.72 & \cellcolor{blue!22}0.74 & \cellcolor{blue!17}0.65 & \cellcolor{blue!23}0.74 & \cellcolor{blue!22}\textbf{0.77} & \cellcolor{blue!23}55.94 & 61.81 \\
Untrained CLIP &  &  & 0.25 & 0.29 & 0.21 & 0.27 & 0.17 & 0.23 & 22.06 & 61.81 \\
\hline
\end{tabular}
}
\endgroup
\end{table}

\textbf{Robustness to the choice of judge.}
Table~\ref{tab:judges} compares six judges using 31 training groups and the same 4{,}950 held-out pairs. Logp. indicates whether token log-probabilities support expected-digit scoring; Bits is the entropy of rounded verdicts. Students agree strongly with their training judge (0.76--0.84 fidelity) and other judges (0.65--0.80), suggesting substantial shared similarity structure. Retrieval improves from 22.06 mAP for untrained CLIP to 55.94--59.66 across judges, while zero-shot accuracy remains at 61.81--62.00. Qwen3.5-397B gives the highest retrieval score, but the gains are not restricted to this judge or to models exposing token log-probabilities.

\begin{figure}[t]
  \centering
  \includegraphics[width=0.9\linewidth]{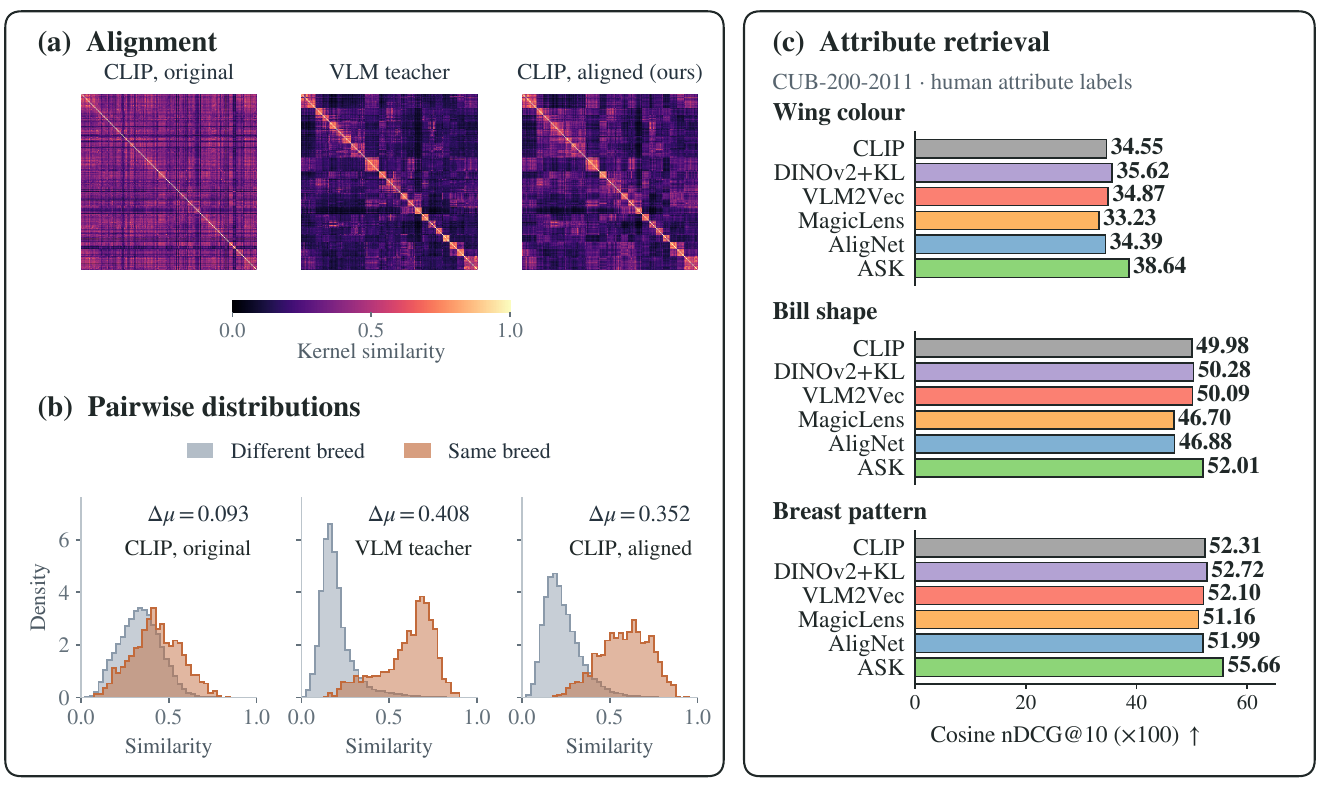}
  \caption{\textbf{Visual geometry and criterion-specific retrieval.} \textbf{(a)} Original CLIP, the VLM breed teacher, and breed-aligned CLIP on 250 held-out dog images. Normalized kernels share breed ordering and a $[0,1]$ colour scale; the teacher kernel is PSD-projected. \textbf{(b)} Corresponding similarity distributions for same-breed and different-breed pairs, each normalized independently; $\Delta\mu$ is the same-breed mean minus the different-breed mean. \textbf{(c)} Cosine retrieval on CUB-200-2011, with human annotations defining relevance for wing colour, bill shape, and breast pattern. Queries retrieve images of other species; bars report nDCG@10 $\times100$ (higher is better). Each \method{} bar uses the corresponding criterion-trained encoder. Human attributes do not supervise \method{}; Appendix~\ref{app:human-attributes} details evaluation and baseline settings.}
  \label{fig:ask-geometry-compact}
\end{figure}

\textbf{Geometry under breed alignment.}
Figure~\ref{fig:ask-geometry-compact}(a,b) shows the effect of breed alignment on 250 held-out dog images. The aligned CLIP kernel recovers the teacher's breed structure, increasing the mean similarity gap between same-breed and different-breed pairs from 0.093 to 0.352, compared with 0.408 for the teacher.

\textbf{Criterion retrieval against human annotations.}
For CUB, we train three criterion-specific students on the same 992 images using Qwen3.8-27B judgments. Figure~\ref{fig:ask-geometry-compact}(c) evaluates cosine retrieval against human annotations for wing colour, bill shape, and breast pattern in CUB-200-2011. Relevance depends on shared attributes, and queries retrieve only other species. These labels do not supervise \method{}. Baselines include CLIP, DINOv2+KL, instruction-conditioned VLM2Vec~\citep{jiang2025vlm2vec} and MagicLens~\citep{zhang2024magiclens}, and human-aligned AligNet~\citep{muttenthaler2025aligning}. Matching \method{} encoders lead all three criteria, exceeding DINOv2+KL, the strongest baseline, by 3.03, 1.73, and 2.94 nDCG@10 points, respectively. They also outperform both mismatched encoders (Table~\ref{tab:human-attributes}(a)), supporting criterion specificity independently of the VLM judge. \method{} and DINOv2+KL use CUB adaptation images; the other learned baselines use released checkpoints. Appendix~\ref{app:human-attributes} gives evaluation details and adds species-disjoint CUB and three-dataset transfer results (Figures~\ref{fig:cub-species-transfer}--\ref{fig:human-criterion-transfer}). \method{} also outperforms the baselines on all three transfer datasets.

\begin{table}[t]
\caption{\textbf{Cross-domain retrieval and criterion specificity on fixed images.} \textbf{(a)} Training domain versus retrieval domain, using unseen-class labels. \textbf{(b)} Training criterion versus held-out teacher, on fixed dog image splits. DINOv2+KL and Shuffled are dog-trained; Shuffled is a negative control with permuted breed targets. Entries are gains over CLIP (absolute scores first). Blue/red denotes improvement/decline; bold marks column bests.}
\label{tab:transfer}
\centering
\begingroup
\footnotesize
\setlength{\tabcolsep}{1.6pt}
\setlength{\arrayrulewidth}{0.4pt}
\arrayrulecolor{black}
\renewcommand{\arraystretch}{1.10}
\resizebox{0.85\linewidth}{!}{%
\begin{tabular}{|l|ccccc|l|ccccc|}
\hline
\multicolumn{6}{|c|}{\textbf{(a) Cross-domain retrieval} ($\Delta$mAP)} & \multicolumn{6}{c|}{\textbf{(b) Criterion specificity} ($\Delta$Spearman)} \\
\cline{1-12}
Elicited on & Dogs & Textures & Flowers & Cars & Pets & Asked about & Breed & \shortstack{Colour/\\lighting} & Pose/view & Coat tex. & Backgr. \\
\hline
\textit{Untrained} & \textit{22.1} & \textit{44.1} & \textit{77.4} & \textit{67.2} & \textit{58.2} & \textit{Untrained} & \textit{0.28} & \textit{0.34} & \textit{0.49} & \textit{0.19} & \textit{0.27} \\
\hline
Dogs & \cellcolor{blue!25}\textbf{+38.1} & \cellcolor{blue!8}+6.2 & \cellcolor{blue!10}\textbf{+8.9} & \cellcolor{blue!13}+14.7 & \cellcolor{blue!20}+29.0 & Breed & \cellcolor{blue!20}\textbf{+0.51} & \cellcolor{blue!7}+0.07 & \cellcolor{red!12}$-$0.25 & \cellcolor{blue!20}+0.53 & \cellcolor{red!9}$-$0.12 \\
Textures & \cellcolor{blue!6}+2.4 & \cellcolor{blue!10}\textbf{+10.4} & \cellcolor{blue!7}+4.0 & \cellcolor{blue!7}+4.1 & \cellcolor{blue!7}+3.8 & Colour/lighting & \cellcolor{blue!11}+0.20 & \cellcolor{blue!17}\textbf{+0.41} & \cellcolor{red!8}$-$0.10 & \cellcolor{blue!10}+0.16 & \cellcolor{blue!10}+0.18 \\
Flowers & \cellcolor{blue!9}+7.3 & \cellcolor{blue!7}+4.6 & \cellcolor{blue!9}+8.4 & \cellcolor{blue!9}+7.4 & \cellcolor{blue!10}+9.9 & Pose/view & \cellcolor{blue!6}+0.02 & \cellcolor{blue!6}+0.02 & \cellcolor{blue!13}\textbf{+0.26} & +0.00 & \cellcolor{blue!6}+0.05 \\
Cars & \cellcolor{blue!11}+10.9 & \cellcolor{blue!7}+3.2 & \cellcolor{blue!9}+7.0 & \cellcolor{blue!14}\textbf{+17.9} & \cellcolor{blue!12}+12.8 & Coat tex. & \cellcolor{blue!17}+0.41 & \cellcolor{red!6}$-$0.04 & \cellcolor{red!13}$-$0.28 & \cellcolor{blue!25}\textbf{+0.68} & \cellcolor{red!10}$-$0.17 \\
Pets & \cellcolor{blue!24}+35.5 & \cellcolor{blue!7}+4.3 & \cellcolor{blue!9}+7.8 & \cellcolor{blue!12}+13.2 & \cellcolor{blue!21}\textbf{+31.1} & Backgr. & \cellcolor{red!8}$-$0.10 & \cellcolor{blue!8}+0.10 & \cellcolor{red!11}$-$0.22 & \cellcolor{red!8}$-$0.10 & \cellcolor{blue!24}\textbf{+0.65} \\
\hline
DINOv2+KL & \cellcolor{blue!21}+29.9 & \cellcolor{blue!7}+3.5 & \cellcolor{blue!8}+5.9 & \cellcolor{blue!11}+10.6 & \cellcolor{blue!17}+23.8 & DINOv2+KL & \cellcolor{blue!13}+0.26 & \cellcolor{red!6}$-$0.03 & \cellcolor{red!11}$-$0.22 & \cellcolor{blue!14}+0.30 & \cellcolor{red!7}$-$0.08 \\
Shuffled (neg.) & \cellcolor{red!7}$-$4.1 & \cellcolor{red!7}$-$4.4 & \cellcolor{red!13}$-$14.3 & \cellcolor{red!8}$-$6.1 & \cellcolor{red!13}$-$14.3 & Shuffled (neg.) & \cellcolor{red!7}$-$0.08 & \cellcolor{red!10}$-$0.16 & \cellcolor{red!12}$-$0.23 & \cellcolor{red!6}$-$0.03 & \cellcolor{red!9}$-$0.12 \\
\hline
\end{tabular}
}
\endgroup
\end{table}

\textbf{Transfer and criterion specificity.}
Table~\ref{tab:transfer}(a) measures cross-domain class retrieval: all 25 combinations improve, with cross-domain gains of 2.4--35.5 mAP points. Panel (b) fixes the dog images: each matching student outperforms every mismatched student and both controls against its teacher. This within-column comparison supports specificity. Off-diagonal declines alone do not: breed alignment lowers pose agreement by 0.25, versus 0.23 for shuffled supervision. Shuffled is a single negative control using permuted breed targets. All adapted models use 93 groups; Dogs and Breed share a checkpoint. Appendix~\ref{app:criterion-transfer} separately tests transfer of the same criteria to non-dog images.

\subsection{Ablations}
\label{sec:ablations}
\label{sec:regablation}

\begin{table}[t]
\caption{\textbf{Ablations of the alignment recipe.} (a) Raw minus PSD for the four cases with highest negative spectral mass (Neg.\%). (b) Supervision targets under matched MSE, KL, and LoRA; mAP on unseen breeds. (c,d) KL-anchored heads versus LoRA. (e) Anchor regularisation. (f) Training objective. Blue/red indicates improvement/decline relative to PSD in (a), LoRA in (c,d), and the default in (b,e,f); bold marks metric bests.}
\label{tab:ablation}
\centering
\resizebox{\linewidth}{!}{%
\begin{minipage}{\linewidth}
\fontsize{6.5}{7.8}\selectfont
\setlength{\tabcolsep}{1.4pt}
\setlength{\arrayrulewidth}{0.4pt}
\arrayrulecolor{black}
\renewcommand{\arraystretch}{1.12}
\begin{minipage}[t]{0.343\linewidth}
\noindent\textbf{(a) Raw vs.\ PSD teacher}\par\vspace{1pt}
\begin{tabular}{|l|c|ccc|}
\hline
Judge & Neg.\% & $\Delta$Fid & $\Delta$mAP & $\Delta$ZS \\
\hline
Gemma 4 12B & 6.4 & \cellcolor{red!25}$-$0.009 & \cellcolor{red!25}$-$1.43 & \cellcolor{red!25}$-$0.41 \\
GPT-5.6 & 3.8 & \cellcolor{red!16}$-$0.005 & \cellcolor{red!12}$-$0.53 & \cellcolor{red!19}$-$0.29 \\
Qwen3.5 & 3.5 & \cellcolor{blue!8}$+$0.002 & \cellcolor{red!25}$-$1.45 & \cellcolor{red!18}$-$0.26 \\
Sonnet 5 & 2.6 & \cellcolor{red!7}$-$0.001 & \cellcolor{red!7}$-$0.17 & \cellcolor{red!16}$-$0.23 \\
\hline
\end{tabular}
\par\vspace{5pt}
\noindent\textbf{(b) Supervision target}\par\vspace{1pt}
\begin{tabular}{|l|cc|}
\hline
Target & Unseen mAP & ZS \\
\hline
Binary labels & \cellcolor{red!25}43.11 & \cellcolor{red!25}59.83 \\
WordNet & \cellcolor{red!15}52.01 & \cellcolor{red!18}60.70 \\
VLM class-pair mean & \cellcolor{red!9}57.17 & \cellcolor{red!9}61.77 \\
\method{} & \textbf{60.16} & \textbf{62.28} \\
\hline
\end{tabular}
\end{minipage}%
\hfill
\begin{minipage}[t]{0.305\linewidth}
\noindent\textbf{(c) Heads vs.\ LoRA: Fid}\par\vspace{1pt}
\begin{tabular}{|l|ccc|}
\hline
Criterion & Linear & MLP & \textbf{LoRA} \\
\hline
Breed & \cellcolor{red!18}0.622 & \cellcolor{red!10}0.734 & \textbf{0.794} \\
Colour/lighting & \cellcolor{red!25}0.491 & \cellcolor{red!11}0.673 & \textbf{0.747} \\
Pose/view & \cellcolor{red!19}0.564 & \cellcolor{red!11}0.674 & \textbf{0.748} \\
Coat texture & \cellcolor{red!19}0.699 & \cellcolor{red!9}0.826 & \textbf{0.874} \\
Background & \cellcolor{red!20}0.730 & \cellcolor{red!8}0.887 & \textbf{0.919} \\
\hline
\end{tabular}
\par\vspace{5pt}
\noindent\textbf{(d) Heads vs.\ LoRA: $\Delta$mAP}\par\vspace{1pt}
\begin{tabular}{|l|ccc|}
\hline
Retrieval on & Linear & MLP & \textbf{LoRA} \\
\hline
Unseen breeds & \cellcolor{red!25}$+$13.9 & \cellcolor{red!18}$+$22.8 & \textbf{$+$38.1} \\
Non-dog & \cellcolor{red!8}$+$1.3 & \cellcolor{red!7}$+$1.9 & \textbf{$+$4.9} \\
Other domains & \cellcolor{red!13}$+$5.5 & \cellcolor{red!10}$+$8.2 & \textbf{$+$14.7} \\
\hline
\end{tabular}
\end{minipage}%
\hfill
\begin{minipage}[t]{0.337\linewidth}
\noindent\textbf{(e) Anchor regulariser}\par\vspace{1pt}
\begin{tabular}{|l|c|ccc|}
\hline
Regulariser & $w$ & Fid & mAP & ZS \\
\hline
\textbf{KL (\method{})} & .2 & 0.794 & \textbf{60.16} & \textbf{62.28} \\
None & --- & \cellcolor{blue!5}\textbf{0.796} & \cellcolor{red!5}59.41 & \cellcolor{red!25}56.67 \\
Feature L2 & 1 & \cellcolor{red!25}0.286 & \cellcolor{red!25}22.29 & \cellcolor{red!6}61.91 \\
Text L1 only & .5 & \cellcolor{blue!5}\textbf{0.796} & \cellcolor{red!5}59.30 & \cellcolor{red!25}56.70 \\
KL, text frozen & .2 & \cellcolor{red!7}0.752 & \cellcolor{red!11}49.71 & \cellcolor{red!8}61.35 \\
\hline
\end{tabular}
\par\vspace{5pt}
\noindent\textbf{(f) Training objective}\par\vspace{1pt}
\begin{tabular}{|l|ccc|}
\hline
Objective & Fid & mAP & ZS \\
\hline
Cosine MSE & \cellcolor{red!25}0.756 & \cellcolor{red!12}58.35 & \cellcolor{blue!25}\textbf{62.57} \\
\textbf{Cubic MSE (\method{})} & 0.794 & \textbf{60.16} & 62.28 \\
Cubic ranking & \cellcolor{blue!8}\textbf{0.799} & \cellcolor{red!25}54.71 & \cellcolor{red!9}62.23 \\
\hline
\end{tabular}
\end{minipage}%
\end{minipage}%
}
\end{table}

\textbf{PSD projection and supervision targets.}
Table~\ref{tab:ablation}(a) shows four cases from Table~\ref{tab:psd}. Under matched 31-group raw/PSD training, projection improves retrieval by 0.17--1.45 mAP points and zero-shot accuracy by 0.23--0.41 points across these cases. Gemma 4 12B has the largest negative mass (6.4\%) and gains 0.009 fidelity, 1.43 mAP, and 0.41 ZS. Qwen car-pose targets have 3.5\% negative mass: projection gains 1.45 car-model mAP and 0.26 ZS, with a 0.002 fidelity decrease. These results support PSD correction as an effective part of alignment when elicited targets have substantial negative spectral mass. Panel (b) fixes kernel MSE, KL, and LoRA: binary labels, WordNet \citep{miller1995wordnet}, and VLM class-pair means yield 43.11, 52.01, and 57.17 unseen-breed mAP, versus 60.16 for individual judgments. Appendix~\ref{app:label-kernels} gives the full comparison.

\textbf{Frozen heads versus encoder adaptation.}
Table~\ref{tab:ablation}(c,d) compares linear and MLP heads with LoRA using the same criterion-specific targets, cubic objective, and KL anchor. Across five criteria, mean fidelity is 0.816 for LoRA, 0.621 for the linear head, and 0.759 for the MLP. The breed-trained models improve unseen-breed mAP over CLIP by 38.1, 13.9, and 22.8 points, respectively. Both heads also improve non-dog and cross-domain retrieval, while LoRA achieves larger gains. The MLP matches LoRA in visual parameter count; these comparisons use a shared learning rate. Full retrieval results and the protocol are in Appendix~\ref{app:heads-results}.

\textbf{Preserving zero-shot performance.}
Table~\ref{tab:ablation}(e) shows that removing the KL anchor leaves fidelity nearly unchanged (0.796 versus 0.794) but lowers zero-shot accuracy from 62.28 to 56.67. An L1 penalty on text-feature drift alone provides little protection (56.70 ZS). An L2 penalty on image-feature drift at the tested weight of 1.0 reduces fidelity to 0.286. Freezing the text tower during KL-anchored training lowers retrieval by 10.45 mAP points. Additional regularisers and weight sweeps are reported in Appendix~\ref{app:regulariser-results}.

\textbf{Kernel form and sensitivity.}
Table~\ref{tab:ablation}(f) compares cosine MSE, normalized-cubic MSE, and normalized-cubic ranking with the same LoRA student, PSD targets, and KL anchor. Fidelity uses each training similarity against raw held-out judgments, with no link fitted on test targets. Cubic MSE reaches 0.794 Spearman and 60.16 mAP, versus 0.756 and 58.35 for cosine MSE; their zero-shot means are 62.28 and 62.57. Ranking raises Spearman slightly to 0.799 but lowers mAP to 54.71. Appendix~\ref{app:kernel-form} additionally evaluates every model with cosine: the cubic-MSE model retains 0.794 Spearman, while Pearson changes from 0.840 to 0.792. These comparisons use a shared learning rate and anchor weight. Sensitivity and reduced-query supervision are examined in Appendices~\ref{app:sweeps} and~\ref{app:streaming-results}, respectively.

\section{Conclusion}
\label{sec:conclusion}
\method{} transfers VLM similarity preferences into CLIP through positive semidefinite kernel matching and joint image--text adaptation. Across five domains, four backbones, and six judges, it improves unseen-class retrieval while an image--text anchor supports recognition retention. Retrieval uses image embeddings and cosine similarity, without VLM calls. Prompt-selected criteria additionally control the learned similarity, with human-annotated evaluations supporting attribute, viewpoint, pose, and background retrieval across four datasets, including species excluded from adaptation.

\subsection*{AI use statement}

Large language models assisted with writing and refining the manuscript. The authors reviewed all text and take full responsibility for the paper's content.

\clearpage
\appendix
\section{Appendix}
\label{app:contents}
\addtocontents{toc}{\protect\setcounter{tocdepth}{2}}
\begingroup
\small
\renewcommand{\contentsname}{Contents}
\hypersetup{linktoc=all}
\tableofcontents
\endgroup
\clearpage

\section{Related work}
\label{app:related-work}

\textbf{Visual representations and perceptual similarity.}
CLIP learns visual representations through image--text contrastive supervision \citep{radford2021learning}, while DINOv2 learns transferable visual features through self-supervision \citep{oquab2024dinov2}. Strong recognition performance does not ensure sensitivity to every visual distinction: \citet{tong2024eyes} identify image pairs that CLIP represents similarly despite differences relevant to visual reasoning. Perceptual metrics address a related problem from human judgments. LPIPS calibrates deep features for perceptual comparison \citep{zhang2018unreasonable}; DISTS combines structure and texture similarity while tolerating texture resampling \citep{ding2022dists}; and DreamSim learns from human comparisons of synthetic image triplets \citep{fu2023dreamsim}. AligNet distils synthetic targets from a human-aligned teacher into criterion-independent visual representations \citep{muttenthaler2025aligning}. These works motivate measuring more than category identity. ASK learns from a selected VLM's criterion-dependent judgments. Its fidelity measures teacher agreement, while human-annotated attribute retrieval independently evaluates the selected visual distinctions; class-based retrieval and zero-shot classification assess category retrieval and recognition retention.

\textbf{Conditional similarity and instruction-guided representations.}
Conditional Similarity Networks learn masks over embedding dimensions to represent different notions of similarity \citep{veit2017conditional}. GeneCIS evaluates whether a model can retrieve images according to changing object and attribute conditions, and learns conditional similarity from image-caption data \citep{vaze2023genecis}. FocalLens uses instruction tuning to produce image representations conditioned on natural-language instructions \citep{hsieh2025focallens}. MagicLens learns image-and-instruction embeddings from web image pairs and synthesized instructions \citep{zhang2024magiclens}. Omni-Attribute learns open-vocabulary attribute embeddings using image pairs annotated with shared and differing attributes, supporting retrieval and visual concept personalization \citep{chen2026omniattribute}. VLM2Vec takes a broader approach, converting VLMs into instruction-aware multimodal embedding models through contrastive training \citep{jiang2025vlm2vec}. ASK shares the goal of controlling which visual information determines similarity, but incorporates the criterion through teacher supervision during adaptation. Each resulting encoder can embed an image without an accompanying instruction; changing to a previously untrained criterion requires further adaptation.

\textbf{Model-based evaluation and continuous score extraction.}
Language and multimodal models can provide evaluation signals as well as generated answers. G-Eval combines prompted evaluation with probability-weighted rating scores for text generation \citep{liu2023geval}. Q-Align trains multimodal models on discrete quality levels and converts their output probabilities into continuous visual scores \citep{wu2024qalign}. LLaVA-Critic learns to evaluate multimodal responses and provides feedback for preference learning \citep{xiong2025llavacritic}. These approaches establish precedents for using model judgments and for extracting finer-grained scores from discrete responses. ASK applies probability-weighted digit scores, when available, to criterion-specific image pairs and uses the resulting similarities to supervise an encoder. Our contribution is the transfer of these judgments into visual geometry, rather than a new principle for converting rating probabilities into scalar scores.

\textbf{Relational distillation and representation comparison.}
Relational knowledge distillation transfers distances and angles between examples \citep{park2019relational}, and similarity-preserving distillation matches pairwise activation similarities \citep{tung2019similarity}. Both show that a student can learn a teacher's relationships without matching its feature coordinates. Related tools compare representation structure: centered kernel alignment measures agreement between representational similarity matrices \citep{kornblith2019similarity}; SPEC identifies differences in sample clustering through spectral analysis \citep{jalali2025towards}; and KODA discovers subsets organized differently across vision-language representations \citep{wu2026koda}. ASK adopts this relational perspective, using pairwise geometry as the target of adaptation. The target is supplied by an elicited criterion-dependent judgment, rather than derived solely from another encoder's features.

\textbf{Kernel alignment with elicited targets.}
KUEA aligns CLIP's visual geometry to a vision teacher such as DINOv2 through kernel matching \citep{gong2025kernel}. Its objective updates the visual encoder and anchors its features to frozen CLIP with an L2 penalty, without using text during adaptation. ASK builds on this foundation in two respects. First, its target is elicited from a VLM under a chosen criterion, allowing the same image set to support different target geometries without access to teacher embeddings. Second, its retention objective acts on image--text prediction distributions and jointly updates adapters in both CLIP encoders. The text supervision is a shared prompt bank; it does not require a newly annotated image--caption dataset. A feature Gram matrix is PSD by construction, whereas elicited scores need not be, so we apply classical spectral PSD projection \citep{higham1988computing} and diagonal normalization within each scored group. The methodological contribution is the use of elicited criterion-dependent targets with joint distributional anchoring; kernel matching and PSD projection themselves are established tools. Our DINOv2+KL control uses the same anchor and student configuration to distinguish the supervision source from the change in regularization relative to KUEA.

\textbf{Efficient adaptation of vision-language models.}
CoOp learns text context vectors for recognition \citep{zhou2022learning}, and MaPLe couples learned prompts across the vision and language branches \citep{khattak2023maple}. CLIP-Adapter learns residual feature adapters \citep{gao2024clipadapter}, while Tip-Adapter constructs a cache from labeled examples for few-shot classification \citep{zhang2022tipadapter}. LoRA provides low-rank parameter updates \citep{hu2022lora}, and CLIP-LoRA applies them to few-shot vision-language adaptation \citep{zanella2024low}. ASK uses LoRA as an adaptation mechanism, with a different objective: reproducing criterion-specific pairwise similarities while retaining useful image--text predictions. The criterion is expressed in the teacher prompt, rather than supplied as a class prediction target.

\textbf{Retaining pretrained capabilities.}
PromptSRC regularizes prompt adaptation against pretrained CLIP to reduce forgetting \citep{khattak2023self}. WiSE-FT interpolates pretrained and fine-tuned weights to improve robustness under distribution shift \citep{wortsman2022robust}. Robust CLIP adapts the visual encoder through an unsupervised adversarial objective, illustrating another setting in which visual features must change while remaining useful in a vision-language system \citep{schlarmann2024robust}. ASK regularizes image--text predictions during similarity alignment, so both visual and textual adapters participate in optimization. Our regularizer ablation shows higher fidelity with trainable text adapters than with a frozen text tower at each tested weight of the default KL anchor. Both adapted encoders are retained at deployment. Zero-shot evaluation measures retention of recognition capabilities after joint adaptation, separately from agreement with the VLM judge.

\section{Experimental setup details}
\label{app:experimental-setup}

\subsection{Datasets and splits}
\label{app:datasets}

\textbf{Image sources.}
We use ImageNet dog breeds \citep{deng2009imagenet}, Describable Textures (DTD; \citealp{cimpoi2014describing}), Oxford Flowers \citep{nilsback2008automated}, Stanford Cars \citep{krause2013collecting}, and Oxford-IIIT Pets \citep{parkhi2012cats}. Dog images are drawn from the ImageNet training partition. The other four domains use the CoOp dataset splits \citep{zhou2022learning}. Retrieval galleries contain 20 images per unseen class: 25 classes each for Dogs, Flowers, and Cars (500 images per domain), 20 for DTD (400 images), and 12 for Pets (240 images). A separate non-dog ImageNet gallery uses 50 classes with four images each (200 images) to evaluate transfer of dog-trained models.

\textbf{Adaptation and held-out fidelity.}
Each domain contains 25 adaptation classes, with 40 training images and 10 held-out images per class, giving 1{,}000 training and 250 held-out images. Training and held-out images are disjoint, but their classes are shared. Fidelity therefore measures agreement with the teacher on new images of the adaptation classes. The full held-out set contains $\binom{250}{2}=31{,}125$ unordered off-diagonal pairs. Matched cross-judge comparisons instead use a shared 100-image test subset, giving 4{,}950 pairs.

\textbf{Validation split.}
For Dogs, we additionally sample 10 images from each of the 25 adaptation breeds, giving 250 validation images. These share classes with the adaptation set but are image-disjoint from training, the held-out fidelity set, and all reported evaluation galleries and benchmark sets. Validation images are excluded from teacher-target construction, kernel fitting, and adapter training, and are not added to training after hyperparameter selection. We provide their image identifiers in the experimental manifests.

\textbf{Retrieval on unseen classes.}
The retrieval sets use classes excluded from adaptation in the corresponding domain. Each image queries the remaining images in its set, with relevance defined by class-label agreement rather than teacher scores. Here, ``unseen'' refers to our adaptation data, not to CLIP's pretraining corpus.

For a query $q$ in a set of $N$ images, let $r_q(k)$ indicate whether the image at rank $k$ has the same class label, and let $R_q=\sum_{k=1}^{N-1}r_q(k)$. We compute
\begin{equation}
\operatorname{AP}(q)=\frac{1}{R_q}\sum_{k=1}^{N-1}r_q(k)\frac{\sum_{j=1}^{k}r_q(j)}{k},
\qquad
\operatorname{mAP}=\frac{100}{N}\sum_{q=1}^{N}\operatorname{AP}(q).
\label{eq:retrieval-ap}
\end{equation}
The query contributes neither a ranked result nor a relevant item: $R_q=19$ in the five domain galleries and $R_q=3$ in the non-dog gallery. Reported mAP values are percentages and gains are percentage-point differences. Paired permutation tests use 10{,}000 random sign flips of the query-level AP differences from CLIP and compare absolute mean differences.

\textbf{Criteria on shared images.}
The main criteria concern breed, build, coat and markings for dogs; breed and coat for pets; pattern structure for textures; body style and model for cars; and petal shape and arrangement for flowers. Additional criteria cover whole-image colour and lighting; pose, viewpoint, and framing; coat texture (dogs); and background. ``Colour/lighting'' and ``Pose/view'' are compact labels for the first two composite criteria, not isolated coat/paint colour or body pose. Within each domain, criterion comparisons reuse the same images and splits, so changing the criterion changes the supervision rather than the image set. Verbatim scoring prompts are provided in Appendix~\ref{app:prompts}.

\textbf{Zero-shot evaluation.}
ZS is the unweighted mean top-1 accuracy on ImageNet \citep{deng2009imagenet}, CIFAR-10 and CIFAR-100 \citep{krizhevsky2009learning}, Caltech-101 \citep{feifei2004learning}, FER2013 \citep{goodfellow2013challenges}, Oxford-IIIT Pets \citep{parkhi2012cats}, DTD \citep{cimpoi2014describing}, RESISC45 \citep{cheng2017remote}, EuroSAT \citep{helber2019eurosat}, PCam \citep{veeling2018rotation}, ImageNet-Sketch \citep{wang2019robust}, and ImageNet-O \citep{hendrycks2021natural}. The reported scores use the jointly adapted visual and text encoders, measuring zero-shot recognition after joint image--text adaptation.

\subsection{Training and evaluation protocols}
\label{app:training-protocol}

\textbf{Feature teachers.}
DINOv2+KL uses the frozen DINOv2 ViT-L/14 with four register tokens \citep{oquab2024dinov2,darcet2024registers}, loaded from \path{facebookresearch/dinov2} through the Torch Hub entry point \path{dinov2_vitl14_reg}, with weights \path{dinov2_vitl14_reg4_pretrain.pth}. We extract the final layer-normalized [CLS] token. RGB images are bicubically resized to a shorter side of 224 pixels, centre-cropped to $224\times224$, and normalized with ImageNet mean $(0.485,0.456,0.406)$ and standard deviation $(0.229,0.224,0.225)$. The SigLIP \citep{zhai2023sigmoid} feature-teacher control uses \path{google/siglip-so400m-patch14-384}, extracting \texttt{pooler\_output} from \texttt{SiglipVisionModel}. Its supplied \texttt{SiglipImageProcessor} resizes RGB images to $384\times384$ and normalizes each channel with mean and standard deviation $0.5$. Both feature teachers remain frozen.

\textbf{Scoring budget.}
Training images are arranged into stratified groups of 32, with eight classes and four images per class. Each group contains $\binom{32}{2}=496$ unordered pairs. The default Qwen3.5-397B protocol uses 93 groups, giving 46{,}128 pair slots and 44{,}450 distinct pairs in the dog run; the difference reflects pairs repeated across groups. Other judges use 31 groups (15{,}376 pairs), and matched cross-judge comparisons use this same budget for Qwen3.5-397B. Reduced-budget experiments reuse the first 31 or 62 groups of the default protocol.

\textbf{Kernel fitting.}
The default targets are PSD-projected teacher similarities. We fit the cubic link parameters by least squares on frozen-encoder features before adaptation, then keep them fixed during training and evaluation. The link is fitted separately for each teacher and backbone. Kernel normalization follows the method section. Fidelity compares the student's normalized kernel with the named teacher's raw scores over unordered off-diagonal held-out pairs. Every arm is evaluated against unprojected judgments, so PSD projection changes the training targets rather than the evaluation target.

\textbf{Shuffled-score control.}
We permute each group's off-diagonal entries of the dog primary-criterion target, preserving their distribution while destroying the correspondence between image pairs and judgments. Table~\ref{tab:transfer} uses this single dog-trained negative control across all evaluation columns. It checks the importance of meaningful pair supervision; it does not isolate criterion-specific suppression. Criterion specificity is assessed by comparing matching and mismatched students against the same held-out teacher.

\textbf{Optimization.}
For \method{}, we train rank-32 LoRA adapters with scaling $\alpha=64$ in the query, key, value, and output projections of both CLIP towers, keeping pretrained weights frozen. Training uses AdamW \citep{loshchilov2019decoupled} for 3{,}000 steps, each using one scored group of 32 images, with 200 warm-up steps. For \method{}, SupCon, and DINOv2+KL, we search learning rates in $\{5\mathrm{e}{-3}, 1\mathrm{e}{-4}, 5\mathrm{e}{-5}, 1\mathrm{e}{-5}\}$ and KL weights in $\{0.01, 0.1, 0.2, 1.0\}$, selecting each method and backbone separately on the same Dogs validation set. \method{} tuning uses the default Qwen3.5-397B breed judgments. Each configuration is trained for 3{,}000 updates and evaluated using its final checkpoint. Selection maximizes validation retrieval mAP: each validation image queries the other 249 images by cosine similarity, with the nine images of the same breed defining relevance. The selected settings are reused across the other domains. Reported fidelity, retrieval, and zero-shot test scores do not enter selection. The KL anchor uses a bank of 1{,}000 ImageNet class-name prompts. In the CLIP vision-transformer backbone comparison \citep{radford2021learning,dosovitskiy2021image}, ViT-B/32 and ViT-B/16 use the full bank at each step, while ViT-L/14 and ViT-L/14-336 sample 200 prompts per step. Experiments were run on four NVIDIA RTX A6000 GPUs.

\textbf{Geometry visualization protocol.}
Figure~\ref{fig:ask-geometry-compact}(a,b) uses the same 250 held-out dog images, ordered by breed. VLM targets are PSD-projected kernels. Original and aligned CLIP use the same fitted cubic link on unnormalized visual features, followed by diagonal normalization. Histograms include all 31{,}125 unordered off-diagonal pairs, with shared bins and axis limits and independently normalized same-breed and different-breed densities. The reported $\Delta\mu$ is the mean same-breed similarity minus the mean different-breed similarity. Panel~(c) evaluates cosine retrieval against human attribute annotations on CUB-200-2011, following Appendix~\ref{app:human-attributes}.

\textbf{Ablation protocols.}
\label{app:ablation-protocol}
Unless varied, ablations use learning rate $1\mathrm{e}{-5}$ and KL weight $0.2$, without separate tuning for each ablation.
Table~\ref{tab:ablation}(a) uses 31 groups per teacher and criterion, matched within each raw/PSD pair. Dog-breed fidelity uses the same 100 test images (4{,}950 pairs); Qwen car-pose fidelity uses 250 test images (31{,}125 pairs). Retrieval uses 500 images from 25 unseen breeds or car-model classes, respectively. Neg.\% is the fraction of absolute eigenvalue mass on negative eigenvalues, averaged over training groups. Table~\ref{tab:psd} includes all six comparisons. Panels (b--f) use 93 groups; their default uses Qwen3.5-397B expected-digit scores, PSD targets, and the default KL anchor with weight 0.2. Panel (b) compares binary labels, WordNet, and VLM class-pair means with individual judgments under shared kernel MSE, KL, and LoRA (Table~\ref{tab:label-kernels}). Verdict extraction is examined in Appendix~\ref{app:verdict-extraction}. In (c), each criterion has separately trained linear, MLP, and LoRA models, all with the same KL anchor. The MLP has 2.36M visual parameters, matching visual LoRA, while the linear head has 0.26M; each model also trains 1.57M text parameters. Panel (d) evaluates the breed-trained models and reports mAP gains over untrained CLIP; ``Other domains'' averages Textures, Flowers, Cars, and Pets. The default breed LoRA checkpoint is shared across panels (b--f). Panel (e) retains selected anchor ablations at the listed weights; the full regulariser sweep is in Table~\ref{tab:regulariser-full}. Panel (f) compares cosine MSE, normalized-cubic MSE, and normalized-cubic ranking with the same LoRA and KL settings. Its fidelity uses each training similarity with fixed training-fitted parameters. The full head and kernel comparisons are in Tables~\ref{tab:heads} and~\ref{tab:kernelform}.

\subsection{Supervised contrastive baseline}
\label{app:supcon-repair}

SupCon uses the same training images and 93 groups as the corresponding ASK run, with rank-32 LoRA in both towers, AdamW, 200 warm-up steps, and 3,000 updates. The supervised contrastive loss uses temperature $0.07$, treats other images of the same class as positives, and excludes self-pairs. It replaces the visual kernel-regression objective while retaining KL anchoring.

Both methods use the same prompt construction, group sampler, optimizer, and scheduling implementation. The anchor uses all 1,000 ImageNet class-name prompts per step on ViT-B/16 and ViT-B/32, and the corresponding ASK run's 200-prompt subsampling on ViT-L/14 and ViT-L/14-336. In the non-ImageNet domains, labels determine the contrastive positives; they do not select prompts from the full anchor bank. The schedule follows the reference implementation's update ordering, including its initial full-learning-rate update before warm-up.

SupCon and DINOv2+KL follow the validation protocol in Appendix~\ref{app:training-protocol}, using the same search grid, validation images, and selection metric as \method{}, with hyperparameters selected separately for each method. Every candidate uses its final 3{,}000-update checkpoint; intermediate checkpoints are not selected. Fidelity uses the corresponding ASK link fitted on training data, while retrieval uses cosine similarity with the query excluded.

\section{Additional results}
\label{app:additional-results}

\subsection{Per-benchmark zero-shot accuracy}
\label{app:zeroshot}

\begin{table}[t]
\caption{\textbf{Per-benchmark zero-shot accuracy across CLIP backbones.} Top-1 accuracy (\%); Mean is the unweighted average over 12 benchmarks. Blue/red cells indicate gains/losses relative to untrained CLIP within each backbone, scaled separately per column across backbones; bold marks the best result per column.}
\label{tab:zeroshot}
\centering
\scriptsize
\setlength{\tabcolsep}{1.6pt}
\setlength{\arrayrulewidth}{0.4pt}
\arrayrulecolor{black}
\renewcommand{\arraystretch}{1.15}
\resizebox{0.9\linewidth}{!}{%
\begin{tabular}{|l|l|cccccccccccc|c|}
\hline
 & Method & IN-1k & C-10 & C-100 & Caltech & FER & Pets & DTD & RESISC & EuroSAT & PCam & IN-Sk. & IN-O & Mean \\
\hline
\multirow{8}{*}{\rotatebox[origin=c]{90}{ViT-B/32}}
 & Untrained CLIP & 63.5 & \textbf{89.7} & 63.3 & \textbf{81.6} & 41.4 & 87.3 & \textbf{44.3} & 53.6 & 50.4 & \textbf{62.3} & 42.3 & 47.8 & \textbf{60.6} \\
 & Shuffled teacher & \cellcolor{red!9}62.6 & \cellcolor{red!12}87.6 & \cellcolor{red!19}59.3 & \cellcolor{red!13}80.6 & \cellcolor{red!25}36.7 & \cellcolor{red!9}86.0 & \cellcolor{red!15}42.6 & \cellcolor{red!18}51.0 & \cellcolor{red!20}42.0 & \cellcolor{red!5}62.1 & \cellcolor{red!10}41.4 & \cellcolor{red!9}47.0 & \cellcolor{red!16}58.2 \\
 & SupCon & \cellcolor{red!25}58.7 & \cellcolor{red!25}83.5 & \cellcolor{red!25}57.5 & \cellcolor{red!25}79.1 & \cellcolor{blue!24}\textbf{45.8} & \cellcolor{red!25}79.9 & \cellcolor{red!25}40.9 & \cellcolor{red!20}50.7 & \cellcolor{red!25}38.9 & \cellcolor{red!14}57.4 & \cellcolor{red!23}38.9 & \cellcolor{red!25}43.5 & \cellcolor{red!25}56.2 \\
 & KUEA & 63.5 & \cellcolor{red!5}89.6 & \cellcolor{blue!5}63.4 & \textbf{81.6} & \cellcolor{blue!5}41.5 & \cellcolor{blue!5}87.4 & \cellcolor{red!7}44.0 & 53.6 & \cellcolor{blue!7}51.4 & \cellcolor{red!5}62.1 & \cellcolor{blue!6}\textbf{42.4} & \cellcolor{red!7}47.3 & \textbf{60.6} \\
 & DINOv2 $+$ KL & 63.5 & \cellcolor{red!12}87.6 & \cellcolor{red!9}62.0 & \cellcolor{red!10}81.0 & \cellcolor{red!22}37.4 & \cellcolor{blue!7}88.0 & \cellcolor{red!17}42.2 & \cellcolor{red!6}53.4 & \cellcolor{red!11}46.7 & \cellcolor{red!8}60.7 & \cellcolor{red!7}42.0 & \cellcolor{red!8}47.2 & \cellcolor{red!11}59.3 \\
 & SigLIP tower & \cellcolor{blue!5}\textbf{63.6} & \cellcolor{red!7}89.0 & \cellcolor{red!6}63.0 & \cellcolor{red!7}81.3 & \cellcolor{red!14}39.3 & \cellcolor{blue!7}\textbf{88.2} & \cellcolor{red!9}43.6 & \cellcolor{blue!9}\textbf{54.3} & \cellcolor{blue!8}\textbf{52.3} & \cellcolor{red!7}61.4 & 42.3 & \cellcolor{blue!7}\textbf{48.2} & \cellcolor{red!5}60.5 \\
 & Qwen2.5-VL tower & \cellcolor{red!5}63.4 & \cellcolor{red!8}88.9 & \cellcolor{blue!7}\textbf{64.0} & \cellcolor{red!10}81.0 & \cellcolor{red!11}39.9 & \cellcolor{blue!7}88.0 & \cellcolor{red!6}44.1 & \cellcolor{red!7}53.3 & \cellcolor{blue!5}50.5 & \cellcolor{red!7}61.3 & \cellcolor{red!6}42.2 & \cellcolor{red!5}47.7 & \cellcolor{red!6}60.4 \\
 & \textbf{\method{} (ours)} & \cellcolor{red!10}62.4 & \cellcolor{red!12}87.6 & \cellcolor{red!10}61.9 & \cellcolor{red!15}80.3 & \cellcolor{red!19}38.0 & \cellcolor{red!7}86.7 & \cellcolor{red!17}42.2 & \cellcolor{red!6}53.4 & \cellcolor{red!8}48.4 & \cellcolor{red!8}60.7 & \cellcolor{red!9}41.5 & \cellcolor{red!7}47.3 & \cellcolor{red!11}59.2 \\
\hline
\multirow{8}{*}{\rotatebox[origin=c]{90}{ViT-B/16}}
 & Untrained CLIP & 68.3 & 90.0 & 65.6 & 82.2 & 46.3 & 89.0 & 44.9 & 58.2 & 55.9 & 50.7 & 48.3 & 42.3 & 61.8 \\
 & Shuffled teacher & \cellcolor{red!11}66.9 & \cellcolor{red!15}86.8 & \cellcolor{red!15}62.7 & \cellcolor{red!11}81.5 & \cellcolor{red!7}45.9 & \cellcolor{red!14}85.6 & 44.9 & \cellcolor{red!25}54.3 & \cellcolor{red!12}51.9 & \cellcolor{blue!11}\textbf{54.2} & \cellcolor{red!10}47.4 & \cellcolor{red!10}41.2 & \cellcolor{red!12}60.3 \\
 & SupCon & \cellcolor{red!24}63.7 & \cellcolor{red!13}87.5 & \cellcolor{red!17}62.2 & \cellcolor{red!25}79.7 & \cellcolor{red!24}41.9 & \cellcolor{red!16}84.8 & \cellcolor{red!14}43.3 & \cellcolor{red!7}57.8 & \cellcolor{red!10}53.2 & \cellcolor{blue!9}53.0 & \cellcolor{red!25}44.6 & \cellcolor{red!20}39.0 & \cellcolor{red!17}59.2 \\
 & KUEA & 68.3 & 90.0 & \cellcolor{blue!6}65.8 & \cellcolor{red!7}82.0 & \cellcolor{blue!6}46.6 & \cellcolor{red!5}88.9 & \cellcolor{red!6}44.8 & \cellcolor{red!6}58.1 & \cellcolor{blue!6}56.2 & \cellcolor{blue!6}51.2 & 48.3 & \cellcolor{blue!6}42.6 & \cellcolor{blue!5}61.9 \\
 & DINOv2 $+$ KL & \cellcolor{blue!6}68.5 & \cellcolor{red!7}89.5 & \cellcolor{blue!7}66.3 & 82.2 & \cellcolor{red!7}45.9 & \cellcolor{blue!10}\textbf{91.0} & \cellcolor{red!11}43.8 & \cellcolor{red!6}58.1 & \cellcolor{blue!7}57.1 & \cellcolor{blue!5}50.9 & \cellcolor{blue!6}\textbf{48.4} & \cellcolor{blue!9}43.1 & \cellcolor{blue!6}62.1 \\
 & SigLIP tower & \cellcolor{blue!6}\textbf{68.6} & \cellcolor{red!6}89.6 & \cellcolor{blue!6}65.9 & \cellcolor{blue!13}\textbf{83.2} & \cellcolor{red!6}46.1 & \cellcolor{blue!7}89.9 & 44.9 & \cellcolor{blue!6}58.3 & \cellcolor{blue!7}56.8 & \cellcolor{red!5}50.5 & \cellcolor{red!7}48.0 & \cellcolor{blue!9}43.1 & \cellcolor{blue!6}62.1 \\
 & Qwen2.5-VL tower & 68.3 & \cellcolor{blue!6}\textbf{90.4} & \cellcolor{blue!10}\textbf{67.1} & \cellcolor{blue!7}82.5 & \cellcolor{blue!11}\textbf{47.8} & \cellcolor{blue!9}90.4 & \cellcolor{blue!6}\textbf{45.1} & \cellcolor{red!9}57.5 & \cellcolor{blue!9}58.1 & \cellcolor{blue!10}53.7 & \cellcolor{red!6}48.1 & \cellcolor{blue!9}\textbf{43.2} & \cellcolor{blue!9}\textbf{62.7} \\
 & \textbf{\method{} (ours)} & \cellcolor{red!7}67.7 & \cellcolor{red!8}89.2 & \cellcolor{blue!8}66.6 & \cellcolor{blue!9}82.7 & \cellcolor{red!11}44.9 & \cellcolor{blue!9}90.3 & \cellcolor{red!11}43.9 & \cellcolor{blue!16}\textbf{60.4} & \cellcolor{blue!10}\textbf{59.0} & \cellcolor{blue!7}51.6 & \cellcolor{red!8}47.8 & \cellcolor{blue!9}43.1 & \cellcolor{blue!7}62.3 \\
\hline
\multirow{8}{*}{\rotatebox[origin=c]{90}{ViT-L/14}}
 & Untrained CLIP & 75.0 & 95.2 & 71.1 & 83.3 & 50.0 & 93.2 & 55.2 & 63.3 & 62.6 & 52.0 & 59.6 & 32.2 & 66.1 \\
 & Shuffled teacher & \cellcolor{red!10}73.7 & \cellcolor{red!7}94.6 & \cellcolor{red!18}67.2 & \cellcolor{blue!10}83.9 & \cellcolor{red!12}48.3 & \cellcolor{red!10}91.5 & 55.2 & \cellcolor{red!13}61.7 & \cellcolor{red!8}60.8 & \cellcolor{blue!8}53.6 & \cellcolor{red!8}59.1 & \cellcolor{blue!5}32.3 & \cellcolor{red!9}65.2 \\
 & SupCon & \cellcolor{red!14}72.9 & \cellcolor{red!15}92.1 & \cellcolor{red!7}70.4 & \cellcolor{blue!8}83.7 & \cellcolor{red!5}49.9 & \cellcolor{red!10}91.5 & \cellcolor{blue!11}\textbf{56.3} & \cellcolor{blue!13}64.8 & \cellcolor{red!6}62.0 & \cellcolor{blue!25}\textbf{63.1} & \cellcolor{red!13}58.1 & \cellcolor{red!8}31.5 & \cellcolor{blue!6}66.4 \\
 & KUEA & \cellcolor{blue!5}75.1 & 95.2 & \cellcolor{blue!6}71.3 & \cellcolor{blue!6}83.4 & \cellcolor{red!6}49.8 & 93.2 & 55.2 & \cellcolor{blue!6}63.4 & \cellcolor{red!6}62.3 & \cellcolor{blue!5}52.1 & 59.6 & \cellcolor{blue!6}32.5 & 66.1 \\
 & DINOv2 $+$ KL & \cellcolor{blue!7}\textbf{75.6} & \cellcolor{red!5}95.1 & \cellcolor{blue!12}73.2 & \cellcolor{blue!17}\textbf{84.8} & \cellcolor{red!11}48.6 & \cellcolor{blue!7}\textbf{94.1} & \cellcolor{blue!10}56.1 & \cellcolor{blue!10}64.2 & \cellcolor{blue!10}65.4 & \cellcolor{blue!9}54.4 & \cellcolor{blue!8}\textbf{60.1} & \cellcolor{blue!8}32.8 & \cellcolor{blue!9}67.0 \\
 & SigLIP tower & \cellcolor{blue!6}75.3 & \cellcolor{blue!6}\textbf{95.4} & \cellcolor{blue!10}72.5 & \cellcolor{blue!11}84.0 & \cellcolor{red!8}49.3 & \cellcolor{blue!6}93.7 & \cellcolor{blue!10}56.0 & \cellcolor{blue!9}64.0 & \cellcolor{blue!6}63.3 & \cellcolor{blue!10}54.9 & \cellcolor{blue!6}59.7 & \cellcolor{blue!9}33.0 & \cellcolor{blue!8}66.8 \\
 & Qwen2.5-VL tower & \cellcolor{blue!5}75.1 & 95.2 & \cellcolor{blue!13}73.4 & \cellcolor{blue!14}84.4 & \cellcolor{blue!9}\textbf{51.0} & \cellcolor{blue!6}93.4 & \cellcolor{blue!8}55.7 & \cellcolor{blue!6}63.4 & \cellcolor{blue!9}65.1 & \cellcolor{red!6}51.6 & \cellcolor{blue!7}59.9 & \cellcolor{blue!11}33.5 & \cellcolor{blue!8}66.8 \\
 & \textbf{\method{} (ours)} & \cellcolor{blue!6}75.2 & \cellcolor{red!6}95.0 & \cellcolor{blue!17}\textbf{74.5} & \cellcolor{blue!16}84.7 & \cellcolor{red!15}47.7 & \cellcolor{blue!7}93.9 & \cellcolor{blue!10}56.1 & \cellcolor{blue!13}\textbf{64.9} & \cellcolor{blue!13}\textbf{67.4} & \cellcolor{blue!9}54.2 & \cellcolor{blue!6}59.8 & \cellcolor{blue!12}\textbf{33.6} & \cellcolor{blue!10}\textbf{67.2} \\
\hline
\multirow{8}{*}{\rotatebox[origin=c]{90}{ViT-L/14-336}}
 & Untrained CLIP & 75.8 & 94.5 & 71.1 & 83.4 & 49.0 & 93.7 & 55.6 & 63.7 & 61.5 & 60.7 & 61.0 & 32.8 & 66.9 \\
 & Shuffled teacher & \cellcolor{red!8}75.0 & \cellcolor{red!9}93.4 & \cellcolor{red!22}66.2 & \cellcolor{blue!7}83.6 & \cellcolor{blue!10}50.2 & \cellcolor{red!10}91.7 & \cellcolor{red!7}55.3 & \cellcolor{red!13}62.1 & \cellcolor{red!12}57.6 & \cellcolor{red!7}59.7 & \cellcolor{red!10}60.1 & \cellcolor{red!12}31.4 & \cellcolor{red!11}65.5 \\
 & SupCon & \cellcolor{red!17}73.0 & \cellcolor{red!18}90.5 & \cellcolor{red!18}67.4 & \cellcolor{blue!9}83.9 & \cellcolor{red!21}45.2 & \cellcolor{red!17}89.2 & 55.6 & \cellcolor{blue!22}\textbf{67.1} & \cellcolor{blue!6}62.2 & \cellcolor{blue!7}61.8 & \cellcolor{red!17}58.8 & \cellcolor{red!14}30.8 & \cellcolor{red!11}65.5 \\
 & KUEA & \cellcolor{blue!5}75.9 & \cellcolor{red!5}94.4 & \cellcolor{blue!6}71.5 & \cellcolor{blue!6}83.5 & \cellcolor{blue!7}49.5 & 93.7 & \cellcolor{red!6}55.4 & \cellcolor{blue!6}63.9 & \cellcolor{red!5}61.3 & \cellcolor{blue!5}60.9 & 61.0 & \cellcolor{blue!6}33.0 & \cellcolor{blue!5}67.0 \\
 & DINOv2 $+$ KL & \cellcolor{blue!8}\textbf{76.4} & \cellcolor{red!6}94.2 & \cellcolor{blue!16}74.2 & \cellcolor{blue!12}\textbf{84.3} & \cellcolor{red!16}46.5 & \cellcolor{blue!6}93.9 & \cellcolor{blue!14}\textbf{57.2} & \cellcolor{blue!15}65.6 & \cellcolor{blue!10}64.2 & \cellcolor{blue!9}\textbf{63.1} & \cellcolor{blue!7}\textbf{61.4} & \cellcolor{red!6}32.5 & \cellcolor{blue!9}67.8 \\
 & SigLIP tower & \cellcolor{blue!6}76.1 & \cellcolor{blue!5}\textbf{94.6} & \cellcolor{blue!10}72.5 & \cellcolor{blue!6}83.5 & \cellcolor{red!5}48.9 & \cellcolor{blue!6}93.9 & \cellcolor{blue!10}56.5 & \cellcolor{blue!6}63.9 & \cellcolor{blue!9}63.9 & 60.7 & 61.0 & \cellcolor{blue!8}\textbf{33.4} & \cellcolor{blue!7}67.4 \\
 & Qwen2.5-VL tower & \cellcolor{blue!6}76.0 & 94.5 & \cellcolor{blue!15}73.9 & \cellcolor{blue!7}83.7 & \cellcolor{blue!14}\textbf{51.1} & 93.7 & \cellcolor{blue!13}57.0 & \cellcolor{blue!8}64.3 & \cellcolor{blue!12}65.3 & \cellcolor{blue!7}61.9 & \cellcolor{blue!7}61.3 & \cellcolor{blue!7}33.3 & \cellcolor{blue!10}\textbf{68.0} \\
 & \textbf{\method{} (ours)} & \cellcolor{blue!6}76.0 & \cellcolor{red!6}94.2 & \cellcolor{blue!18}\textbf{75.0} & \cellcolor{blue!12}\textbf{84.3} & \cellcolor{red!13}47.1 & \cellcolor{blue!6}\textbf{94.0} & \cellcolor{blue!12}56.8 & \cellcolor{blue!21}66.8 & \cellcolor{blue!12}\textbf{65.6} & \cellcolor{red!11}57.5 & 61.0 & \cellcolor{blue!8}\textbf{33.4} & \cellcolor{blue!8}67.6 \\
\hline
\end{tabular}%
}
\end{table}

\textbf{Evaluation.}
Table~\ref{tab:zeroshot} expands the ZS averages in Table~\ref{tab:backbones} for dog-trained models. ImageNet uses 10 images per class across 1{,}000 classes; the other 11 benchmarks use their full test splits. C-10/100 denote CIFAR-10/100, FER denotes FER2013, IN-Sk. denotes ImageNet-Sketch, and IN-O denotes ImageNet-O. The methods retain their respective targets and regularisers from the main comparison. Colour intensity is scaled separately for each benchmark and the mean, using a shared scale across backbones.

\textbf{Retention varies across benchmarks.}
\method{} improves mean zero-shot accuracy on ViT-B/16, ViT-L/14, and ViT-L/14-336, but reduces it on ViT-B/32. The mean hides consistent trade-offs: CIFAR-10 and FER2013 decline on all four backbones, while CIFAR-100, Caltech-101, Pets, RESISC45, EuroSAT, and ImageNet-O improve on the three larger backbones. For ViT-B/16, EuroSAT rises from 55.9 to 59.0 and RESISC45 from 58.2 to 60.4, while CIFAR-10 falls from 90.0 to 89.2. On ViT-L/14-336, PCam falls from 60.7 to 57.5 despite the higher overall mean. Thus, average retention does not imply unchanged accuracy on every benchmark.

\subsection{Retrieval and fidelity on unseen classes}
\label{app:retrieval}

\begin{table}[t]
\caption{\textbf{Retrieval and fidelity on unseen classes.} CLIP ViT-B/16. Retrieval uses 500 unseen-breed and 200 non-dog images; scores are percentages and $\Delta$ is the mAP gain over CLIP. Fidelity is Spearman $\rho$ against Qwen3.8-27B on 250 unseen-breed images. All uses every pair; Within uses pairs of the same breed; Between uses pairs of different breeds. Blue/red indicates improvement/decline from CLIP, scaled per metric column; bold marks the best score within each block.}
\label{tab:retrieval-full}
\centering
\footnotesize
\setlength{\tabcolsep}{2.6pt}
\setlength{\arrayrulewidth}{0.4pt}
\arrayrulecolor{black}
\renewcommand{\arraystretch}{1.15}
\resizebox{\linewidth}{!}{%
\begin{tabular}{|l|l|cccc c|cccc c|ccc|}
\hline
 & & \multicolumn{5}{c|}{In domain: 25 unseen breeds} & \multicolumn{5}{c|}{Out of domain: 50 non-dog classes} & \multicolumn{3}{c|}{\shortstack{Unseen-breed\\fidelity ($\rho$)}} \\
 & Method & R@1 & R@5 & P@5 & mAP & $\Delta$ & R@1 & R@5 & P@5 & mAP & $\Delta$ & All & Within & Between \\
\hline
\multirow{8}{*}{\rotatebox[origin=c]{90}{Target}}
 & Untrained CLIP & 47.0 & 82.2 & 38.8 & 22.1 & --- & 77.0 & 93.5 & 44.1 & 70.0 & --- & 0.268 & 0.231 & 0.245 \\
 & Shuffled teacher & \cellcolor{red!12}36.6 & \cellcolor{red!16}76.0 & \cellcolor{red!11}28.8 & \cellcolor{red!7}18.0 & $-4.1$ & \cellcolor{red!23}73.5 & \cellcolor{red!10}93.0 & \cellcolor{red!12}43.1 & \cellcolor{red!14}67.7 & $-2.3$ & \cellcolor{red!6}0.239 & \cellcolor{red!5}0.227 & \cellcolor{red!6}0.211 \\
 & SupCon & \cellcolor{blue!19}69.4 & \cellcolor{blue!14}87.2 & \cellcolor{blue!18}62.0 & \cellcolor{blue!18}47.7 & $+25.6$ & \cellcolor{blue!13}78.5 & \cellcolor{red!20}92.0 & \cellcolor{red!6}43.9 & \cellcolor{blue!10}71.3 & $+1.3$ & \cellcolor{blue!16}0.546 & \cellcolor{blue!16}0.471 & \cellcolor{blue!16}0.508 \\
 & KUEA & \cellcolor{blue!5}47.4 & 82.2 & \cellcolor{blue!5}38.9 & \cellcolor{blue!5}22.3 & $+0.2$ & \cellcolor{blue!10}78.0 & 93.5 & \cellcolor{red!6}44.0 & \cellcolor{blue!5}70.1 & $+0.1$ & \cellcolor{blue!5}0.271 & \cellcolor{blue!5}0.232 & \cellcolor{blue!5}0.247 \\
 & DINOv2 $+$ KL & \cellcolor{blue!23}75.2 & \cellcolor{blue!22}92.0 & \cellcolor{blue!21}67.5 & \cellcolor{blue!21}52.0 & $+29.9$ & \cellcolor{blue!25}\textbf{81.0} & \cellcolor{blue!20}95.0 & \cellcolor{blue!25}\textbf{47.1} & \cellcolor{blue!25}74.8 & $+4.8$ & \cellcolor{blue!17}0.560 & \cellcolor{blue!19}0.536 & \cellcolor{blue!16}0.523 \\
 & SigLIP tower & \cellcolor{blue!13}59.6 & \cellcolor{blue!12}86.2 & \cellcolor{blue!9}46.3 & \cellcolor{blue!9}29.6 & $+7.5$ & 77.0 & \cellcolor{blue!10}94.0 & \cellcolor{blue!6}44.3 & \cellcolor{blue!6}70.3 & $+0.3$ & \cellcolor{blue!9}0.356 & \cellcolor{blue!8}0.291 & \cellcolor{blue!8}0.326 \\
 & Qwen2.5-VL tower & \cellcolor{blue!13}59.6 & \cellcolor{blue!13}87.0 & \cellcolor{blue!10}47.0 & \cellcolor{blue!9}30.3 & $+8.2$ & \cellcolor{blue!17}79.5 & \cellcolor{blue!15}94.5 & \cellcolor{blue!8}44.5 & \cellcolor{blue!12}71.6 & $+1.5$ & \cellcolor{blue!10}0.401 & \cellcolor{blue!11}0.357 & \cellcolor{blue!10}0.370 \\
 & \textbf{\method{}} & \cellcolor{blue!25}\textbf{78.6} & \cellcolor{blue!25}\textbf{93.6} & \cellcolor{blue!25}\textbf{74.0} & \cellcolor{blue!25}\textbf{60.2} & $+38.1$ & \cellcolor{blue!23}80.5 & \cellcolor{blue!25}\textbf{95.5} & \cellcolor{blue!22}46.7 & \cellcolor{blue!25}\textbf{74.9} & $+4.9$ & \cellcolor{blue!25}\textbf{0.756} & \cellcolor{blue!24}\textbf{0.657} & \cellcolor{blue!25}\textbf{0.734} \\
\hline
\multirow{6}{*}{\rotatebox[origin=c]{90}{\method{} by judge}}
 & Qwen3.5-397B & \cellcolor{blue!25}\textbf{78.2} & \cellcolor{blue!24}\textbf{93.0} & \cellcolor{blue!25}\textbf{73.9} & \cellcolor{blue!25}\textbf{59.7} & $+37.6$ & \cellcolor{blue!13}78.5 & \cellcolor{blue!10}94.0 & \cellcolor{blue!24}\textbf{46.9} & \cellcolor{blue!24}\textbf{74.7} & $+4.7$ & \cellcolor{blue!25}\textbf{0.765} & \cellcolor{blue!25}0.657 & \cellcolor{blue!25}\textbf{0.745} \\
 & GPT-5.6 Sol & \cellcolor{blue!24}77.8 & \cellcolor{blue!21}91.6 & \cellcolor{blue!24}72.6 & \cellcolor{blue!24}58.9 & $+36.9$ & \cellcolor{blue!10}78.0 & \cellcolor{blue!20}95.0 & \cellcolor{blue!19}46.2 & \cellcolor{blue!20}73.7 & $+3.7$ & \cellcolor{blue!23}0.725 & \cellcolor{blue!22}0.610 & \cellcolor{blue!23}0.701 \\
 & Kimi-K3 & \cellcolor{blue!24}77.8 & \cellcolor{blue!23}92.2 & \cellcolor{blue!24}72.3 & \cellcolor{blue!24}59.0 & $+36.9$ & \cellcolor{blue!10}78.0 & \cellcolor{blue!15}94.5 & \cellcolor{blue!16}45.8 & \cellcolor{blue!18}73.2 & $+3.2$ & \cellcolor{blue!25}0.764 & \cellcolor{blue!25}\textbf{0.668} & \cellcolor{blue!25}0.743 \\
 & Claude Sonnet 5 & \cellcolor{blue!23}76.0 & \cellcolor{blue!22}91.8 & \cellcolor{blue!24}72.5 & \cellcolor{blue!24}58.2 & $+36.2$ & \cellcolor{blue!15}\textbf{79.0} & \cellcolor{blue!15}94.5 & \cellcolor{blue!18}46.0 & \cellcolor{blue!18}73.3 & $+3.3$ & \cellcolor{blue!24}0.750 & \cellcolor{blue!25}0.665 & \cellcolor{blue!24}0.729 \\
 & Qwen3.8-27B & \cellcolor{blue!24}77.0 & \cellcolor{blue!23}92.6 & \cellcolor{blue!24}72.6 & \cellcolor{blue!24}58.9 & $+36.9$ & \cellcolor{blue!13}78.5 & \cellcolor{blue!15}94.5 & \cellcolor{blue!14}45.5 & \cellcolor{blue!18}73.1 & $+3.1$ & \cellcolor{blue!24}0.745 & \cellcolor{blue!25}0.660 & \cellcolor{blue!24}0.723 \\
 & Gemma 4 12B & \cellcolor{blue!24}77.0 & \cellcolor{blue!22}91.8 & \cellcolor{blue!23}70.4 & \cellcolor{blue!23}55.9 & $+33.9$ & \cellcolor{blue!7}77.5 & \cellcolor{blue!25}\textbf{95.5} & \cellcolor{blue!13}45.3 & \cellcolor{blue!13}72.0 & $+1.9$ & \cellcolor{blue!22}0.698 & \cellcolor{blue!22}0.602 & \cellcolor{blue!22}0.671 \\
\hline
\end{tabular}%
}
\end{table}

\textbf{Evaluation.}
Table~\ref{tab:retrieval-full} reports cosine retrieval using CLIP ViT-B/16 on 500 images from 25 unseen dog breeds and 200 images from 50 non-dog ImageNet classes. Each image queries all other images in its set, and matching class labels define relevance. R@$k$ is the percentage of queries with at least one relevant image among the first $k$ results; P@5 is the mean fraction of relevant images in the first five. mAP averages query-level average precision over the full ranking. The upper block compares supervision sources, with \method{} using the default 93-group Qwen protocol. The SupCon row uses the baseline configuration described in Appendix~\ref{app:supcon-repair}, and the same final checkpoint as Table~\ref{tab:backbones}. The lower block compares \method{} students trained with different judges under the matched 31-group protocol.

\textbf{Retrieval and transfer.}
In the upper block, \method{} reaches 60.2 mAP on unseen breeds, compared with 52.0 for DINOv2+KL and 47.7 for SupCon. The corresponding P@5 scores are 74.0, 67.5, and 62.0, showing that the gain also appears among the first few neighbours. On non-dog classes, \method{} and DINOv2+KL perform similarly (74.9 and 74.8 mAP). SupCon also improves on untrained CLIP, reaching 71.3 versus 70.0 mAP. Under the matched scoring budget, all six judges improve retrieval over CLIP on both sets, with mAP gains of 33.9--37.6 points on unseen breeds and 1.9--4.7 on non-dog classes. For each unseen-breed gain, none of 10{,}000 paired sign-flip permutations is as extreme as the observed difference. The non-dog gains are not uniformly statistically significant: Gemma 4 12B's 1.9-point gain has $p=0.116$.

\textbf{Fidelity on unseen breeds.}
The three fidelity columns in Table~\ref{tab:retrieval-full} evaluate agreement with a common Qwen3.8-27B judge under the unchanged dog-similarity rubric. A deterministic subset of 10 images per unseen breed is selected before scoring, giving 250 images. All uses all 31,125 unordered, distinct image pairs. Within uses the 1,125 pairs whose images belong to the same breed; Between uses the 30,000 pairs whose images belong to different breeds. Entries are Spearman correlations with raw expected-digit judgments. All saved encoders use the same normalized-cubic link fitted on the original ASK training data; neither weights nor link parameters are updated on these images or judgments. Every student is compared with the same evaluation judge, irrespective of its training judge.

\textbf{Graded similarity beyond the adaptation breeds.}
\method{} reaches 0.756 overall agreement, compared with 0.268 for CLIP and 0.560 for DINOv2+KL. Its within-breed and between-breed correlations are 0.657 and 0.734, compared with 0.536 and 0.523 for DINOv2+KL. Computing the correlation separately within each breed and averaging the 25 values gives 0.652 for \method{}, 0.299 for CLIP, and 0.543 for DINOv2+KL. This additional calculation holds the breed fixed, separating within-breed variation from differences in mean similarity across breeds. These measurements assess agreement with the common evaluation VLM on new breeds.

\subsection{Criterion transfer beyond dogs}
\label{app:criterion-transfer}

\textbf{Holding the criterion fixed across domains.}
Table~\ref{tab:criterion-ood} evaluates the five dog-trained criterion encoders and two controls from Table~\ref{tab:transfer} on the 200 non-dog images used for class retrieval in Table~\ref{tab:retrieval-full}. Here the targets are raw Qwen3.5-397B judgments under the unchanged colour/lighting, pose/view, and background prompts. This measures whether the requested notion of similarity transfers to new image classes; background agreement, for example, need not preserve class identity. Each encoder retains its training-fitted, diagonally normalized cubic similarity, with no adaptation or link fitting on non-dog images; CLIP uses the breed link. Each criterion uses the same 19,900 unordered pairs from 50 classes. These images share neither paths nor identical file contents with the adaptation images.

\begin{table}[htbp]
\caption{\textbf{Criterion transfer beyond dogs.} Spearman fidelity to raw Qwen3.5-397B judgments on 200 non-dog images (50 classes; 19,900 pairs). Dog-trained encoders receive no non-dog adaptation, and each column reuses its dog scoring prompt unchanged. Each model keeps its training-fitted cubic link; CLIP uses the breed link. Blue/red shading denotes improvement/decline relative to CLIP, scaled within the table; bold marks column bests.}
\label{tab:criterion-ood}
\centering
\begingroup
\footnotesize
\setlength{\tabcolsep}{3pt}
\setlength{\arrayrulewidth}{0.4pt}
\arrayrulecolor{black}
\renewcommand{\arraystretch}{1.15}
\resizebox{0.62\linewidth}{!}{%
\begin{tabular}{|l|ccc|}
\hline
Training criterion / model & Colour/lighting & Pose/view & Background \\
\hline
\textit{Untrained CLIP} & \textit{0.343} & \textit{0.464} & \textit{0.535} \\
\hline
Breed & \cellcolor{red!8}0.302 & \cellcolor{red!11}0.390 & \cellcolor{red!10}0.472 \\
Colour/lighting & \cellcolor{blue!12}\textbf{0.428} & \cellcolor{red!6}0.451 & \cellcolor{blue!8}0.576 \\
Pose/view & \cellcolor{red!6}0.332 & \cellcolor{blue!7}\textbf{0.492} & \cellcolor{blue!6}0.552 \\
Coat texture & \cellcolor{red!7}0.320 & \cellcolor{red!9}0.412 & \cellcolor{red!8}0.498 \\
Background & \cellcolor{blue!8}0.387 & \cellcolor{red!9}0.413 & \cellcolor{blue!25}\textbf{0.795} \\
\hline
DINOv2+KL & \cellcolor{red!5}0.342 & \cellcolor{red!8}0.431 & \cellcolor{red!8}0.494 \\
Shuffled (neg.) & \cellcolor{red!7}0.323 & \cellcolor{red!7}0.435 & \cellcolor{red!9}0.489 \\
\hline
\end{tabular}
}
\endgroup
\end{table}

\textbf{Transfer varies across criteria.}
The matching student leads every teacher column, exceeding CLIP, the mismatched students, and both controls. Background fidelity rises from 0.535 to 0.795, colour/lighting from 0.343 to 0.428, and pose/view from 0.464 to 0.492. Transfer is strongest for background and more modest for pose/view. Cosine similarity preserves the matching-model ordering, with corresponding fidelities of 0.795, 0.427, and 0.491. Criterion specificity on dogs therefore coexists with transfer beyond dogs. These measurements assess fidelity to the training judge on new classes; they do not provide independent human or attribute-ground-truth validation.

\subsection{Criterion specificity on car images}
\label{app:car-criteria}

\textbf{Comparing criteria on fixed images.}
Table~\ref{tab:car-criteria} complements the car retrieval examples by evaluating every encoder against every criterion. Qwen3.5-397B scores the same 250 held-out images under all four prompts. These images come from the 25 adaptation classes and are disjoint from the 1,000 training images. Fidelity uses all 31,125 unordered pairs, raw judgments, and each encoder's fixed training-fitted cubic link. Models are evaluated at the final 3,000-step checkpoint. Their scoring budgets differ, so this comparison assesses criterion fidelity rather than the effect of budget. Colour/lighting concerns the whole image, while pose/view includes the camera angle and composition (Appendix~\ref{app:prompts}).

\begin{table}[htbp]
\caption{\textbf{Criterion specificity on cars.} Spearman fidelity to raw Qwen3.5-397B judgments on the same 250 held-out car images (25 adaptation classes; 31,125 pairs). Rows identify car-trained encoders and columns the evaluation criteria. Each model keeps its training-fitted cubic link; CLIP uses the car-model link. Blue/red shading denotes improvement/decline relative to CLIP, scaled within the table; bold marks column bests.}
\label{tab:car-criteria}
\centering
\begingroup
\footnotesize
\setlength{\tabcolsep}{3pt}
\setlength{\arrayrulewidth}{0.4pt}
\arrayrulecolor{black}
\renewcommand{\arraystretch}{1.15}
\resizebox{0.74\linewidth}{!}{%
\begin{tabular}{|l|cccc|}
\hline
Training criterion / model & Car model & Colour/lighting & Pose/view & Background \\
\hline
\textit{Untrained CLIP} & \textit{0.616} & \textit{0.232} & \textit{0.317} & \textit{0.081} \\
\hline
Car model & \cellcolor{blue!12}\textbf{0.859} & \cellcolor{red!5}0.227 & \cellcolor{blue!7}0.384 & \cellcolor{red!6}0.051 \\
Colour/lighting & \cellcolor{red!11}0.409 & \cellcolor{blue!18}\textbf{0.704} & \cellcolor{red!7}0.257 & \cellcolor{blue!9}0.231 \\
Pose/view & \cellcolor{red!8}0.497 & \cellcolor{red!6}0.196 & \cellcolor{blue!17}\textbf{0.756} & \cellcolor{blue!5}0.096 \\
Background & \cellcolor{red!16}0.209 & \cellcolor{blue!7}0.303 & \cellcolor{red!9}0.154 & \cellcolor{blue!25}\textbf{0.813} \\
\hline
DINOv2+KL & \cellcolor{red!5}0.597 & \cellcolor{red!6}0.190 & \cellcolor{red!5}0.310 & \cellcolor{red!6}0.057 \\
\hline
\end{tabular}
}
\endgroup
\end{table}

\textbf{Matching criteria yield the highest fidelity.}
Each matching student leads its teacher column: 0.859 for car model, 0.704 for colour/lighting, 0.756 for pose/view, and 0.813 for background. This ordering also holds under cosine similarity. The car retrieval examples are thus supported by quantitative criterion-specific agreement on held-out images. As with the dog comparisons, agreement with the training judge does not establish independent criterion-retrieval accuracy.

\subsection{Criterion retrieval with human annotations}
\label{app:human-attributes}

\textbf{Human attributes and adaptation.}
We evaluate wing colour, bill shape, and breast pattern using existing image-level human annotations from CUB-200-2011~\citep{wah2011cub}. Three CLIP ViT-B/16 students use the same 992 uniformly sampled training images and 44,702 Qwen3.8-27B judgments per criterion, with the kernel objective, PSD processing, and KL anchor described in the method. Prompts compare the visible attribute while ignoring species and unrelated appearance. Each request contains two training images and the corresponding criterion prompt; Appendix~\ref{app:cub-prompts} gives the complete prompts. Human attribute labels do not enter student training. DINOv2+KL uses the same images with criterion-independent feature targets.

\textbf{Cosine retrieval across species.}
Evaluation uses the 5,794 official test images. For criterion $c$, $A_c(x)$ contains attributes annotated as present with certainty at least 3; images with empty sets are excluded. Each query retrieves eligible images of other species, giving 4,985, 5,266, and 5,001 queries for wing colour, bill shape, and breast pattern. Relevance is the Jaccard overlap $r_c(x,y)=|A_c(x)\cap A_c(y)|/|A_c(x)\cup A_c(y)|$. We report nDCG@10 with linear relevance gain and logarithmic rank discount, normalized by each query's ideal ranking. All methods use cosine similarity and identical eligible galleries; neither a fitted kernel link nor VLM judgments enter evaluation.

\textbf{Baseline representations.}
Alongside CLIP~\citep{radford2021learning} and DINOv2+KL~\citep{oquab2024dinov2}, two training-free controls use CLIP's original text encoder to describe attribute values: \emph{text subspace} projects image embeddings onto the span of centred text vectors, and \emph{text profile} uses image--text similarities as coordinates. Neither fits image labels. VLM2Vec~\citep{jiang2025vlm2vec} (Qwen2-VL-2B) embeds images with task instructions; MagicLens~\citep{zhang2024magiclens} (CLIP-B/16) learns image-and-instruction embeddings from web pairs and synthesized instructions. AligNet~\citep{muttenthaler2025aligning} learns criterion-independent representations from a human-aligned teacher's synthetic targets; we use its released CLIP-B's 512-dimensional image-encoder output. Each released baseline retains its native preprocessing and receives no weight adaptation on our evaluation datasets.

\textbf{Selecting instruction settings.}
Following the initial baseline evaluation, we select CUB instruction settings by mean attribute nDCG@10 on 512 unused training images, disjoint from adaptation and test images. We compare detailed, short, and positive prompts, gallery conditioning, and VLM2Vec resolution. The selected native-resolution prompts are ``Represent the [attribute] of the bird in this image'' for VLM2Vec and ``Find a bird with the same [attribute]'' for MagicLens, applied to both query and gallery. Human validation labels guide selection; revised test scores do not.

\begin{table}[htbp]
\caption{\textbf{Criterion retrieval and baseline checks on CUB-200-2011.} Cosine nDCG@10 ($\times100$) with cross-species galleries. \textbf{(a)} Model comparison; each \method{} row is one fixed criterion-trained student. \textbf{(b)} Instruction settings and AligNet output spaces. Symmetric conditioning uses the same instruction for query and gallery; \textsuperscript{*}marks validation-selected settings used in (a), where AligNet uses its image encoder. Blue/red denotes improvement/decline from CLIP in (a) or each block\textquotesingle s first row in (b), scaled per column; bold marks column bests in (a) and block bests in (b).}
\label{tab:human-attributes}
\centering
\begingroup
\footnotesize
\setlength{\tabcolsep}{2pt}
\setlength{\arrayrulewidth}{0.4pt}
\arrayrulecolor{black}
\renewcommand{\arraystretch}{1.10}
\begin{minipage}[t]{0.49\linewidth}
\vspace{0pt}
\resizebox{\linewidth}{!}{%
\begin{tabular}{|l|ccc|}
\hline
\multicolumn{4}{|c|}{\textbf{(a) Attribute retrieval}} \\
\hline
Model & \shortstack{Wing\\colour} & \shortstack{Bill\\shape} & \shortstack{Breast\\pattern} \\
\hline
\textit{Untrained CLIP} & 34.55 & 49.98 & 52.31 \\
\hline
DINOv2+KL & \cellcolor{blue!10}35.62 & \cellcolor{blue!6}50.28 & \cellcolor{blue!6}52.72 \\
CLIP text subspace & \cellcolor{blue!9}35.42 & \cellcolor{red!24}38.94 & \cellcolor{red!25}45.97 \\
CLIP text profile & \cellcolor{red!6}34.44 & \cellcolor{red!25}38.25 & \cellcolor{red!25}45.91 \\
\hline
VLM2Vec & \cellcolor{blue!7}34.87 & \cellcolor{blue!5}50.09 & \cellcolor{red!6}52.10 \\
MagicLens & \cellcolor{red!11}33.23 & \cellcolor{red!11}46.70 & \cellcolor{red!9}51.16 \\
\hline
AligNet: base CLIP-B & \cellcolor{red!18}31.91 & \cellcolor{red!10}46.86 & \cellcolor{red!8}51.25 \\
AligNet: aligned CLIP-B & \cellcolor{red!6}34.39 & \cellcolor{red!10}46.88 & \cellcolor{red!6}51.99 \\
\hline
\method{}: wing colour & \cellcolor{blue!25}\textbf{38.64} & \cellcolor{red!10}47.26 & \cellcolor{red!6}51.90 \\
\method{}: bill shape & \cellcolor{blue!7}35.02 & \cellcolor{blue!8}\textbf{52.01} & \cellcolor{blue!6}52.56 \\
\method{}: breast pattern & \cellcolor{blue!6}34.77 & \cellcolor{red!9}47.89 & \cellcolor{blue!15}\textbf{55.66} \\
\hline
\end{tabular}%
}
\end{minipage}\hfill%
\begin{minipage}[t]{0.49\linewidth}
\vspace{0pt}
\resizebox{\linewidth}{!}{%
\begin{tabular}{|l|ccc|}
\hline
\multicolumn{4}{|c|}{\textbf{(b) Baseline checks}} \\
\hline
Configuration & \shortstack{Wing\\colour} & \shortstack{Bill\\shape} & \shortstack{Breast\\pattern} \\
\hline
\multicolumn{4}{|l|}{\textit{VLM2Vec}} \\
\hline
Detailed, symmetric & 33.79 & 45.96 & 50.86 \\
Positive, symmetric\textsuperscript{*} & \cellcolor{blue!25}\textbf{34.87} & \cellcolor{blue!25}\textbf{50.09} & \cellcolor{blue!25}\textbf{52.10} \\
\hline
\multicolumn{4}{|l|}{\textit{MagicLens}} \\
\hline
Detailed, query only & 17.58 & 34.44 & 45.45 \\
Short, symmetric\textsuperscript{*} & \cellcolor{blue!23}33.23 & \cellcolor{blue!25}\textbf{46.70} & \cellcolor{blue!25}51.16 \\
Empty text on both sides & \cellcolor{blue!25}\textbf{34.87} & \cellcolor{blue!25}46.64 & \cellcolor{blue!25}\textbf{51.27} \\
\hline
\multicolumn{4}{|l|}{\textit{AligNet CLIP-B}} \\
\hline
Base: image encoder & 31.91 & 46.86 & 51.25 \\
Aligned: image encoder & \cellcolor{blue!25}\textbf{34.39} & \cellcolor{blue!6}\textbf{46.88} & \cellcolor{blue!25}\textbf{51.99} \\
Base: triplet head & \cellcolor{red!6}31.80 & \cellcolor{red!25}46.33 & \cellcolor{red!6}51.22 \\
Aligned: triplet head & \cellcolor{blue!24}34.31 & 46.86 & \cellcolor{blue!25}51.98 \\
\hline
\end{tabular}%
}
\end{minipage}
\endgroup
\end{table}

\textbf{Criterion accuracy.}
In Table~\ref{tab:human-attributes}(a), the matching \method{} student leads each column and exceeds both mismatched students. Gains over CLIP are 4.09, 2.03, and 3.35 points. Gains over VLM2Vec are 3.78, 1.92, and 3.56, with paired 95\% intervals [2.92, 4.70], [0.90, 2.91], and [2.43, 4.67], resampling query-species clusters with fixed checkpoints and galleries. \method{} and DINOv2+KL use CUB adaptation, whereas the other learned baselines use released weights. Panel~(b) compares instruction settings and AligNet output spaces; MagicLens\textquotesingle s instruction-free embeddings remain competitive with its selected configuration.

\begin{figure}[htbp]
\centering
\includegraphics[width=\linewidth]{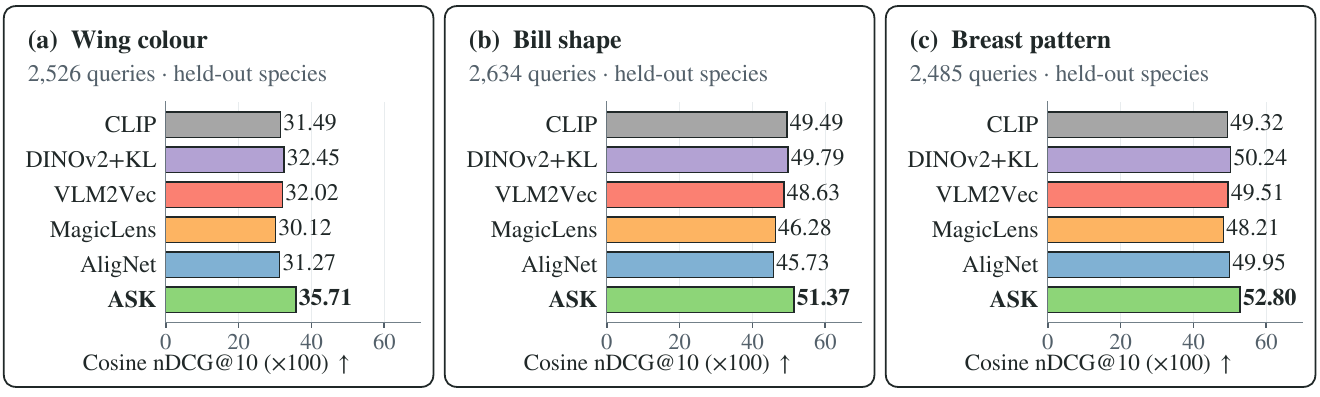}
\caption{\textbf{Attribute retrieval on species excluded from adaptation.} \method{} and DINOv2+KL adapt on 100 CUB species and evaluate on the other 100, with cross-species galleries. Each bar reports cosine nDCG@10 for one fixed encoder; \method{} uses the matching criterion. VLM2Vec and MagicLens retain the earlier CUB settings, selected on validation images spanning all 200 species; their prompt selection is not species-disjoint.}
\label{fig:cub-species-transfer}
\end{figure}

\textbf{Transfer to species excluded from adaptation.}
We repeat CUB adaptation using 992 images from 100 species, reserving the other 100 species for evaluation. Fresh \method{} and DINOv2+KL adapters start from original CLIP. The 2,903 held-out-species test images yield 2,526 colour, 2,634 shape, and 2,485 pattern queries under the same relevance and cross-species gallery rules. Figure~\ref{fig:cub-species-transfer} shows gains of 3.26, 1.58, and 2.55 points over DINOv2+KL; matching students also exceed both mismatched students, with all paired species-bootstrap intervals above zero. Species separation concerns adaptation, not foundation-model pretraining.

\begin{figure}[htbp]
\centering
\includegraphics[width=\linewidth]{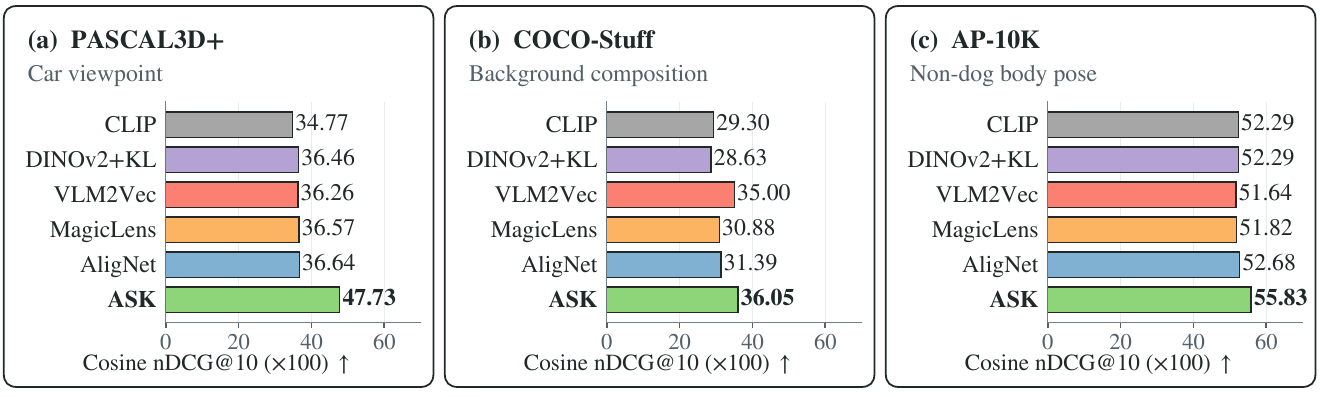}
\caption{\textbf{Criterion transfer evaluated with independent annotations.} \textbf{(a)} Car viewpoint from PASCAL3D+ azimuth annotations. \textbf{(b)} Background composition from COCO-Stuff masks. \textbf{(c)} Non-dog body pose from AP-10K keypoints. Each bar reports cosine nDCG@10 for one fixed encoder, with no weight updates on these datasets. VLM2Vec and MagicLens use separate validation images to select instructions. \method{} uses the car Pose/view, dog background, and dog Pose/view encoders, respectively; DINOv2+KL uses the corresponding adaptation domain.}
\label{fig:human-criterion-transfer}
\end{figure}

\textbf{Viewpoint, background, and pose on new datasets.}
Figure~\ref{fig:human-criterion-transfer} evaluates existing car Pose/view and dog background and Pose/view encoders without updating their weights. PASCAL3D+~\citep{xiang2014beyond} supplies 86 clear car images from the PASCAL validation split; within-category relevance is $\max(0,1-d_{\rm az}/90^\circ)$, where $d_{\rm az}$ is circular azimuth error. COCO-Stuff~\citep{caesar2018cocostuff} supplies 3,701 non-dog validation images with at least 20\% stuff and 5\% foreground coverage in the CLIP crop. Relevance is Jaccard overlap of stuff classes covering at least 1\% of the crop; galleries exclude the query's dominant foreground class. AP-10K~\citep{yu2021ap10k} supplies 517 non-dog pose queries from 33 species in test split 1. We use single-animal images with at least ten visible keypoints, subtract the bounding-box centre, and divide coordinates by the longer box side. Within-species galleries control species identity; relevance is $\exp[-d^2/(2\cdot0.1^2)]$, where $d^2$ is mean squared distance over at least eight jointly visible keypoints. These annotations define retrieval relevance, not direct human similarity votes; stuff overlap measures background composition.

\textbf{Transfer comparisons.}
For each new dataset, VLM2Vec and MagicLens receive the same search budget of five prompt families and two gallery-conditioning formats, using 274 PASCAL, 512 COCO, and 768 AP-10K training images disjoint from evaluation and \method{} adaptation. Selection averages nDCG@10 across vehicle categories for PASCAL, uses background retrieval for COCO, and averages non-dog pose and background for AP-10K. All models then use fixed weights and cosine retrieval. \method{} exceeds the strongest plotted baseline by 11.09, 1.05, and 3.16 points on car viewpoint, background, and non-dog pose. Transfer is not uniform: on PASCAL buses and motorbikes, \method{} scores 57.14 and 33.37 versus VLM2Vec's 57.85 and 33.50.

\subsection{Per-benchmark retention across adaptation domains}
\label{app:domain-zeroshot}

\begin{table}[t]
\caption{\textbf{Per-benchmark accuracy after domain adaptation.} CLIP ViT-B/16; Mean averages the 12 benchmarks. Blue/red indicates improvement/decline from untrained CLIP, scaled per column across domains. Bold marks the highest adapted score within each domain when it exceeds CLIP. Same-domain evaluations are discussed in the text.}
\label{tab:domains-zs}
\centering
\scriptsize
\setlength{\tabcolsep}{1.6pt}
\setlength{\arrayrulewidth}{0.4pt}
\arrayrulecolor{black}
\renewcommand{\arraystretch}{1.15}
\resizebox{0.9\linewidth}{!}{%
\begin{tabular}{|l|l|cccccccccccc|c|}
\hline
 & Method & IN-1k & C-10 & C-100 & Caltech & FER & Pets & DTD & RESISC & EuroSAT & PCam & IN-Sk. & IN-O & Mean \\
\hline
 & Untrained CLIP & 68.3 & 90.0 & 65.6 & 82.2 & 46.3 & 89.0 & 44.9 & 58.2 & 55.9 & 50.7 & 48.3 & 42.3 & 61.8 \\
\hline
\multirow{5}{*}{\rotatebox[origin=c]{90}{Dogs}}
 & SupCon & \cellcolor{red!25}63.7 & \cellcolor{red!22}87.5 & \cellcolor{red!15}62.2 & \cellcolor{red!25}79.7 & \cellcolor{red!21}41.9 & \cellcolor{red!25}84.8 & \cellcolor{red!23}43.3 & \cellcolor{red!7}57.8 & \cellcolor{red!13}53.2 & \cellcolor{blue!14}53.0 & \cellcolor{red!20}44.6 & \cellcolor{red!16}39.0 & \cellcolor{red!21}59.2 \\
 & KUEA & 68.3 & 90.0 & \cellcolor{blue!6}65.8 & \cellcolor{red!7}82.0 & \cellcolor{blue!6}46.6 & \cellcolor{red!5}88.9 & \cellcolor{red!6}44.8 & \cellcolor{red!5}58.1 & \cellcolor{blue!6}56.2 & \cellcolor{blue!7}51.2 & 48.3 & \cellcolor{blue!6}42.6 & \cellcolor{blue!6}61.9 \\
 & Qwen2.5-VL tower & 68.3 & \cellcolor{blue!8}\textbf{90.4} & \cellcolor{blue!9}\textbf{67.1} & \cellcolor{blue!7}82.5 & \cellcolor{blue!10}\textbf{47.8} & \cellcolor{blue!12}90.4 & \cellcolor{blue!7}\textbf{45.1} & \cellcolor{red!8}57.5 & \cellcolor{blue!12}58.1 & \cellcolor{blue!17}\textbf{53.7} & \cellcolor{red!6}48.1 & \cellcolor{blue!8}\textbf{43.2} & \cellcolor{blue!11}\textbf{62.7} \\
 & DINOv2 $+$ KL & \cellcolor{blue!6}\textbf{68.5} & \cellcolor{red!8}89.5 & \cellcolor{blue!7}66.3 & 82.2 & \cellcolor{red!6}45.9 & \cellcolor{blue!15}\textbf{91.0} & \cellcolor{red!17}43.8 & \cellcolor{red!5}58.1 & \cellcolor{blue!9}57.1 & \cellcolor{blue!6}50.9 & \cellcolor{blue!5}\textbf{48.4} & \cellcolor{blue!8}43.1 & \cellcolor{blue!7}62.1 \\
 & \textbf{\method{}} & \cellcolor{red!8}67.7 & \cellcolor{red!10}89.2 & \cellcolor{blue!8}66.6 & \cellcolor{blue!9}\textbf{82.7} & \cellcolor{red!10}44.9 & \cellcolor{blue!11}90.3 & \cellcolor{red!16}43.9 & \cellcolor{blue!15}\textbf{60.4} & \cellcolor{blue!15}\textbf{59.0} & \cellcolor{blue!9}51.6 & \cellcolor{red!7}47.8 & \cellcolor{blue!8}43.1 & \cellcolor{blue!8}62.3 \\
\hline
\multirow{5}{*}{\rotatebox[origin=c]{90}{Textures}}
 & SupCon & \cellcolor{red!24}63.9 & \cellcolor{red!25}87.0 & \cellcolor{red!25}58.6 & \cellcolor{red!20}80.3 & \cellcolor{blue!8}47.2 & \cellcolor{red!17}86.4 & \cellcolor{blue!25}\textbf{46.7} & \cellcolor{red!25}53.6 & \cellcolor{red!25}49.5 & \cellcolor{red!5}50.6 & \cellcolor{red!25}43.3 & \cellcolor{red!25}36.5 & \cellcolor{red!25}58.6 \\
 & KUEA & 68.3 & 90.0 & \cellcolor{red!5}65.5 & 82.2 & \cellcolor{blue!6}46.5 & \cellcolor{red!5}88.9 & \cellcolor{blue!9}45.3 & \cellcolor{blue!6}58.4 & \cellcolor{blue!6}56.1 & \cellcolor{red!7}50.1 & \cellcolor{red!5}48.2 & \cellcolor{blue!6}42.6 & \cellcolor{blue!6}61.9 \\
 & Qwen2.5-VL tower & \cellcolor{blue!6}68.5 & \cellcolor{blue!9}\textbf{90.6} & \cellcolor{blue!7}66.3 & \cellcolor{red!7}82.0 & \cellcolor{blue!11}47.9 & \cellcolor{red!5}88.9 & \cellcolor{blue!11}45.4 & \cellcolor{red!5}58.1 & \cellcolor{red!7}55.3 & \cellcolor{blue!11}52.2 & \cellcolor{red!6}48.0 & \cellcolor{blue!9}43.5 & \cellcolor{blue!8}62.2 \\
 & DINOv2 $+$ KL & \cellcolor{blue!5}68.4 & \cellcolor{blue!7}90.3 & \cellcolor{blue!6}65.8 & \cellcolor{red!6}82.1 & \cellcolor{blue!10}47.6 & \cellcolor{red!6}88.8 & \cellcolor{blue!6}45.0 & \cellcolor{red!7}57.8 & \cellcolor{red!6}55.6 & \cellcolor{blue!10}52.0 & \cellcolor{red!7}47.9 & \cellcolor{blue!9}43.6 & \cellcolor{blue!7}62.1 \\
 & \textbf{\method{}} & \cellcolor{blue!6}\textbf{68.6} & \cellcolor{blue!8}90.4 & \cellcolor{blue!10}\textbf{67.2} & \cellcolor{red!6}82.1 & \cellcolor{blue!19}\textbf{50.1} & \cellcolor{blue!7}\textbf{89.4} & \cellcolor{blue!18}46.1 & \cellcolor{blue!8}\textbf{59.0} & \cellcolor{blue!10}\textbf{57.5} & \cellcolor{blue!13}\textbf{52.6} & \cellcolor{red!5}48.2 & \cellcolor{blue!12}\textbf{44.2} & \cellcolor{blue!12}\textbf{62.9} \\
\hline
\multirow{5}{*}{\rotatebox[origin=c]{90}{Flowers}}
 & SupCon & \cellcolor{red!7}67.8 & \cellcolor{red!11}89.1 & \cellcolor{red!8}64.4 & \cellcolor{blue!8}\textbf{82.6} & \cellcolor{red!9}45.3 & \cellcolor{blue!5}89.1 & \cellcolor{red!11}44.4 & \cellcolor{red!13}56.4 & \cellcolor{red!17}52.1 & \cellcolor{red!18}47.5 & \cellcolor{red!7}47.9 & \cellcolor{blue!5}42.4 & \cellcolor{red!12}60.7 \\
 & KUEA & 68.3 & \cellcolor{red!6}89.9 & \cellcolor{red!6}65.4 & 82.2 & \cellcolor{blue!6}\textbf{46.5} & \cellcolor{red!6}88.8 & 44.9 & 58.2 & \cellcolor{blue!6}\textbf{56.3} & \cellcolor{blue!6}\textbf{51.0} & 48.3 & \cellcolor{red!5}42.2 & 61.8 \\
 & Qwen2.5-VL tower & \cellcolor{blue!5}68.4 & \cellcolor{blue!6}90.2 & \cellcolor{blue!6}65.8 & \cellcolor{red!7}81.9 & \cellcolor{red!6}46.1 & 89.0 & \cellcolor{red!6}44.8 & \cellcolor{blue!8}58.9 & \cellcolor{red!6}55.6 & \cellcolor{red!14}48.5 & \cellcolor{blue!5}48.4 & \cellcolor{blue!8}43.2 & \cellcolor{red!6}61.7 \\
 & DINOv2 $+$ KL & \cellcolor{blue!6}68.5 & 90.0 & \cellcolor{red!6}65.1 & \cellcolor{red!6}82.1 & \cellcolor{red!9}45.1 & \cellcolor{red!6}88.8 & \cellcolor{red!14}44.1 & \cellcolor{blue!9}59.2 & 55.9 & \cellcolor{red!13}48.8 & \cellcolor{blue!5}48.4 & \cellcolor{blue!9}\textbf{43.5} & \cellcolor{red!6}61.6 \\
 & \textbf{\method{}} & \cellcolor{blue!8}\textbf{68.9} & \cellcolor{blue!8}\textbf{90.4} & \cellcolor{blue!8}\textbf{66.5} & \cellcolor{blue!7}82.4 & \cellcolor{red!6}46.1 & \cellcolor{blue!8}\textbf{89.7} & \cellcolor{red!12}44.3 & \cellcolor{blue!10}\textbf{59.3} & \cellcolor{red!5}55.8 & \cellcolor{red!8}50.0 & \cellcolor{blue!7}\textbf{48.7} & \cellcolor{blue!9}\textbf{43.5} & \cellcolor{blue!7}\textbf{62.1} \\
\hline
\multirow{5}{*}{\rotatebox[origin=c]{90}{Cars}}
 & SupCon & \cellcolor{red!7}67.8 & \cellcolor{red!8}89.5 & \cellcolor{blue!6}66.0 & \cellcolor{red!19}80.4 & \cellcolor{blue!25}\textbf{51.8} & \cellcolor{red!10}88.0 & \cellcolor{red!8}44.6 & \cellcolor{blue!8}58.8 & \cellcolor{red!17}52.1 & \cellcolor{red!22}46.5 & \cellcolor{red!9}47.4 & 42.3 & \cellcolor{red!8}61.3 \\
 & KUEA & 68.3 & 90.0 & 65.6 & \cellcolor{red!6}82.1 & \cellcolor{blue!5}46.4 & 89.0 & \cellcolor{blue!7}45.1 & 58.2 & \cellcolor{blue!7}\textbf{56.4} & \cellcolor{blue!6}51.0 & 48.3 & \cellcolor{red!5}42.2 & \cellcolor{blue!6}61.9 \\
 & Qwen2.5-VL tower & \cellcolor{red!5}68.2 & \cellcolor{blue!8}90.5 & \cellcolor{blue!6}65.9 & \cellcolor{red!7}82.0 & \cellcolor{red!9}45.3 & 89.0 & 44.9 & 58.2 & \cellcolor{red!12}53.6 & \cellcolor{blue!7}\textbf{51.3} & \cellcolor{red!5}48.2 & \cellcolor{blue!7}\textbf{42.9} & \cellcolor{red!6}61.7 \\
 & DINOv2 $+$ KL & \cellcolor{blue!5}\textbf{68.4} & \cellcolor{blue!11}\textbf{90.9} & \cellcolor{blue!8}66.7 & \cellcolor{blue!7}\textbf{82.4} & \cellcolor{red!6}46.0 & \cellcolor{red!5}88.9 & \cellcolor{blue!7}45.1 & \cellcolor{blue!10}59.4 & \cellcolor{red!11}54.0 & \cellcolor{red!8}50.0 & \cellcolor{blue!6}\textbf{48.6} & \cellcolor{blue!5}42.4 & \cellcolor{blue!6}61.9 \\
 & \textbf{\method{}} & \cellcolor{blue!5}\textbf{68.4} & \cellcolor{blue!9}90.6 & \cellcolor{blue!10}\textbf{67.4} & \cellcolor{red!11}81.5 & \cellcolor{blue!14}48.8 & \cellcolor{red!6}88.7 & \cellcolor{blue!8}\textbf{45.2} & \cellcolor{blue!12}\textbf{59.7} & \cellcolor{red!15}52.8 & \cellcolor{red!7}50.3 & \cellcolor{blue!6}\textbf{48.6} & \cellcolor{blue!7}\textbf{42.9} & \cellcolor{blue!7}\textbf{62.1} \\
\hline
\multirow{5}{*}{\rotatebox[origin=c]{90}{Pets}}
 & SupCon & \cellcolor{red!22}64.4 & \cellcolor{red!11}89.1 & \cellcolor{red!12}63.3 & \cellcolor{red!17}80.7 & \cellcolor{blue!11}47.9 & \cellcolor{red!13}87.4 & \cellcolor{red!14}44.1 & \cellcolor{red!6}58.0 & \cellcolor{red!19}51.3 & \cellcolor{blue!25}\textbf{55.7} & \cellcolor{red!18}45.1 & \cellcolor{red!10}40.9 & \cellcolor{red!12}60.7 \\
 & KUEA & 68.3 & \cellcolor{blue!6}90.1 & 65.6 & 82.2 & \cellcolor{blue!6}46.6 & 89.0 & \cellcolor{red!7}44.7 & \cellcolor{blue!5}58.3 & \cellcolor{blue!6}\textbf{56.1} & \cellcolor{blue!5}50.8 & \cellcolor{red!5}48.2 & \cellcolor{blue!6}42.6 & \cellcolor{blue!6}61.9 \\
 & Qwen2.5-VL tower & \cellcolor{red!6}68.1 & \cellcolor{blue!8}\textbf{90.5} & \cellcolor{blue!10}\textbf{67.3} & \cellcolor{red!7}81.9 & \cellcolor{blue!5}46.4 & \cellcolor{blue!10}90.1 & \cellcolor{blue!14}\textbf{45.7} & \cellcolor{red!10}57.1 & \cellcolor{red!10}54.3 & \cellcolor{blue!19}54.2 & 48.3 & \cellcolor{blue!6}42.7 & \cellcolor{blue!8}62.2 \\
 & DINOv2 $+$ KL & \cellcolor{red!5}68.2 & \cellcolor{red!8}89.5 & \cellcolor{red!6}65.4 & \cellcolor{red!6}82.1 & \cellcolor{red!9}45.2 & \cellcolor{blue!18}\textbf{91.7} & \cellcolor{red!8}44.6 & \cellcolor{red!5}58.1 & \cellcolor{red!7}55.3 & \cellcolor{blue!24}55.5 & \cellcolor{red!6}48.1 & \cellcolor{red!5}42.2 & \cellcolor{blue!8}62.2 \\
 & \textbf{\method{}} & \cellcolor{blue!5}\textbf{68.4} & \cellcolor{blue!6}90.1 & \cellcolor{blue!9}66.9 & \cellcolor{blue!12}\textbf{83.1} & \cellcolor{blue!12}\textbf{48.3} & \cellcolor{blue!9}89.9 & \cellcolor{blue!9}45.3 & \cellcolor{blue!13}\textbf{60.0} & \cellcolor{red!6}55.5 & \cellcolor{blue!11}52.3 & \cellcolor{red!5}48.2 & \cellcolor{blue!7}\textbf{42.8} & \cellcolor{blue!10}\textbf{62.6} \\
\hline
\end{tabular}%
}
\end{table}

\textbf{Evaluation scope.}
Table~\ref{tab:domains-zs} breaks down the 12-benchmark means in Table~\ref{tab:main}, using the evaluation protocol of Table~\ref{tab:zeroshot}. Each block changes the adaptation domain; the untrained ViT-B/16 baseline is shared. These results measure retention after adaptation. In particular, Pets evaluation after Pets adaptation and DTD evaluation after Textures adaptation use the same source datasets, so these entries should not be interpreted as transfer to an unseen domain. Dog adaptation also uses ImageNet classes; the ImageNet column measures retention on that benchmark.

\textbf{Domain-dependent changes.}
\method{} improves the 12-benchmark mean over CLIP in all five domains, but the gains are not uniform across benchmarks. Texture adaptation raises FER2013 accuracy from 46.3 to 50.1 and ImageNet-O from 42.3 to 44.2. Car adaptation raises CIFAR-100 from 65.6 to 67.4 and RESISC45 from 58.2 to 59.7, while lowering EuroSAT from 55.9 to 52.8. Pet adaptation improves Caltech-101 from 82.2 to 83.1 and RESISC45 to 60.0, but slightly lowers EuroSAT and ImageNet-Sketch. These variations show why the benchmark-level results complement the aggregate retention score.

\subsection{Judge agreement and order consistency}
\label{app:judge-agreement}
\label{app:crossteacher}

\begin{table}[t]
\caption{\textbf{Agreement and order consistency of VLM judges.} Pairwise agreement is Spearman correlation. Blue intensity encodes absolute agreement on a shared scale; bold marks the highest cross-judge value per column. Self-agreement uses swapped image order; Ratio is student fidelity divided by self-Spearman agreement.}
\label{tab:agreement}
\centering
\footnotesize
\setlength{\tabcolsep}{2.4pt}
\setlength{\arrayrulewidth}{0.4pt}
\arrayrulecolor{black}
\renewcommand{\arraystretch}{1.15}
\resizebox{0.9\linewidth}{!}{%
\begin{tabular}{|l|cccccc|ccc|cc|}
\hline
\multirow{2}{*}{Judge} & \multicolumn{6}{c|}{Agreement with judge} & \multicolumn{3}{c|}{Self, swapped order} & \multicolumn{2}{c|}{Own student} \\
\cline{2-12}
 & Q3.5-397B & GPT & Kimi & Sonnet & Q3.8-27B & G4-12B & $\rho_{\mathrm{S}}$ & $\rho_{\mathrm{P}}$ & $|\Delta|$ & Fid & Ratio \\
\hline
Qwen3.5-397B & \textcolor{black!40}{---} & \cellcolor{blue!20}0.753 & \cellcolor{blue!21}0.799 & \cellcolor{blue!19}0.706 & \cellcolor{blue!21}0.819 & \cellcolor{blue!19}0.702 & 0.837 & 0.921 & 0.35 & 0.791 & 94\% \\
GPT-5.6 Sol & \cellcolor{blue!20}0.753 & \textcolor{black!40}{---} & \cellcolor{blue!21}0.811 & \cellcolor{blue!20}0.725 & \cellcolor{blue!20}0.770 & \cellcolor{blue!19}0.679 & 0.901 & 0.926 & 0.32 & 0.794 & 88\% \\
Kimi-K3 & \cellcolor{blue!21}0.799 & \cellcolor{blue!21}\textbf{0.811} & \textcolor{black!40}{---} & \cellcolor{blue!20}\textbf{0.750} & \cellcolor{blue!22}\textbf{0.829} & \cellcolor{blue!19}0.719 & \textbf{0.948} & \textbf{0.975} & \textbf{0.20} & \textbf{0.838} & 88\% \\
Claude Sonnet 5 & \cellcolor{blue!19}0.706 & \cellcolor{blue!20}0.725 & \cellcolor{blue!20}0.750 & \textcolor{black!40}{---} & \cellcolor{blue!19}0.690 & \cellcolor{blue!17}0.605 & 0.820 & 0.912 & 0.33 & 0.778 & \textbf{95\%} \\
Qwen3.8-27B & \cellcolor{blue!21}\textbf{0.819} & \cellcolor{blue!20}0.770 & \cellcolor{blue!22}\textbf{0.829} & \cellcolor{blue!19}0.690 & \textcolor{black!40}{---} & \cellcolor{blue!20}\textbf{0.738} & 0.829 & 0.858 & 0.87 & 0.760 & 92\% \\
Gemma 4 12B & \cellcolor{blue!19}0.702 & \cellcolor{blue!19}0.679 & \cellcolor{blue!19}0.719 & \cellcolor{blue!17}0.605 & \cellcolor{blue!20}0.738 & \textcolor{black!40}{---} & 0.811 & 0.834 & 0.57 & 0.770 & 95\% \\
\hline
\end{tabular}%
}
\end{table}

\textbf{Evaluation.}
Table~\ref{tab:agreement} compares the six judges on the same 100 held-out dog images (4{,}950 pairs). Cross-judge agreement is Spearman correlation between their scores. To assess sensitivity to image order, 1{,}000 pairs are re-scored with the two images swapped. We report Spearman and Pearson self-agreement and the mean absolute score change $|\Delta|$ on the 1--9 scale. The final columns give each student's fidelity to its own judge and its ratio to that judge's self-Spearman agreement. The ratio is descriptive: order consistency is not a strict upper bound on student fidelity, and the two correlations use different pair sets.

\textbf{Shared structure and variability.}
\looseness=-1
Cross-judge correlations range from 0.605 for Sonnet--Gemma 4 12B to 0.829 for Kimi--Qwen3.8-27B, indicating shared but non-identical similarity judgments. Self-Spearman agreement ranges from 0.811 to 0.948. Kimi has the highest order consistency and the smallest mean absolute score change (0.20), whereas Qwen3.8-27B changes by 0.87 on average. Student fidelity is 88--95\% of the corresponding self-agreement reference. These measurements help contextualize fidelity differences across judges without treating any judge as noise-free ground truth.

\subsection{Judge comparisons across backbones}
\label{app:judge-scale}
\label{app:judge-scale-zeroshot}

\begin{table}[t]
\caption{\textbf{VLM judges across CLIP backbones.} All students use 31 scoring groups. Spearman shading compares $\rho_{\mathrm{S}}$ with Base $\rho_{\mathrm{S}}$ in the same judge row. These judge-specific CLIP baselines replace a single value in the Untrained CLIP row. Retrieval and zero-shot shading compare with that backbone's Untrained CLIP row. Blue/red denotes improvement/decline, scaled per metric; Pearson is unshaded. Bold marks column bests within each backbone.}
\label{tab:scale}
\centering
\footnotesize
\setlength{\tabcolsep}{3.5pt}
\setlength{\arrayrulewidth}{0.4pt}
\arrayrulecolor{black}
\renewcommand{\arraystretch}{1.15}
\resizebox{0.7\linewidth}{!}{%
\begin{tabular}{|l|l|ccc|c|cc|}
\hline
 & & \multicolumn{3}{c|}{Fidelity} & Retrieval & \multicolumn{2}{c|}{Zero-shot} \\
 & Judge & $\rho_{\mathrm{S}}$ & $\rho_{\mathrm{P}}$ & Base $\rho_{\mathrm{S}}$ & mAP & 12 sets & IN-1k \\
\hline
\multirow{5}{*}{\rotatebox[origin=c]{90}{ViT-B/32}}
 & Untrained CLIP & \textcolor{black!40}{---} & \textcolor{black!40}{---} & \textcolor{black!40}{---} & 18.7 & \textbf{60.6} & \textbf{63.5} \\
 & Qwen3.5-397B & \cellcolor{blue!21}0.761 & 0.769 & 0.230 & \cellcolor{blue!19}48.0 & \cellcolor{red!20}58.1 & \cellcolor{red!22}61.9 \\
 & GPT-5.6 Sol & \cellcolor{blue!19}0.745 & 0.733 & 0.274 & \cellcolor{blue!18}47.0 & \cellcolor{red!25}57.2 & \cellcolor{red!25}61.6 \\
 & Kimi-K3 & \cellcolor{blue!24}\textbf{0.815} & \textbf{0.813} & 0.182 & \cellcolor{blue!19}\textbf{49.1} & \cellcolor{red!22}57.7 & \cellcolor{red!24}61.7 \\
 & Claude Sonnet 5 & \cellcolor{blue!19}0.719 & 0.752 & 0.251 & \cellcolor{blue!19}48.0 & \cellcolor{red!20}58.0 & \cellcolor{red!24}61.7 \\
\hline
\multirow{5}{*}{\rotatebox[origin=c]{90}{ViT-B/16}}
 & Untrained CLIP & \textcolor{black!40}{---} & \textcolor{black!40}{---} & \textcolor{black!40}{---} & 22.1 & 61.8 & \textbf{68.3} \\
 & Qwen3.5-397B & \cellcolor{blue!21}0.791 & 0.838 & 0.251 & \cellcolor{blue!23}\textbf{59.7} & \cellcolor{blue!6}\textbf{62.0} & \cellcolor{red!13}67.5 \\
 & GPT-5.6 Sol & \cellcolor{blue!20}0.794 & 0.804 & 0.287 & \cellcolor{blue!22}58.9 & \cellcolor{blue!6}61.9 & \cellcolor{red!18}67.0 \\
 & Kimi-K3 & \cellcolor{blue!24}\textbf{0.838} & \textbf{0.878} & 0.209 & \cellcolor{blue!22}59.0 & \cellcolor{blue!5}61.9 & \cellcolor{red!17}67.2 \\
 & Claude Sonnet 5 & \cellcolor{blue!20}0.778 & 0.828 & 0.269 & \cellcolor{blue!22}58.2 & \cellcolor{blue!6}61.9 & \cellcolor{red!17}67.2 \\
\hline
\multirow{5}{*}{\rotatebox[origin=c]{90}{ViT-L/14}}
 & Untrained CLIP & \textcolor{black!40}{---} & \textcolor{black!40}{---} & \textcolor{black!40}{---} & 30.5 & 66.1 & 75.0 \\
 & Qwen3.5-397B & \cellcolor{blue!22}0.815 & 0.873 & 0.240 & \cellcolor{blue!24}71.1 & \cellcolor{blue!11}\textbf{67.2} & \textbf{75.0} \\
 & GPT-5.6 Sol & \cellcolor{blue!21}0.805 & 0.822 & 0.274 & \cellcolor{blue!24}\textbf{71.4} & \cellcolor{blue!10}67.0 & 75.0 \\
 & Kimi-K3 & \cellcolor{blue!25}\textbf{0.869} & \textbf{0.903} & 0.191 & \cellcolor{blue!24}70.7 & \cellcolor{blue!11}67.2 & \cellcolor{red!9}74.6 \\
 & Claude Sonnet 5 & \cellcolor{blue!21}0.800 & 0.856 & 0.254 & \cellcolor{blue!23}69.5 & \cellcolor{blue!9}66.8 & 75.0 \\
\hline
\multirow{5}{*}{\rotatebox[origin=c]{90}{ViT-L/14-336}}
 & Untrained CLIP & \textcolor{black!40}{---} & \textcolor{black!40}{---} & \textcolor{black!40}{---} & 30.0 & 66.9 & \textbf{75.8} \\
 & Qwen3.5-397B & \cellcolor{blue!22}0.827 & 0.877 & 0.243 & \cellcolor{blue!25}71.9 & \cellcolor{blue!10}\textbf{67.7} & 75.8 \\
 & GPT-5.6 Sol & \cellcolor{blue!21}0.809 & 0.825 & 0.276 & \cellcolor{blue!25}\textbf{72.5} & \cellcolor{blue!7}67.3 & \cellcolor{red!6}75.7 \\
 & Kimi-K3 & \cellcolor{blue!25}\textbf{0.869} & \textbf{0.902} & 0.190 & \cellcolor{blue!25}71.7 & \cellcolor{blue!7}67.3 & \cellcolor{red!7}75.7 \\
 & Claude Sonnet 5 & \cellcolor{blue!21}0.800 & 0.854 & 0.246 & \cellcolor{blue!24}70.6 & \cellcolor{blue!6}67.1 & \cellcolor{red!7}75.7 \\
\hline
\end{tabular}%
}
\end{table}

\textbf{Matched scoring budget.}
Table~\ref{tab:scale} compares four judges across four CLIP backbones using 31 groups (15{,}376 training pairs) per judge. Fidelity is evaluated on each judge's scores for the same 100 held-out images. Kernel parameters are fitted separately for each teacher--backbone combination. The base Spearman column reports each untrained backbone's agreement with the corresponding judge; this baseline differs across judges even for a fixed backbone. Retrieval and zero-shot evaluation use the datasets and scoring protocols above. The ViT-B/16 rows share their zero-shot results with Table~\ref{tab:judges}. Pearson values are left unshaded because the table does not report their untrained baselines.

\textbf{Fidelity and retrieval.}
Every judge improves Spearman fidelity and unseen-breed retrieval over the corresponding CLIP baseline at every scale. Kimi yields the highest own-judge fidelity for all four backbones, but does not consistently give the best retrieval. The highest mAP is obtained with Kimi on ViT-B/32, Qwen on ViT-B/16, and GPT on both ViT-L/14 variants. Own-judge fidelity measures agreement with different targets and should therefore not be treated as a common measure of teacher quality.

\textbf{Retention across scales.}
Mean zero-shot accuracy exceeds CLIP for every judge on both large backbones: the gains are 0.7--1.1 points on ViT-L/14 and 0.2--0.8 on ViT-L/14-336. ImageNet accuracy differs from CLIP by at most 0.42 and 0.17 points, respectively. ViT-B/16 stays close to CLIP, whereas ViT-B/32 loses 2.6--3.4 points in mean accuracy. The 93-group results in Table~\ref{tab:backbones} use different checkpoints; these comparisons do not isolate the effect of scoring budget.

\begin{table}[t]
\caption{\textbf{Per-benchmark zero-shot accuracy across judges and backbones.} Top-1 accuracy (\%) for Table~\ref{tab:scale}; Mean averages the 12 benchmarks. Blue/red indicates improvement/decline from each backbone\textquotesingle s untrained CLIP, scaled per column across backbones. Bold marks the highest adapted-model score within each backbone when it exceeds CLIP.}
\label{tab:scale-zs}
\centering
\scriptsize
\setlength{\tabcolsep}{1.6pt}
\setlength{\arrayrulewidth}{0.4pt}
\arrayrulecolor{black}
\renewcommand{\arraystretch}{1.15}
\resizebox{0.9\linewidth}{!}{%
\begin{tabular}{|l|l|cccccccccccc|c|}
\hline
 & Judge & IN-1k & C-10 & C-100 & Caltech & FER & Pets & DTD & RESISC & EuroSAT & PCam & IN-Sk. & IN-O & Mean \\
\hline
\multirow{5}{*}{\rotatebox[origin=c]{90}{ViT-B/32}}
 & Untrained CLIP & 63.5 & 89.7 & 63.3 & 81.6 & 41.4 & 87.3 & 44.3 & 53.6 & 50.4 & 62.3 & 42.3 & 47.8 & 60.6 \\
 & Qwen3.5-397B & \cellcolor{red!22}61.9 & \cellcolor{red!21}86.5 & \cellcolor{red!16}60.8 & \cellcolor{red!22}80.1 & \cellcolor{red!16}36.8 & \cellcolor{red!16}86.2 & \cellcolor{red!25}41.1 & \cellcolor{red!12}52.4 & \cellcolor{red!22}42.1 & \cellcolor{red!10}61.2 & \cellcolor{red!20}41.1 & \cellcolor{red!25}46.5 & \cellcolor{red!20}58.1 \\
 & GPT-5.6 Sol & \cellcolor{red!25}61.6 & \cellcolor{red!25}85.6 & \cellcolor{red!25}58.6 & \cellcolor{red!25}79.9 & \cellcolor{red!25}32.6 & \cellcolor{red!21}85.6 & \cellcolor{red!21}41.7 & \cellcolor{red!11}52.6 & \cellcolor{red!25}40.9 & \cellcolor{red!17}59.4 & \cellcolor{red!20}41.0 & \cellcolor{red!22}46.7 & \cellcolor{red!25}57.2 \\
 & Kimi-K3 & \cellcolor{red!24}61.7 & \cellcolor{red!23}86.1 & \cellcolor{red!20}59.8 & \cellcolor{red!25}79.9 & \cellcolor{red!16}36.6 & \cellcolor{red!25}85.2 & \cellcolor{red!25}41.1 & \cellcolor{red!17}51.6 & \cellcolor{red!20}43.3 & \cellcolor{red!15}59.9 & \cellcolor{red!25}40.6 & \cellcolor{red!23}46.6 & \cellcolor{red!22}57.7 \\
 & Claude Sonnet 5 & \cellcolor{red!24}61.7 & \cellcolor{red!22}86.2 & \cellcolor{red!16}60.7 & \cellcolor{red!22}80.1 & \cellcolor{red!12}38.2 & \cellcolor{red!20}85.7 & \cellcolor{red!20}41.8 & \cellcolor{red!10}52.7 & \cellcolor{red!24}41.6 & \cellcolor{red!14}60.1 & \cellcolor{red!20}41.0 & \cellcolor{red!20}46.8 & \cellcolor{red!20}58.0 \\
\hline
\multirow{5}{*}{\rotatebox[origin=c]{90}{ViT-B/16}}
 & Untrained CLIP & 68.3 & 90.0 & 65.6 & 82.2 & 46.3 & 89.0 & 44.9 & 58.2 & 55.9 & 50.7 & 48.3 & 42.3 & 61.8 \\
 & Qwen3.5-397B & \cellcolor{red!13}67.5 & \cellcolor{red!12}88.5 & \cellcolor{blue!7}\textbf{66.1} & 82.2 & \cellcolor{red!10}44.1 & \cellcolor{blue!20}\textbf{90.6} & \cellcolor{red!10}44.1 & \cellcolor{blue!22}61.1 & \cellcolor{red!5}55.8 & \cellcolor{blue!17}53.4 & \cellcolor{red!15}47.5 & \cellcolor{blue!18}\textbf{43.1} & \cellcolor{blue!6}\textbf{62.0} \\
 & GPT-5.6 Sol & \cellcolor{red!18}67.0 & \cellcolor{red!15}87.9 & \cellcolor{red!10}64.5 & \cellcolor{red!11}81.6 & \cellcolor{red!7}45.6 & \cellcolor{blue!11}89.7 & \cellcolor{red!11}44.0 & \cellcolor{blue!18}60.3 & \cellcolor{blue!12}\textbf{59.1} & \cellcolor{blue!19}\textbf{54.0} & \cellcolor{red!16}47.4 & \cellcolor{red!8}42.1 & \cellcolor{blue!6}61.9 \\
 & Kimi-K3 & \cellcolor{red!17}67.2 & \cellcolor{red!14}88.2 & \cellcolor{red!7}65.2 & \cellcolor{red!10}81.7 & \cellcolor{red!7}45.4 & \cellcolor{blue!14}90.0 & \cellcolor{red!14}43.5 & \cellcolor{blue!20}60.8 & \cellcolor{blue!10}58.3 & \cellcolor{blue!10}51.8 & \cellcolor{red!15}47.4 & \cellcolor{blue!10}42.6 & \cellcolor{blue!5}61.9 \\
 & Claude Sonnet 5 & \cellcolor{red!17}67.2 & \cellcolor{red!12}88.5 & \cellcolor{red!6}65.4 & \cellcolor{red!7}82.0 & \cellcolor{red!9}44.5 & \cellcolor{blue!16}90.2 & \cellcolor{red!12}43.8 & \cellcolor{blue!23}\textbf{61.3} & \cellcolor{blue!7}56.9 & \cellcolor{blue!15}53.1 & \cellcolor{red!14}47.5 & \cellcolor{blue!8}42.5 & \cellcolor{blue!6}61.9 \\
\hline
\multirow{5}{*}{\rotatebox[origin=c]{90}{ViT-L/14}}
 & Untrained CLIP & 75.0 & 95.2 & 71.1 & 83.3 & 50.0 & 93.2 & 55.2 & 63.3 & 62.6 & 52.0 & 59.6 & 32.2 & 66.1 \\
 & Qwen3.5-397B & 75.0 & \cellcolor{red!6}95.0 & \cellcolor{blue!19}\textbf{74.4} & \cellcolor{blue!22}84.8 & \cellcolor{red!9}48.3 & \cellcolor{blue!7}93.5 & \cellcolor{red!6}55.1 & \cellcolor{blue!17}65.4 & \cellcolor{blue!17}\textbf{68.1} & \cellcolor{blue!10}53.1 & \cellcolor{blue!7}59.8 & \cellcolor{blue!23}33.5 & \cellcolor{blue!11}\textbf{67.2} \\
 & GPT-5.6 Sol & 75.0 & \cellcolor{red!8}94.6 & \cellcolor{blue!7}71.4 & \cellcolor{blue!22}84.7 & \cellcolor{red!11}47.2 & 93.2 & 55.2 & \cellcolor{blue!15}65.1 & \cellcolor{blue!15}67.4 & \cellcolor{blue!25}\textbf{56.7} & \cellcolor{blue!7}\textbf{59.8} & \cellcolor{blue!23}33.5 & \cellcolor{blue!10}67.0 \\
 & Kimi-K3 & \cellcolor{red!9}74.6 & \cellcolor{red!6}95.0 & \cellcolor{blue!15}73.5 & \cellcolor{blue!24}\textbf{84.9} & \cellcolor{red!12}47.1 & \cellcolor{blue!8}93.5 & 55.2 & \cellcolor{blue!18}\textbf{65.6} & \cellcolor{blue!15}67.5 & \cellcolor{blue!21}55.8 & 59.6 & \cellcolor{blue!25}\textbf{33.6} & \cellcolor{blue!11}67.2 \\
 & Claude Sonnet 5 & 75.0 & \cellcolor{red!8}94.5 & \cellcolor{blue!13}73.0 & \cellcolor{blue!24}84.9 & \cellcolor{red!7}49.0 & \cellcolor{blue!9}\textbf{93.6} & \cellcolor{red!9}54.6 & \cellcolor{blue!15}65.0 & \cellcolor{blue!12}66.0 & \cellcolor{blue!7}52.5 & \cellcolor{blue!6}59.7 & \cellcolor{blue!20}33.2 & \cellcolor{blue!9}66.8 \\
\hline
\multirow{5}{*}{\rotatebox[origin=c]{90}{ViT-L/14-336}}
 & Untrained CLIP & 75.8 & 94.5 & 71.1 & 83.4 & 49.0 & 93.7 & 55.6 & 63.7 & 61.5 & 60.7 & 61.0 & 32.8 & 66.9 \\
 & Qwen3.5-397B & 75.8 & \cellcolor{red!7}94.2 & \cellcolor{blue!16}\textbf{73.7} & \cellcolor{blue!20}84.7 & \cellcolor{red!8}47.9 & \cellcolor{red!11}93.1 & \cellcolor{red!9}55.1 & \cellcolor{blue!25}\textbf{67.1} & \cellcolor{blue!15}\textbf{66.3} & \cellcolor{blue!6}\textbf{61.0} & \cellcolor{blue!5}\textbf{61.1} & \cellcolor{red!9}32.5 & \cellcolor{blue!10}\textbf{67.7} \\
 & GPT-5.6 Sol & \cellcolor{red!6}75.7 & \cellcolor{red!9}93.6 & \cellcolor{blue!9}72.2 & \cellcolor{blue!13}84.1 & \cellcolor{red!9}47.3 & \cellcolor{red!10}93.2 & \cellcolor{red!6}55.4 & \cellcolor{blue!24}66.9 & \cellcolor{blue!14}65.8 & \cellcolor{red!12}59.0 & 61.0 & \cellcolor{blue!15}\textbf{33.4} & \cellcolor{blue!7}67.3 \\
 & Kimi-K3 & \cellcolor{red!7}75.7 & \cellcolor{red!7}94.1 & \cellcolor{blue!15}73.4 & \cellcolor{blue!20}\textbf{84.7} & \cellcolor{red!10}47.0 & \cellcolor{red!10}93.2 & \cellcolor{red!10}54.9 & \cellcolor{blue!24}66.9 & \cellcolor{blue!14}65.7 & \cellcolor{red!19}57.4 & \cellcolor{red!7}60.9 & \cellcolor{blue!14}33.4 & \cellcolor{blue!7}67.3 \\
 & Claude Sonnet 5 & \cellcolor{red!7}75.7 & \cellcolor{red!9}93.6 & \cellcolor{blue!9}72.0 & \cellcolor{blue!16}84.3 & \cellcolor{red!11}46.4 & \cellcolor{red!12}93.0 & \cellcolor{red!7}55.3 & \cellcolor{blue!24}66.9 & \cellcolor{blue!12}65.0 & \cellcolor{red!13}58.7 & 61.0 & \cellcolor{blue!7}32.9 & \cellcolor{blue!6}67.1 \\
\hline
\end{tabular}%
}
\end{table}

\textbf{Per-benchmark evaluation.}
Table~\ref{tab:scale-zs} expands the 12-benchmark means in Table~\ref{tab:scale}. ImageNet uses 10 images per class over 1{,}000 classes; the other benchmarks use their full test splits. Means are computed before rounding the per-benchmark values.

\textbf{Benchmark-dependent retention.}
All four ViT-B/32 students lose accuracy relative to CLIP on every benchmark. On both large backbones, every judge improves CIFAR-100, Caltech-101, RESISC45, and EuroSAT, while reducing CIFAR-10 and FER2013 accuracy. Thus, the higher aggregate accuracy of the large-backbone students coexists with losses on individual benchmarks.

\subsection{Teacher-scoring cost and student training time}
\label{app:cost}

\textbf{Teacher-scoring budget.}
Table~\ref{tab:teacher-cost} estimates OpenRouter cost and time for the default dog criterion: 44,450 distinct training-pair judgments for 1,000 images. Each request contains two images and a criterion prompt. Costs use the listed provider rates; times extrapolate published P50 latency plus output duration with ten concurrent requests. These are planning estimates, not measured scoring times; provider conditions affect realized time, and retries and fees are excluded.

\begin{table}[htbp]
\caption{\textbf{OpenRouter teacher-scoring budget estimates.} For 44,450 judgments: 300--700 input tokens per pair (including both images), two output tokens without reasoning, and ten concurrent requests. Names follow the experiments; $\dagger$ marks Gemma 3 12B pricing used as a proxy for Gemma 4 12B.}
\label{tab:teacher-cost}
\centering
\footnotesize
\setlength{\tabcolsep}{3.3pt}
\renewcommand{\arraystretch}{1.15}
\resizebox{\linewidth}{!}{%
\begin{tabular}{|l|l|c|r|r|r|r|}
\hline
Judge & Provider & \shortstack{\$/M tokens\\input / output} & \shortstack{P50\\latency (s)} & \shortstack{Output\\tokens/s} & \shortstack{Est.\\cost (\$)} & \shortstack{Est.\\time (h)} \\
\hline
\href{https://openrouter.ai/qwen/qwen3.5-397b-a17b}{Qwen3.5-397B} & Alibaba Cloud Int. & 0.39 / 2.34 & 1.11 & 61 & 5.4--12.3 & 1.4 \\
\href{https://openrouter.ai/openai/gpt-5.6-sol}{GPT-5.6 Sol} & OpenAI & 2.00 / 10.00 & 3.58 & 49 & 27.6--63.1 & 4.5 \\
\href{https://openrouter.ai/moonshotai/kimi-k3}{Kimi-K3} & Parasail & 3.00 / 15.00 & 1.11 & 68 & 41.3--94.7 & 1.4 \\
\href{https://openrouter.ai/anthropic/claude-sonnet-5}{Claude Sonnet 5} & Anthropic & 2.00 / 10.00 & 2.30 & 63 & 27.6--63.1 & 2.9 \\
\href{https://openrouter.ai/qwen/qwen3.8-27b}{Qwen3.8-27B} & Parasail & 0.24 / 2.20 & 1.35 & 58 & 3.4--7.7 & 1.7 \\
\href{https://openrouter.ai/google/gemma-3-12b-it}{Gemma 4 12B}$^{\dagger}$ & DeepInfra & 0.05 / 0.15 & 0.75 & 14 & 0.7--1.6 & 1.1 \\
\hline
\end{tabular}%
}
\end{table}

\textbf{Cost per criterion and reuse.}
The 250-image fidelity set adds 31,125 judgments, bringing the total to 75,575 (about $1.70\times$ the training-only budget). Each new criterion requires teacher judgments and student adaptation; cached judgments are reusable across backbones and ablations. Sparse supervision retains most of the retrieval gain with fewer calls (Figure~\ref{fig:query-budget}, Appendix~\ref{app:streaming-results}), reducing estimated cost and time proportionally.

\textbf{Student training.}
On one RTX A6000, median times for 3{,}000 steps are 67, 78, 92, and 181 minutes for ViT-B/32, ViT-B/16, ViT-L/14, and ViT-L/14-336, respectively. These timestamp-based estimates summarize observed configurations, exclude teacher scoring, and retain durations between 20 and 400 minutes.

\subsection{Reducing teacher API calls}
\label{app:streaming-results}

\textbf{Strong performance with fewer queries.}
Sparse supervision uses 14,320 distinct judgments instead of 44,450 (67.8\% fewer), while retaining 59.41 mAP versus 60.16 for the full budget and 62.56 zero-shot accuracy versus 62.28 (Figure~\ref{fig:query-budget}). Under the API assumptions in Appendix~\ref{app:cost}, this requires about one third of the estimated teacher-scoring cost and time. Simply reducing the dense budget to 62 groups also retains 60.07 mAP with 30,174 judgments.

\begin{figure}[htbp]
\centering
\includegraphics[width=\linewidth]{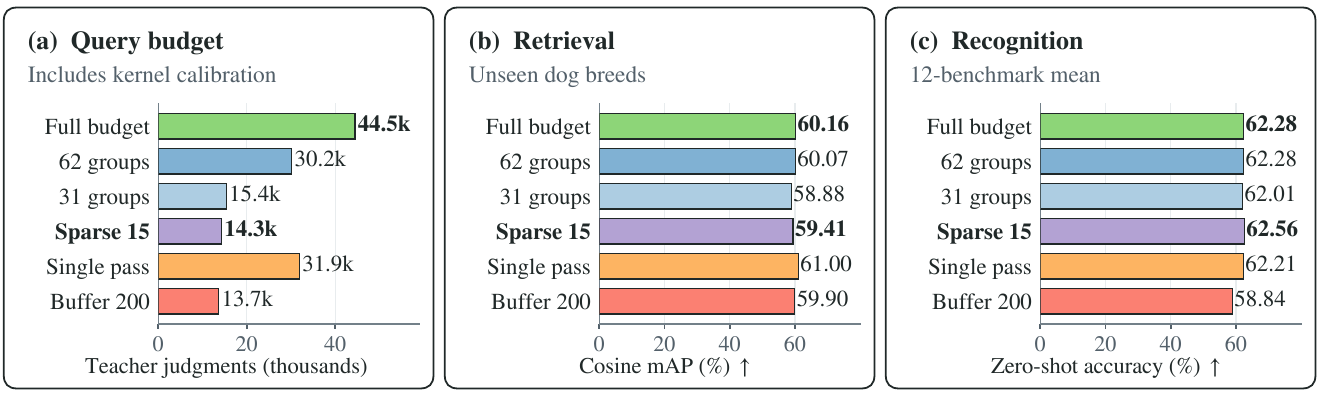}
\caption{\textbf{Strong performance with fewer teacher API calls.} \textbf{(a)} Distinct judgments, including kernel calibration. \textbf{(b)} Cosine retrieval on unseen dog breeds. \textbf{(c)} Zero-shot recognition across 12 benchmarks. Full budget is the default 93-group baseline; Sparse 15 uses 15 pairs per block.}
\label{fig:query-budget}
\end{figure}

\textbf{Baselines and variants.}
Full budget scores all 496 pairs within each of 93 groups of 32 images; the 62- and 31-group baselines use nested subsets. Sparse 15 selects 15 available pairs per random block. All cached pairs uses every available score (about 45 per block); wider blocks use 64 images and 30 pairs. Coverage selection chooses pairs to cover more distinct images. Sparse variants apply masked MSE only to observed, affine-scaled scores, without dense PSD projection or imputation.

\begin{table}[htbp]
\caption{\textbf{Query budgets and performance.} Dog-trained ViT-B/16, 3,000 updates. Update pairs count distinct training judgments; total pairs also include the 9,920 calibration pairs, without double counting. Blue/red indicates improvement/decline from the full-budget baseline, scaled per metric; bold marks the best score per column.}
\label{tab:streaming-full}
\centering
\footnotesize
\setlength{\tabcolsep}{3pt}
\setlength{\arrayrulewidth}{0.4pt}
\arrayrulecolor{black}
\renewcommand{\arraystretch}{1.15}
\resizebox{\linewidth}{!}{%
\begin{tabular}{|l|r|r|rr|c|c|c|}
\hline
Setting & Blocks & Pairs/block & \multicolumn{2}{c|}{Distinct judgments} & Fidelity & Retrieval & Zero-shot \\
\cline{4-8}
 & & & Update pairs & Total pairs & $\rho_{\mathrm{S}}$ & mAP & 12 sets \\
\hline
\textbf{Full budget (dense)} & 93 & 496 & 44,450 & 44,450 & \textbf{0.794} & 60.16 & 62.28 \\
Dense, 62 groups & 62 & 496 & 30,174 & 30,174 & \cellcolor{red!5}0.790 & \cellcolor{red!5}60.07 & 62.28 \\
Dense, 31 groups & 31 & 496 & 15,376 & 15,376 & \cellcolor{red!6}0.781 & \cellcolor{red!7}58.88 & \cellcolor{red!6}62.01 \\
\hline
\textbf{Sparse 15 (random pairs)} & 400 & 15 & 5,624 & 14,320 & \cellcolor{red!6}0.780 & \cellcolor{red!6}59.41 & \cellcolor{blue!6}\textbf{62.56} \\
All cached pairs per block & 400 & $\approx$45 & 14,703 & 21,311 & \cellcolor{red!5}0.793 & \cellcolor{red!5}60.06 & \cellcolor{blue!6}62.47 \\
Wider blocks (64 images) & 400 & 30 & 10,525 & 18,062 & \textbf{0.794} & \cellcolor{blue!5}60.18 & \cellcolor{red!8}61.48 \\
Single pass (each block once) & 3,000 & 15 & 28,332 & 31,912 & \cellcolor{red!5}0.793 & \cellcolor{blue!6}\textbf{61.00} & \cellcolor{red!5}62.21 \\
Sparse 15 (image coverage) & 400 & 15 & 5,609 & 14,285 & \cellcolor{red!6}0.776 & \cellcolor{red!7}59.05 & \cellcolor{red!18}59.07 \\
\hline
Sliding buffer (200 images) & 400 & 15 & 4,857 & 13,661 & \cellcolor{red!8}0.749 & \cellcolor{red!5}59.90 & \cellcolor{red!18}58.84 \\
Breed-sorted blocks & 400 & 15 & 3,497 & 12,627 & \cellcolor{red!20}0.597 & \cellcolor{red!14}54.82 & \cellcolor{red!25}57.18 \\
Same blocks, shuffled order & 400 & 15 & 3,497 & 12,627 & \cellcolor{red!25}0.526 & \cellcolor{red!25}48.54 & \cellcolor{red!22}57.84 \\
\hline
\end{tabular}%
}
\end{table}

\textbf{Streaming-style use.}
Single pass visits each block once, allowing images to recur, and reaches 61.00 mAP with 31,912 judgments. A sliding 200-image buffer retains 59.90 mAP but lowers zero-shot accuracy to 58.84. Breed-sorted blocks restrict batch diversity; shuffling their order preserves their composition. These cached-score replays support using sparse supervision when API budgets are limited; live acquisition remains untested.

\clearpage
\section{Ablation studies}
\label{app:ablations}

\subsection{Class labels versus VLM judgments}
\label{app:label-kernels}

\begin{table}[htbp]
\caption{\textbf{Class labels versus VLM judgments.} CLIP ViT-B/16 on Dogs, with shared kernel MSE and KL anchoring. Fidelity reports Spearman/Pearson agreement with raw held-out judgments. Retrieval mAP and the 12-benchmark zero-shot mean (ZS) are percentages. Blue/red indicates improvement/decline relative to untrained CLIP, scaled per column; bold marks the best score.}
\label{tab:label-kernels}
\centering
\footnotesize
\setlength{\tabcolsep}{4pt}
\setlength{\arrayrulewidth}{0.4pt}
\arrayrulecolor{black}
\renewcommand{\arraystretch}{1.15}
\begin{tabular}{|l|cc|cc|c|}
\hline
\multirow{2}{*}{Supervision} & \multicolumn{2}{c|}{Fidelity} & \multicolumn{2}{c|}{Retrieval mAP} & \multirow{2}{*}{ZS} \\
\cline{2-5}
 & Spearman & Pearson & Unseen breeds & Non-dog & \\
\hline
Untrained CLIP & 0.281 & 0.280 & 22.06 & 70.01 & 61.81 \\
Binary class labels & \cellcolor{blue!12}0.453 & \cellcolor{blue!20}0.707 & \cellcolor{blue!16}43.11 & \cellcolor{red!23}64.40 & \cellcolor{red!25}59.83 \\
Grouped class labels & \cellcolor{blue!12}0.451 & \cellcolor{blue!20}0.706 & \cellcolor{blue!16}43.71 & \cellcolor{red!25}63.81 & \cellcolor{red!18}60.49 \\
WordNet path similarity & \cellcolor{blue!11}0.444 & \cellcolor{blue!20}0.707 & \cellcolor{blue!21}52.01 & \cellcolor{red!7}69.41 & \cellcolor{red!16}60.70 \\
Class-pair mean (VLM) & \cellcolor{blue!23}0.730 & \cellcolor{blue!24}0.801 & \cellcolor{blue!23}57.17 & \cellcolor{blue!14}72.66 & \cellcolor{red!5}61.77 \\
\method{} (VLM judgments) & \cellcolor{blue!25}\textbf{0.794} & \cellcolor{blue!25}\textbf{0.840} & \cellcolor{blue!25}\textbf{60.16} & \cellcolor{blue!21}\textbf{74.91} & \cellcolor{blue!10}\textbf{62.28} \\
\hline
\end{tabular}
\end{table}

\textbf{Class-derived similarities.}
Table~\ref{tab:label-kernels} compares VLM judgments with class-derived targets under the same kernel objective. Unlike SupCon, this comparison retains kernel MSE as the image loss. Binary targets are $K_{ij}=\mathbf{1}[y_i=y_j]$. Grouped targets are
\[
K_{ij}=\tfrac12\mathbf{1}[y_i=y_j]+\tfrac12\mathbf{1}[h(y_i)=h(y_j)],
\]
where $h$ assigns six fixed breed groups and treats each remaining class as its own group. Similarity is therefore 1 within a class, 0.5 between distinct classes in a group, and 0 otherwise. Both targets are PSD with unit diagonal. The groups are: Beagle, Bluetick Coonhound, Black and Tan Coonhound, and Redbone Coonhound; Toy and Miniature Poodle; Staffordshire Bull Terrier and American Staffordshire Terrier; Pekingese and Shih Tzu; Border and Norwich Terrier; and Kuvasz and Komondor.

\textbf{Taxonomy-derived similarities.}
A graded label control uses WordNet 3.0 \citep{miller1995wordnet}, mapping each ImageNet breed identifier directly to its noun synset. Its target is $K_{ij}=1/(1+d(y_i,y_j))$, where $d$ is the shortest path through the hypernym hierarchy. This fixed path-similarity rule uses no VLM calls, manual breed groups, or held-out judgments. The 25 training breeds yield seven between-class distances (2--8 edges), giving scores from $1/9$ to $1/3$; same-class targets are 1. All image pairs with the same unordered breed labels share a target. The same training groups and PSD preprocessing are used; the resulting class kernel is already PSD, so projection changes the targets only at numerical precision. WordNet encodes semantic category relations, rather than genetic ancestry.

\textbf{Averaging within class pairs.}
The class-pair-mean control replaces each raw training judgment with the mean over distinct scored training image pairs having the same unordered pair of class labels. Self-pairs are excluded from these means. The same affine scaling, unit self-similarity, and per-group PSD projection are then applied. All 325 class pairs are represented among the 44,450 scored training pairs; the 93 training groups are unchanged. This retains graded relationships between classes while removing the original variation among individual image pairs before PSD processing. No held-out ratings enter the averages or the link fit.

\textbf{Shared adaptation protocol.}
All adapted models use the same 1,000 dog images and 93 groups, normalized-cubic kernel MSE, rank-32 LoRA in both towers, and the KL anchor of weight 0.2 over 1,000 ImageNet prompts. Optimization uses 3,000 AdamW updates, learning rate $1\mathrm{e}{-5}$, weight decay $1\mathrm{e}{-4}$, and 200 warm-up steps. Each target has its own cubic link fitted on frozen features from the first 20 training groups and fixed during adaptation. All models use the final checkpoint without selection on held-out data.

Fidelity compares each model's fixed link with raw VLM judgments on 250 held-out images of the adaptation classes; untrained CLIP uses the ASK link. This measures teacher imitation, although the label students do not train on those judgments. Cosine retrieval uses the same 500 unseen-breed images and 200 non-dog images as Table~\ref{tab:retrieval-full}, excluding each query from its gallery and relevant-image count. Zero-shot accuracy is the unweighted mean over the same 12 benchmarks as Table~\ref{tab:zeroshot}.

\textbf{Retrieval and recognition retention.}
Binary and grouped labels reach 43.11 and 43.71 unseen-breed mAP, compared with 60.16 for \method{} and 22.06 for CLIP. Their non-dog mAP falls to 64.40 and 63.81, below CLIP's 70.01, while \method{} reaches 74.91. Zero-shot means are 59.83 and 60.49 for the label targets, versus 62.28 for \method{}. Thus, VLM judgments improve retrieval and retention over these two label-derived targets under the shared recipe. The class-pair-mean student reaches 57.17 unseen-breed mAP, 72.66 non-dog mAP, and 61.77 ZS, compared with 60.16, 74.91, and 62.28 for \method{}. This comparison tests the effect of retaining individual image-pair judgments beyond their class-pair averages under the shared recipe; it does not establish optimal performance for all class-derived targets.

The WordNet control reaches 52.01 unseen-breed mAP, 69.41 non-dog mAP, and 60.70 ZS; its Spearman fidelity to the existing held-out VLM judgments is 0.444. This adds a query-free, graded taxonomy target to the binary and manually grouped label comparisons under the same adaptation recipe. No taxonomy metric or distance scale is selected using evaluation performance.

\subsection{Effect of PSD projection}
\label{app:psd-results}

\begin{table}[t]
\caption{\textbf{PSD projection across judges and criteria.} Neg.\% denotes negative spectral mass. Retrieval uses unseen breed labels for Dogs and unseen car-model labels for Cars. Raw-target cells show gains (blue) or losses (red) relative to PSD targets, scaled per metric; bold marks the higher score within each pair. Displayed ties are uncoloured.}
\label{tab:psd}
\centering
\footnotesize
\setlength{\tabcolsep}{3pt}
\setlength{\arrayrulewidth}{0.4pt}
\arrayrulecolor{black}
\renewcommand{\arraystretch}{1.15}
\resizebox{\linewidth}{!}{%
\begin{tabular}{|l|l|c|cc|cc|cc|}
\hline
\multirow{2}{*}{Judge} & \multirow{2}{*}{Criterion} & \multirow{2}{*}{Neg.\%} & \multicolumn{2}{c|}{Fidelity} & \multicolumn{2}{c|}{mAP} & \multicolumn{2}{c|}{ZS} \\
\cline{4-9}
 & & & PSD & raw & PSD & raw & PSD & raw \\
\hline
Gemma 4 12B & Dogs / breed & 6.4 & \textbf{0.771} & \cellcolor{red!25}0.761 & \textbf{55.93} & \cellcolor{red!25}54.49 & \textbf{61.71} & \cellcolor{red!25}61.30 \\
GPT-5.6 Sol & Dogs / breed & 3.8 & \textbf{0.790} & \cellcolor{red!16}0.785 & \textbf{58.99} & \cellcolor{red!12}58.45 & \textbf{62.10} & \cellcolor{red!19}61.81 \\
Qwen3.5-397B & Cars / pose-view & 3.5 & 0.758 & \cellcolor{blue!8}\textbf{0.760} & \textbf{45.75} & \cellcolor{red!25}44.30 & \textbf{60.88} & \cellcolor{red!18}60.63 \\
Claude Sonnet 5 & Dogs / breed & 2.6 & \textbf{0.777} & \cellcolor{red!7}0.776 & \textbf{58.57} & \cellcolor{red!7}58.40 & \textbf{62.07} & \cellcolor{red!16}61.84 \\
Qwen3.5-397B & Dogs / breed & 1.8 & \textbf{0.792} & \cellcolor{red!7}0.791 & \textbf{59.79} & \cellcolor{red!8}59.57 & 62.26 & \cellcolor{blue!6}\textbf{62.29} \\
Kimi-K3 & Dogs / breed & 0.7 & \textbf{0.838} & \textbf{0.838} & 59.35 & \cellcolor{blue!6}\textbf{59.40} & \textbf{61.97} & \cellcolor{red!11}61.85 \\
\hline
\end{tabular}%
}
\end{table}

\textbf{Matched comparison.}
Table~\ref{tab:psd} gives the absolute results underlying Table~\ref{tab:ablation}(a), together with the lower-negative-mass Kimi-K3 and Qwen dog-breed comparisons. Each raw/PSD pair uses the same 31 training groups, student, anchor, and training schedule. Dog-breed fidelity uses the shared 100-image holdout (4{,}950 pairs); car-pose fidelity uses 250 held-out car images (31{,}125 pairs). Retrieval uses 500 images from 25 unseen classes in the corresponding domain, with breed labels for Dogs and car-model labels for Cars. Thus, car retrieval measures preservation of model identity after pose alignment. Negative spectral mass is the mean, over training groups, of the fraction of absolute eigenvalue mass contributed by negative eigenvalues. The main table's deltas are calculated before rounding, so they can differ slightly from differences between the means displayed here.

\textbf{Gains from correcting indefinite targets.}
Gemma 4 12B has the largest negative spectral mass (6.4\%) and benefits on all three metrics: fidelity increases from 0.761 to 0.771, unseen-breed mAP from 54.49 to 55.93, and ZS from 61.30 to 61.71. GPT-5.6 Sol (3.8\%) also improves on all three metrics, including 0.005 fidelity and 0.53 mAP. For Qwen car-pose targets (3.5\%), PSD raises car-model mAP from 44.30 to 45.75 and ZS from 60.63 to 60.88, while fidelity decreases from 0.760 to 0.758. Sonnet (2.6\%) gains 0.17 mAP and 0.23 ZS. The lower-negative-mass Qwen dog-breed (1.8\%) and Kimi-K3 (0.7\%) cases show smaller retrieval changes; Kimi's fidelity is unchanged at the displayed precision. The comparisons show that correcting negative spectral components can improve downstream performance, with the size and metric of the benefit depending on the teacher and criterion.

\subsection{Verdict extraction}
\label{app:verdict-extraction}

\begin{figure}[t]
\centering
\includegraphics[width=\linewidth]{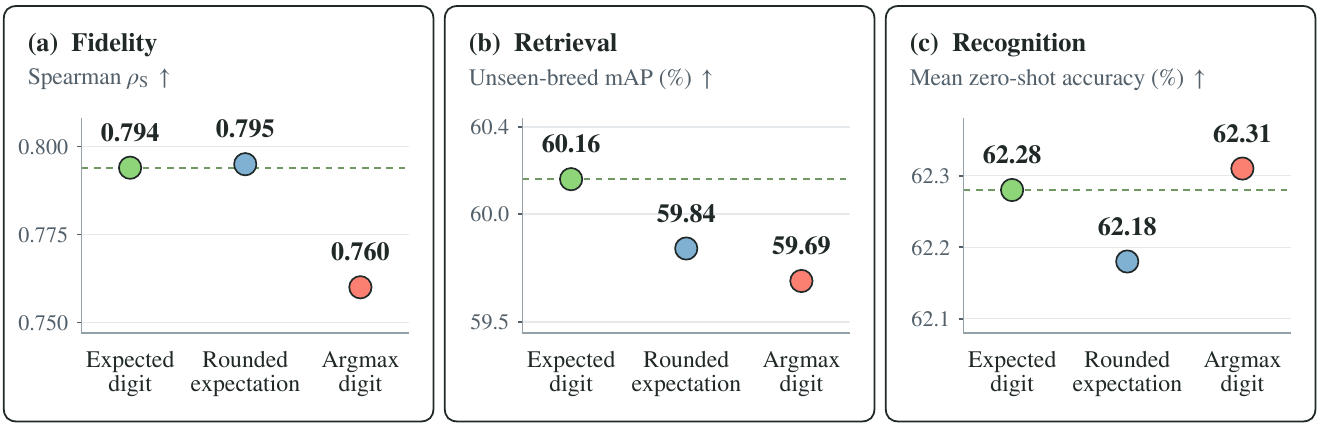}
\caption{\textbf{Verdict extraction with Qwen3.5-397B.} Expected-digit scores (green), rounded expectations (blue), and argmax digits (coral). \textbf{(a)} Fidelity uses the normalized kernel against expected-digit judgments. \textbf{(b)} Cosine-retrieval mAP uses unseen-breed labels. \textbf{(c)} Zero-shot accuracy averages 12 benchmarks. Labels give exact reported values; dashed lines mark expected-digit results. Each panel uses its own zoomed vertical scale; higher is better.}
\label{fig:verdict-extraction}
\end{figure}

\textbf{Comparison protocol.}
Figure~\ref{fig:verdict-extraction} compares expected-digit scores, rounded expectations, and argmax digits derived from the same cached Qwen responses. All variants use the same 93 training groups, PSD projection, kernel-fitting procedure, and optimization settings, including a KL weight of 0.2 and 1{,}000 anchor prompts per step. Fidelity uses each student's normalized kernel against the same expected-digit judgments on 250 held-out images. Retrieval uses cosine similarity on 500 images from 25 unseen breeds, excluding each query from its gallery; breed labels define relevance independently of the judge. ZS averages the 12 recognition benchmarks.

\textbf{Fidelity, retrieval, and retention.}
Expected-digit scoring reaches 0.794 fidelity and 60.16 mAP, compared with 0.760 and 59.69 for argmax. Rounding the expectation gives similar fidelity (0.795), with 59.84 mAP. ZS is close across the three variants (62.18--62.31).

\textbf{Distributional information and score precision.}
Taking the expectation combines support across rating levels; rounding that expectation can still produce a different digit from argmax. Thus, the expectation--argmax comparison and the expectation--rounded comparison test different choices. Moreover, PSD projection can produce continuous kernel entries from discrete input scores, so the ablation does not isolate the removal of ties in the final training targets.

\subsection{Regulariser families and weights}
\label{app:regulariser-results}
\label{app:fullablation}

\begin{table}[t]
\caption{\textbf{Regulariser families and weights.} Dog-trained CLIP ViT-B/16. Here, $P^0$ and $P$ denote the frozen and adapted image--text prediction distributions, respectively. Blue/red indicates improvement/decline from the default KL anchor, scaled per metric across all rows. Bold marks the best score within each block.}
\label{tab:regulariser-full}
\centering
\footnotesize
\setlength{\tabcolsep}{3.5pt}
\setlength{\arrayrulewidth}{0.4pt}
\arrayrulecolor{black}
\renewcommand{\arraystretch}{1.12}
\resizebox{0.9\linewidth}{!}{%
\begin{tabular}{|l|l|cc|c|cc|}
\hline
Regulariser & Weight & \multicolumn{2}{c|}{Fidelity} & Retrieval & \multicolumn{2}{c|}{Zero-shot} \\
 & & $\rho_{\mathrm{S}}$ & $\rho_{\mathrm{P}}$ & mAP & 12 sets & IN-1k \\
\hline
\multicolumn{2}{|l|}{\textbf{Default: $\mathrm{KL}(P^0\Vert P)$, weight $0.2$, text trainable (\method{})}} & 0.794 & 0.840 & 60.16 & 62.28 & 67.70 \\
\hline
\multicolumn{2}{|l|}{Untrained CLIP} & \cellcolor{red!25}0.281 & \cellcolor{red!25}0.280 & \cellcolor{red!25}22.06 & \cellcolor{red!7}61.81 & \cellcolor{blue!6}68.29 \\
\hline
\multirow{4}{*}{L2 on features (KUEA's anchor)}
 & 0.001 & \cellcolor{red!8}\textbf{0.709} & \cellcolor{red!9}\textbf{0.733} & \cellcolor{red!12}\textbf{46.83} & \cellcolor{red!8}61.33 & \cellcolor{red!5}67.61 \\
 & 0.01 & \cellcolor{red!21}0.391 & \cellcolor{red!21}0.398 & \cellcolor{red!23}26.72 & \cellcolor{red!6}\textbf{61.95} & \cellcolor{blue!7}\textbf{68.60} \\
 & 0.1 & \cellcolor{red!24}0.298 & \cellcolor{red!24}0.298 & \cellcolor{red!25}22.76 & \cellcolor{red!6}61.89 & \cellcolor{blue!7}68.41 \\
 & 1.0 (published) & \cellcolor{red!25}0.286 & \cellcolor{red!25}0.285 & \cellcolor{red!25}22.29 & \cellcolor{red!6}61.91 & \cellcolor{blue!7}68.35 \\
\hline
\multirow{4}{*}{L1 on features}
 & 0.001 & \cellcolor{blue!5}\textbf{0.795} & \cellcolor{blue!6}\textbf{0.855} & \cellcolor{red!5}59.65 & \cellcolor{red!23}57.31 & \cellcolor{red!22}61.01 \\
 & 0.01 & \cellcolor{red!5}0.792 & \cellcolor{blue!5}0.849 & \cellcolor{red!5}\textbf{59.87} & \cellcolor{red!18}58.73 & \cellcolor{red!16}63.18 \\
 & 0.1 & \cellcolor{red!6}0.760 & \cellcolor{red!7}0.786 & \cellcolor{red!10}51.15 & \cellcolor{red!9}61.24 & \cellcolor{red!7}66.98 \\
 & 1.0 & \cellcolor{red!22}0.360 & \cellcolor{red!22}0.361 & \cellcolor{red!24}23.82 & \cellcolor{red!7}\textbf{61.70} & \cellcolor{blue!7}\textbf{68.31} \\
\hline
\multirow{4}{*}{$\mathrm{KL}(P^0\Vert P)$, text frozen}
 & 0.01 & \cellcolor{red!6}\textbf{0.772} & \textbf{0.840} & \cellcolor{red!7}\textbf{56.26} & \cellcolor{red!13}60.17 & \cellcolor{red!15}63.85 \\
 & 0.1 & \cellcolor{red!6}0.760 & \cellcolor{red!6}0.818 & \cellcolor{red!9}52.15 & \cellcolor{red!8}61.39 & \cellcolor{red!8}66.65 \\
 & 0.2 & \cellcolor{red!7}0.752 & \cellcolor{red!6}0.799 & \cellcolor{red!10}49.71 & \cellcolor{red!8}61.35 & \cellcolor{red!6}67.16 \\
 & 1.0 & \cellcolor{red!10}0.671 & \cellcolor{red!12}0.636 & \cellcolor{red!19}34.03 & \cellcolor{red!7}\textbf{61.78} & \cellcolor{blue!6}\textbf{68.27} \\
\hline
\multirow{4}{*}{$\mathrm{KL}(P\Vert P^0)$, text trainable}
 & 0.01 & \cellcolor{blue!5}\textbf{0.801} & \cellcolor{blue!6}\textbf{0.857} & \cellcolor{blue!5}\textbf{60.78} & \cellcolor{red!13}60.02 & \cellcolor{red!13}64.45 \\
 & 0.1 & \cellcolor{blue!5}0.799 & \cellcolor{blue!5}0.849 & \cellcolor{blue!5}60.51 & \cellcolor{red!6}61.93 & \cellcolor{red!7}66.92 \\
 & 0.2 & 0.794 & 0.840 & \cellcolor{red!5}60.09 & \cellcolor{red!5}62.20 & \cellcolor{red!5}67.62 \\
 & 1.0 & \cellcolor{red!7}0.741 & \cellcolor{red!8}0.749 & \cellcolor{red!9}52.15 & \cellcolor{blue!5}\textbf{62.41} & \cellcolor{blue!8}\textbf{68.94} \\
\hline
\multirow{4}{*}{$\mathrm{KL}(P^0\Vert P)$, text trainable (\method{})}
 & 0.01 & \cellcolor{blue!5}\textbf{0.801} & \cellcolor{blue!6}\textbf{0.857} & \cellcolor{blue!5}\textbf{60.82} & \cellcolor{red!11}60.47 & \cellcolor{red!12}65.01 \\
 & 0.1 & \cellcolor{blue!5}0.798 & \cellcolor{blue!5}0.849 & \cellcolor{blue!5}60.43 & \cellcolor{red!5}62.21 & \cellcolor{red!6}67.13 \\
 & \textbf{0.2 (default)} & 0.794 & 0.840 & 60.16 & 62.28 & 67.70 \\
 & 1.0 & \cellcolor{red!7}0.739 & \cellcolor{red!8}0.747 & \cellcolor{red!9}51.88 & \cellcolor{blue!6}\textbf{62.43} & \cellcolor{blue!8}\textbf{68.86} \\
\hline
\multirow{5}{*}{Single settings}
 & none & \cellcolor{blue!5}0.796 & \cellcolor{blue!6}0.854 & \cellcolor{red!5}\textbf{59.41} & \cellcolor{red!25}56.67 & \cellcolor{red!25}59.82 \\
 & text L1, 0.5 & \cellcolor{blue!5}\textbf{0.796} & \cellcolor{blue!6}\textbf{0.855} & \cellcolor{red!5}59.30 & \cellcolor{red!25}56.70 & \cellcolor{red!25}59.85 \\
 & KL 0.2 $+$ text L1 0.5 & \cellcolor{red!5}0.783 & \cellcolor{red!5}0.828 & \cellcolor{red!6}57.67 & \cellcolor{blue!6}\textbf{62.47} & \cellcolor{blue!6}68.20 \\
 & PromptSRC & \cellcolor{red!6}0.775 & \cellcolor{red!6}0.819 & \cellcolor{red!7}56.54 & \cellcolor{red!5}62.25 & \cellcolor{blue!6}68.22 \\
 & L2 1.0 $+$ KL 0.2 & \cellcolor{red!25}0.285 & \cellcolor{red!25}0.283 & \cellcolor{red!25}22.26 & \cellcolor{red!6}61.89 & \cellcolor{blue!7}\textbf{68.36} \\
\hline
\multirow{5}{*}{DINOv2 teacher}
 & L2, 0.001 & \cellcolor{red!18}0.452 & \cellcolor{red!16}0.530 & \cellcolor{red!18}36.15 & \cellcolor{red!7}61.63 & \cellcolor{blue!6}68.01 \\
 & L2, 0.01 & \cellcolor{red!23}0.330 & \cellcolor{red!23}0.350 & \cellcolor{red!23}25.14 & \cellcolor{red!6}61.90 & \cellcolor{blue!7}\textbf{68.59} \\
 & L2, 0.1 & \cellcolor{red!25}0.290 & \cellcolor{red!25}0.292 & \cellcolor{red!25}22.63 & \cellcolor{red!6}61.91 & \cellcolor{blue!7}68.40 \\
 & L2, 1.0 (KUEA) & \cellcolor{red!25}0.286 & \cellcolor{red!25}0.285 & \cellcolor{red!25}22.27 & \cellcolor{red!6}61.91 & \cellcolor{blue!7}68.32 \\
 & KL, 0.2 (ours) & \cellcolor{red!15}\textbf{0.539} & \cellcolor{red!11}\textbf{0.678} & \cellcolor{red!9}\textbf{51.99} & \cellcolor{red!6}\textbf{62.07} & \cellcolor{blue!7}68.45 \\
\hline
\end{tabular}%
}
\end{table}

\textbf{Comparison setup.}
Table~\ref{tab:regulariser-full} expands the anchor comparison using dog-trained CLIP ViT-B/16, Qwen3.5-397B judgments, 93 groups. We sweep four weights for each main regulariser family; the grids differ because feature penalties and logit divergences have different scales. The single-setting block includes removing the anchor, anchoring text features, PromptSRC, and combined penalties. The final block replaces the judgment target with DINOv2 features while retaining the listed anchor. Fidelity is always measured against the VLM judgments.

\textbf{Feature anchoring.}
Reducing the feature-L2 weight from 1.0 to 0.001 raises fidelity from 0.286 to 0.709 and mAP from 22.29 to 46.83, but both remain below the default KL configuration (0.794 and 60.16). Feature L1 at weight 0.01 approaches the default fidelity and retrieval, while its zero-shot mean is lower (58.73 versus 62.28). Thus, the tested feature penalties offer a less favourable balance of alignment and retention. With strong L2 anchoring, VLM and DINOv2 targets both yield 0.286 fidelity; at weight 0.001, their scores separate to 0.709 and 0.452. The anchor can therefore obscure differences between supervision sources.

\textbf{KL strength and text adaptation.}
With a trainable text tower, increasing the weight of the default KL anchor from 0.01 to 1.0 raises mean zero-shot accuracy from 60.47 to 62.43, while reducing fidelity from 0.801 to 0.739 and mAP from 60.82 to 51.88. The default weight 0.2 retains most of the alignment gain with a 62.28 zero-shot mean. Adapting the text tower improves fidelity over freezing it at every tested weight of the default KL anchor. The two KL directions differ by at most 0.002 in fidelity, although their zero-shot means differ by up to 0.45 points.

\subsection{Frozen-feature heads versus encoder adaptation}
\label{app:heads-results}

\begin{table}[t]
\caption{\textbf{Frozen-feature heads versus LoRA.} All models use KL anchoring. (a) Criterion-specific Spearman fidelity. (b) Retrieval gains of breed-trained models. Blue/red indicates improvement/decline relative to LoRA, scaled per panel; bold marks row bests.}
\label{tab:heads}
\centering
\footnotesize
\setlength{\tabcolsep}{3.0pt}
\setlength{\arrayrulewidth}{0.4pt}
\arrayrulecolor{black}
\renewcommand{\arraystretch}{1.15}
\resizebox{0.8\linewidth}{!}{%
\begin{tabular}{|l|ccc|l|ccc|}
\hline
\multicolumn{4}{|c|}{(a) Fidelity to the criterion's own teacher} & \multicolumn{4}{c|}{(b) $\Delta$mAP over untrained CLIP} \\
\hline
Criterion & Linear & MLP & \textbf{LoRA} & Retrieval on & Linear & MLP & \textbf{LoRA} \\
\hline
Breed & \cellcolor{red!18}0.622 & \cellcolor{red!10}0.734 & \textbf{0.794} & Unseen breeds & \cellcolor{red!25}$+$13.9 & \cellcolor{red!18}$+$22.8 & \textbf{$+$38.1} \\
Colour/lighting & \cellcolor{red!25}0.491 & \cellcolor{red!11}0.673 & \textbf{0.747} & Non-dog & \cellcolor{red!8}$+$1.3 & \cellcolor{red!7}$+$1.9 & \textbf{$+$4.9} \\
Pose/view & \cellcolor{red!19}0.564 & \cellcolor{red!11}0.674 & \textbf{0.748} & Textures & \cellcolor{red!9}$+$1.3 & \cellcolor{red!8}$+$2.2 & \textbf{$+$6.2} \\
Coat texture & \cellcolor{red!19}0.699 & \cellcolor{red!9}0.826 & \textbf{0.874} & Flowers & \cellcolor{red!10}$+$2.6 & \cellcolor{red!10}$+$3.2 & \textbf{$+$8.9} \\
Background & \cellcolor{red!20}0.730 & \cellcolor{red!8}0.887 & \textbf{0.919} & Cars & \cellcolor{red!14}$+$3.7 & \cellcolor{red!13}$+$5.5 & \textbf{$+$14.7} \\
\cline{1-4}
Mean & \cellcolor{red!20}0.621 & \cellcolor{red!10}0.759 & \textbf{0.816} & Pets & \cellcolor{red!17}$+$14.6 & \cellcolor{red!11}$+$22.0 & \textbf{$+$29.0} \\
\hline
\end{tabular}%
}
\end{table}

\textbf{Matched training protocol.}
Table~\ref{tab:heads}(a) trains separate linear, MLP, and LoRA models for breed, colour and lighting; pose, viewpoint, and framing; coat texture; and background. All models use CLIP ViT-B/16, the same 1,000 dog images and 93 groups, the corresponding PSD-projected Qwen targets, normalized-cubic MSE, and the default KL anchor of weight $0.2$ over 1,000 class-name prompts. Each criterion uses its training-fitted link, fixed during adaptation and evaluation. Training uses 3,000 AdamW updates, learning rate $1\mathrm{e}{-5}$, weight decay $1\mathrm{e}{-4}$, 200 warm-up steps, and rank-32 text LoRA. Existing anchored LoRA checkpoints and the completed anchored breed heads are reused; the other eight heads follow the same head-training implementation. All results use the final checkpoint.

\textbf{Adaptation capacity.}
The identity-initialized $512\times512$ linear map has 262,144 visual parameters. The residual MLP $x+W_2\operatorname{GELU}(W_1x+b_1)+b_2$ has hidden width 2,304 and a zero-initialized second layer; its 2,362,112 visual parameters closely match LoRA\textquotesingle s 2,359,296. All models also train 1,572,864 text parameters. The visual backbone is frozen for both heads. Table~\ref{tab:ablation}(c,d) summarizes these same anchored models, with the four cross-domain retrieval gains averaged in panel (d).

\textbf{Evaluation.}
Panel (a) reports Spearman correlation with each criterion\textquotesingle s raw judgments on 250 held-out images (31,125 pairs); no link is fitted on test targets. Panel (b) reports cosine-retrieval mAP gains over untrained CLIP for the same breed-trained models: 500 unseen-breed images, 200 non-dog ImageNet images, 400 Textures, 500 Flowers, 500 Cars, and 240 Pets. Each query is excluded from its gallery and relevant-image count. The default 93-group LoRA checkpoint is shared with Table~\ref{tab:kernelform}.

\textbf{Fidelity and transfer.}
Mean fidelity is 0.816 for LoRA, 0.621 for the linear head, and 0.759 for the MLP. Unseen-breed mAP gains over CLIP are 38.1, 13.9, and 22.8 points, respectively. These experiments use a common learning rate and anchor weight; they do not establish per-architecture optimal performance.

\subsection{Kernel form}
\label{app:kernel-form}

\begin{table}[t]
\caption{\textbf{Training similarity and objective under matched LoRA adaptation.} All adapted rows share PSD targets and the KL anchor. Fidelity compares the listed similarity and deployed cosine similarity with the same raw held-out judgments, without test-time fitting. Untrained CLIP uses the default cubic link in the listed-similarity columns. mAP is cosine retrieval on unseen breeds; ZS is the 12-benchmark mean (both \%). Blue/red indicates improvement/decline relative to cubic MSE; bold marks column bests.}
\label{tab:kernelform}
\centering
\footnotesize
\setlength{\tabcolsep}{2.5pt}
\renewcommand{\arraystretch}{1.15}
\resizebox{\linewidth}{!}{%
\begin{tabular}{|l|l|l|cc|cc|cc|}
\hline
\multirow{3}{*}{Training similarity} & \multirow{3}{*}{Formula} & \multirow{3}{*}{Loss} & \multicolumn{4}{c|}{Fidelity} & \multirow{3}{*}{mAP} & \multirow{3}{*}{ZS} \\
\cline{4-7}
 & & & \multicolumn{2}{c|}{Listed similarity} & \multicolumn{2}{c|}{Cosine} & & \\
\cline{4-7}
 & & & $\rho_{\mathrm S}$ & $\rho_{\mathrm P}$ & $\rho_{\mathrm S}$ & $\rho_{\mathrm P}$ & & \\
\hline
Untrained CLIP & --- & -- & \cellcolor{red!25}0.281 & \cellcolor{red!25}0.280 & \cellcolor{red!25}0.281 & \cellcolor{red!25}0.256 & \cellcolor{red!25}22.06 & \cellcolor{red!11}61.81 \\
Dot product & $\gamma x_i^\top x_j+c$ & MSE & \cellcolor{red!9}0.696 & \cellcolor{red!11}0.677 & \cellcolor{red!9}0.688 & \cellcolor{red!10}0.672 & \cellcolor{red!15}40.50 & \cellcolor{red!25}60.63 \\
Cosine & $\hat{x}_i^\top\hat{x}_j$ & MSE & \cellcolor{red!6}0.756 & \cellcolor{red!6}0.811 & \cellcolor{red!6}0.756 & \cellcolor{blue!6}\textbf{0.811} & \cellcolor{red!6}58.35 & \cellcolor{blue!8}\textbf{62.57} \\
Normalized cubic (\method{}) & $\tilde{k}_{\gamma,c}(x_i,x_j)$ & MSE & 0.794 & \textbf{0.840} & 0.794 & 0.792 & \textbf{60.16} & 62.28 \\
Normalized cubic & $\tilde{k}_{\gamma,c}(x_i,x_j)$ & Ranking & \cellcolor{blue!5}\textbf{0.799} & \cellcolor{red!8}0.768 & \cellcolor{blue!5}\textbf{0.799} & \cellcolor{red!8}0.725 & \cellcolor{red!8}54.71 & \cellcolor{red!6}62.23 \\
Unnormalized cubic & $(\gamma x_i^\top x_j+c)^3$ & MSE & \cellcolor{red!7}0.749 & \cellcolor{red!6}0.805 & \cellcolor{red!7}0.750 & \cellcolor{red!6}0.762 & \cellcolor{red!8}54.19 & \cellcolor{blue!8}62.54 \\
\hline
\end{tabular}%
}
\end{table}

\textbf{Matched LoRA comparison.}
All adapted rows use CLIP ViT-B/16, the same 1,000 dog images, 93 PSD-projected Qwen training groups, rank-32 adapters in both towers, 3,000 updates, AdamW at $1\mathrm{e}{-5}$ with 200 warm-up steps, and the default KL anchor of weight $0.2$ over 1,000 class-name prompts. The training similarity or discrepancy changes between rows. Positive $\gamma$ and nonnegative $c$ are fitted on the first 20 training groups using frozen features and 600 Adam steps at learning rate $5\mathrm{e}{-2}$, then fixed during adaptation. Dot-product and unnormalized-cubic scales are fitted for their own forms; the normalized-cubic ranking row uses the same fixed link as the MSE row. The completed cosine and cubic MSE controls supply those reference rows.

\textbf{Objectives and evaluation.}
MSE averages the full group matrix, including its diagonal. Ranking compares all ordered pairs of off-diagonal image-pair scores, applies a hinge margin of $0.05$, and ignores teacher-score gaps at most $10^{-3}$. Fidelity reports Spearman and Pearson correlations between final-step checkpoint similarities and raw judgments on 31,125 held-out pairs. We report two readouts on the same embeddings: each row\textquotesingle s listed similarity, with link parameters fixed from training, and cosine similarity for every row. Untrained CLIP uses the default normalized-cubic link for the first readout. No link is fitted on test targets. Retrieval uses cosine similarity on 500 images from 25 unseen breeds, with each query excluded. Zero-shot evaluation averages the same 12 benchmarks as Table~\ref{tab:zeroshot}. The experiments use a shared learning rate and loss weight; they do not constitute per-objective hyperparameter optimization.

\textbf{Training objective and deployed similarity.}
With a positive offset and varying embedding norms, the normalized cubic kernel need not order candidates as cosine does. For cubic MSE, cosine fidelity is 0.794 Spearman and 0.792 Pearson, compared with 0.794 and 0.840 for the cubic kernel. Direct cosine MSE yields 0.756 Spearman and 0.811 Pearson under cosine, with 58.35 mAP versus 60.16 for cubic MSE. Thus cubic training improves cosine rank agreement and retrieval mAP in this comparison, while direct cosine training has higher cosine Pearson and a slightly higher zero-shot mean. The cubic-kernel Pearson score does not describe agreement under deployed cosine similarity.

\subsection{Sensitivity to training choices}
\label{app:sweeps}

\begin{table}[t]
\caption{\textbf{Sensitivity to training choices.} One factor changes from the default at a time. Blue/red indicates improvement/decline from the default, scaled per metric; bold marks the best score per column.}
\label{tab:sensitivity}
\centering
\footnotesize
\setlength{\tabcolsep}{2.6pt}
\setlength{\arrayrulewidth}{0.4pt}
\arrayrulecolor{black}
\renewcommand{\arraystretch}{1.15}
\resizebox{0.9\linewidth}{!}{%
\begin{tabular}{|l|l|cc|c|cc|}
\hline
Factor & Variant & \multicolumn{2}{c|}{Fidelity} & Retrieval & \multicolumn{2}{c|}{Zero-shot} \\
\cline{3-7}
 & & $\rho_{\mathrm{S}}$ & $\rho_{\mathrm{P}}$ & mAP & 12 sets & IN-1k \\
\hline
\multicolumn{2}{|l|}{\textbf{Default recipe (\method{})}} & 0.794 & 0.840 & 60.16 & 62.28 & 67.70 \\
\hline
\multirow{2}{*}{\shortstack[l]{Kernel loss\\{\scriptsize default: Frobenius}}}
 & Spectral norm & \cellcolor{blue!8}\textbf{0.803} & \cellcolor{blue!8}0.856 & \cellcolor{blue!6}60.75 & \cellcolor{red!10}61.60 & \cellcolor{red!10}66.47 \\
 & Top-4 eigenvalues & \cellcolor{blue!8}0.801 & \cellcolor{blue!8}0.854 & \cellcolor{blue!7}60.79 & \cellcolor{red!8}61.85 & \cellcolor{red!11}66.21 \\
\hline
\multirow{2}{*}{\shortstack[l]{Pair budget\\{\scriptsize default: 93 groups}}}
 & 31 groups (15{,}376 unique pairs) & \cellcolor{red!10}0.781 & \cellcolor{red!9}0.822 & \cellcolor{red!8}58.88 & \cellcolor{red!7}62.01 & \cellcolor{blue!5}67.71 \\
 & 62 groups (30{,}174 unique pairs) & \cellcolor{red!6}0.790 & \cellcolor{red!6}0.836 & \cellcolor{red!5}60.07 & 62.28 & \cellcolor{red!5}67.68 \\
\hline
\multirow{3}{*}{\shortstack[l]{Anchor weight $\lambda$\\{\scriptsize default: 0.2}}}
 & $\lambda = 0.01$ & \cellcolor{blue!8}0.801 & \cellcolor{blue!9}\textbf{0.857} & \cellcolor{blue!7}\textbf{60.82} & \cellcolor{red!17}60.47 & \cellcolor{red!16}65.01 \\
 & $\lambda = 0.1$ & \cellcolor{blue!6}0.798 & \cellcolor{blue!7}0.849 & \cellcolor{blue!6}60.43 & \cellcolor{red!5}62.21 & \cellcolor{red!7}67.13 \\
 & $\lambda = 1$ & \cellcolor{red!25}0.739 & \cellcolor{red!25}0.747 & \cellcolor{red!25}51.88 & \cellcolor{blue!6}\textbf{62.43} & \cellcolor{blue!10}\textbf{68.86} \\
\hline
\multirow{2}{*}{\shortstack[l]{Anchor text\\{\scriptsize default: class prompts}}}
 & 1{,}000 Conceptual Captions & \cellcolor{red!5}0.793 & \cellcolor{red!5}0.839 & \cellcolor{red!9}58.68 & \cellcolor{red!13}61.16 & \cellcolor{red!8}66.97 \\
 & 50/50 prompts and captions & \cellcolor{blue!5}0.795 & \cellcolor{red!5}0.839 & \cellcolor{red!5}60.05 & 62.28 & 67.70 \\
\hline
\multirow{3}{*}{\shortstack[l]{LoRA rank\\{\scriptsize default: 32, $\alpha{=}2r$}}}
 & Rank 8 & \cellcolor{red!9}0.784 & \cellcolor{red!8}0.824 & \cellcolor{blue!6}60.66 & 62.28 & \cellcolor{red!5}67.65 \\
 & Rank 64 & \cellcolor{blue!6}0.797 & \cellcolor{blue!7}0.848 & \cellcolor{red!9}58.44 & \cellcolor{red!11}61.44 & \cellcolor{red!7}67.26 \\
 & Rank 128 & \cellcolor{red!6}0.792 & \cellcolor{blue!6}0.846 & \cellcolor{red!20}53.87 & \cellcolor{red!25}59.37 & \cellcolor{red!25}62.61 \\
\hline
\end{tabular}%
}
\end{table}

\textbf{Protocol.}
Table~\ref{tab:sensitivity} changes one factor at a time from the default dog-trained ViT-B/16 configuration with PSD-projected Qwen3.5-397B judgments. Unless varied, settings are 93 scoring groups, rank-32 LoRA, 3{,}000 steps, and default KL anchor weight 0.2 with class-name prompts. The 31- and 62-group budgets are nested subsets containing 15{,}376 and 30{,}174 distinct pairs. Evaluation uses the full 250-image fidelity set for both budgets.

\textbf{Budget and rank.}
Reducing the budget from 93 to 31 groups lowers fidelity from 0.794 to 0.781 and mAP from 60.16 to 58.88. With 62 groups, these recover to 0.790 and 60.07. Rank 8 retains a 62.28 zero-shot mean and achieves 60.66 mAP, despite lower fidelity (0.784). Larger ranks do not consistently help: rank 128 yields 53.87 mAP and a 59.37 zero-shot mean. The default is therefore a practical balance among the tested settings, rather than the best choice for every metric.

\textbf{Loss and anchor.}
The spectral-norm loss raises fidelity to 0.803 but lowers the zero-shot mean to 61.60. Replacing class-name anchors with Conceptual Captions \citep{sharma2018conceptual} changes fidelity little (0.793) while reducing mAP to 58.68 and zero-shot accuracy to 61.16; mixing prompts and captions stays closer to the default. The anchor-weight sweep reproduces the alignment--retention trade-off discussed above.

\clearpage
\section{Additional visual results}
\label{app:visual-results}

\subsection{Kernel geometry}
\label{app:geometry}

These figures extend Figure~\ref{fig:ask-geometry-compact} with judge comparisons, supervision baselines, and separately aligned models for all five criteria. All targets are PSD-projected. Encoder kernels use saved polynomial links on raw features followed by diagonal normalization, without test-time link fitting. Images have a fixed ordering by breed and image path. Kernel heatmaps share a $[0,1]$ colour scale and signed residuals share $[-1,1]$. MSE counts each unordered off-diagonal pair once; it is measured against the projected target, unlike fidelity to raw judgments.

\begin{figure}[t]
\centering
\includegraphics[width=\linewidth,height=0.80\textheight,keepaspectratio]{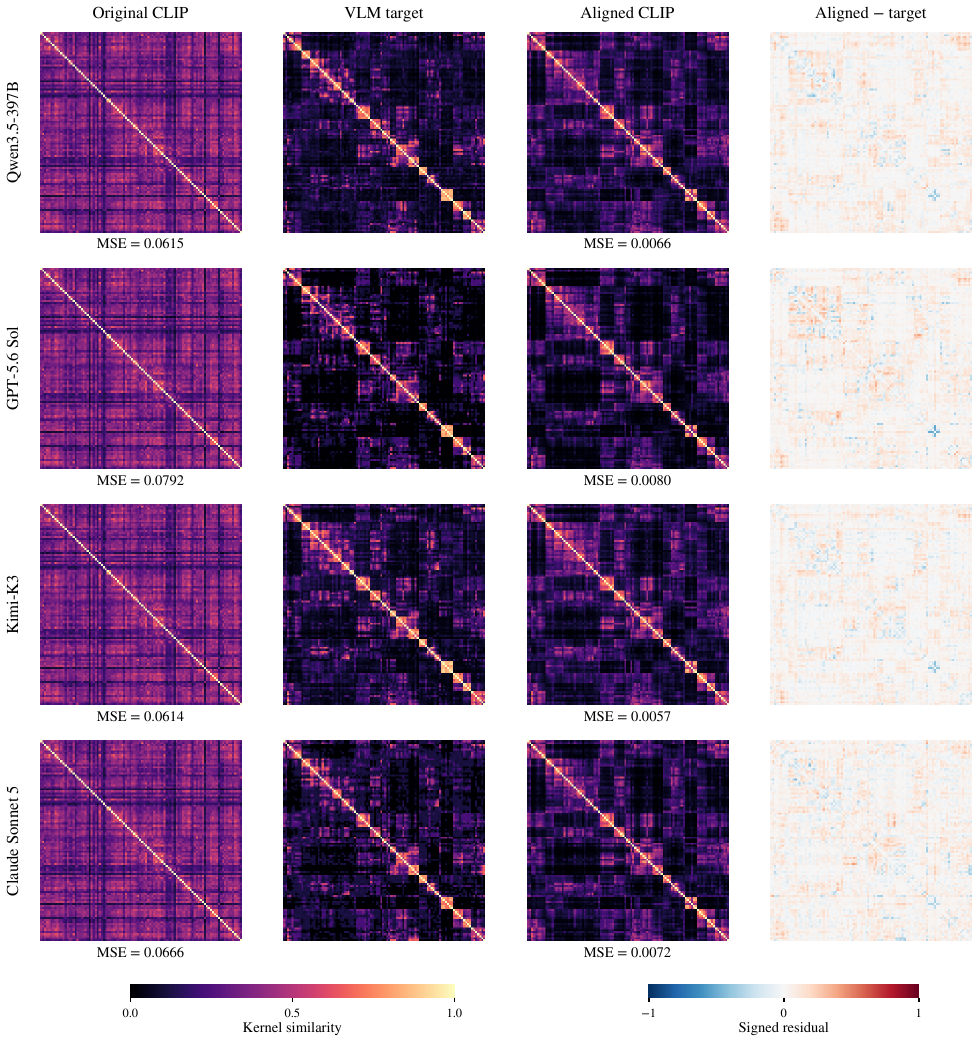}
\caption{\textbf{Kernel alignment across VLM judges.} Original CLIP, teacher target, aligned CLIP, and signed residual on 100 held-out dog images. Each row uses its judge's saved link; residuals are aligned minus target.}
\label{fig:app-geometry-judges}
\end{figure}

\textbf{Across judges.}
Figure~\ref{fig:app-geometry-judges} uses the same 100 held-out images and a matched 31-group training budget for all four judges. Alignment recovers the major structures of each target and reduces MSE relative to original CLIP. Target structure and scale vary by judge, so errors are best compared before and after alignment within each row.

\begin{figure}[t]
\centering
\includegraphics[width=\linewidth,height=0.80\textheight,keepaspectratio]{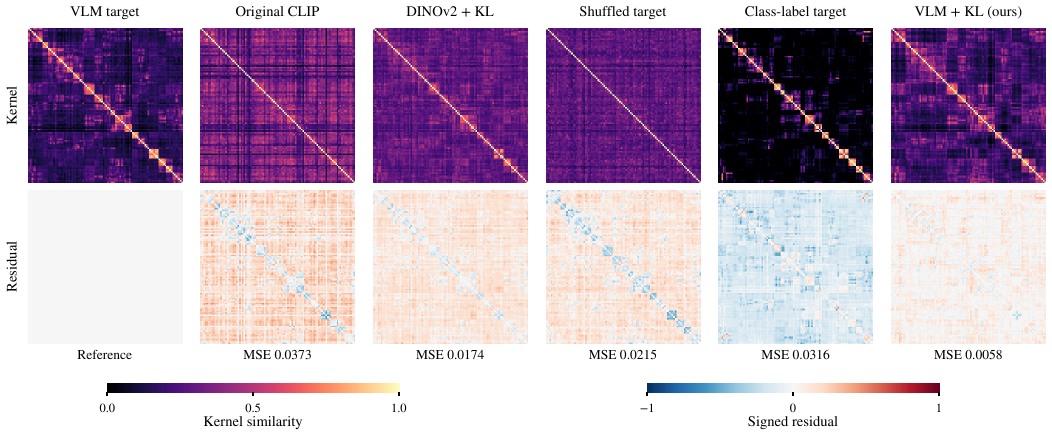}
\caption{\textbf{Kernel geometry under alternative supervision.} Top: target and encoder kernels on 100 held-out dog images. Bottom: signed residuals relative to the Qwen breed target. The VLM-aligned model uses the 93-group protocol.}
\label{fig:app-geometry-baselines}
\end{figure}

\textbf{Across supervision sources.}
Figure~\ref{fig:app-geometry-baselines} compares the stored supervision baselines on the same 100 images. The target is the corresponding submatrix of Qwen's 250-image projected breed kernel, and \method{} uses the 93-group reference model. Each model uses its own saved link; original CLIP uses the VLM-alignment link. This comparison complements the matched-judge figure and is not a 31-group budget comparison. \method{} has the smallest residual MSE among the displayed models; class-label supervision produces sharper blocks but does not reproduce the teacher's graded similarities.

\begin{figure}[t]
\centering
\includegraphics[width=\linewidth,height=0.80\textheight,keepaspectratio]{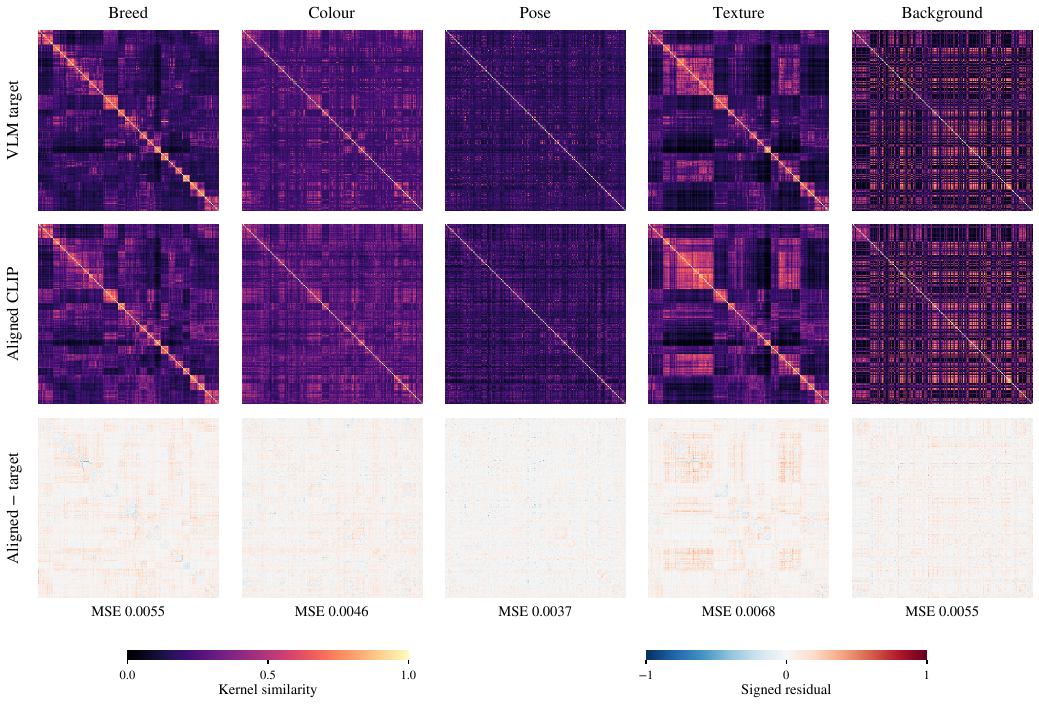}
\caption{\textbf{Target and learned geometry for five criteria.} Rows show the VLM target, criterion-aligned CLIP, and signed residual on the same 250 held-out dog images. Columns retain the same image ordering across criteria.}
\label{fig:app-geometry-criteria}
\label{fig:persona-kernels}
\end{figure}

\textbf{Across criteria.}
Figure~\ref{fig:app-geometry-criteria} compares each criterion's target with its own aligned model on 250 held-out images. It extends the breed example in Figure~\ref{fig:ask-geometry-compact}(a,b) to learned kernels and residuals for every criterion. The shared breed ordering makes breed and coat-texture blocks prominent; other criteria can express similarity through different patterns without forming the same blocks.

\clearpage

\subsection{Criterion-steered retrieval examples}
\label{app:visual-examples}

\begin{figure}[t]
\centering
\includegraphics[width=\linewidth,height=0.76\textheight,keepaspectratio]{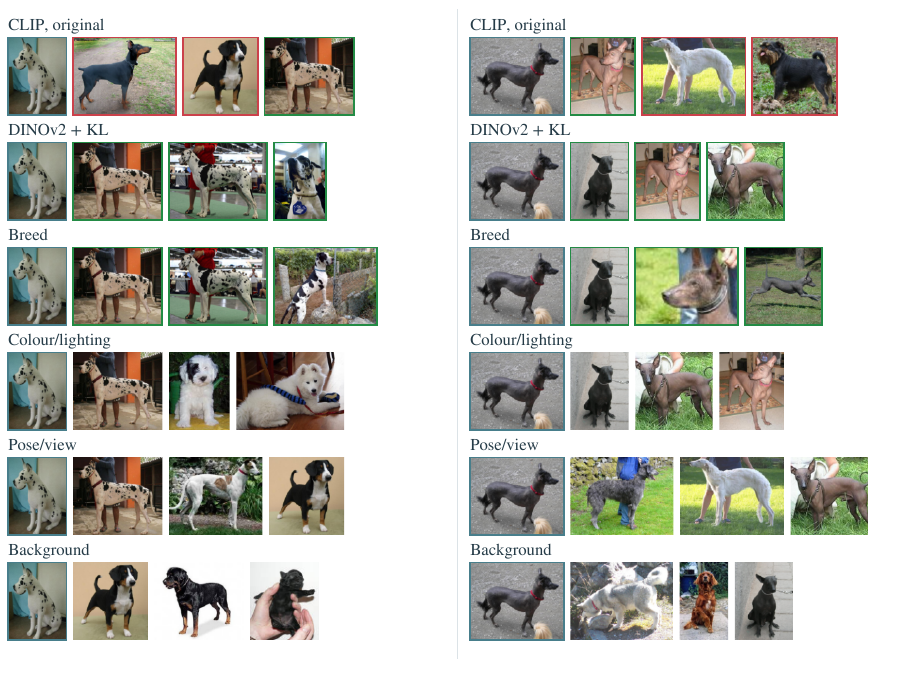}
\caption{\textbf{Dog retrieval under different criteria.} Each row shows the query (teal border) and its top three neighbours. Green/red borders indicate matching/different breeds only in the class-retrieval rows.}
\label{fig:app-criteria-dogs}
\end{figure}

\begin{figure}[t]
\centering
\includegraphics[width=\linewidth,height=0.76\textheight,keepaspectratio]{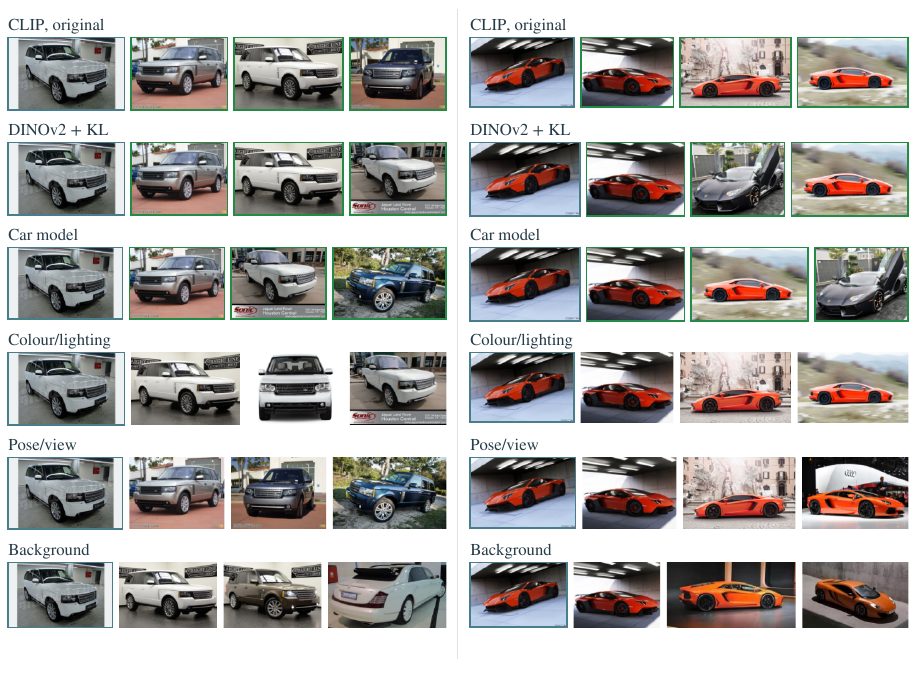}
\caption{\textbf{Car retrieval under different criteria.} Each row shows the query (teal border) and its top three neighbours. Green/red borders indicate matching/different car classes only in the class-retrieval rows.}
\label{fig:app-criteria-cars}
\end{figure}

Figures~\ref{fig:app-criteria-dogs} and~\ref{fig:app-criteria-cars} show two queries per domain, including background-based retrieval. Each row repeats the query and shows the exact top three cosine neighbours from the other 499 images. The queries were sampled from the existing example gallery, rather than the full dataset; these illustrations do not estimate typical retrieval performance. Teal borders mark queries. Green and red borders indicate matching and different class labels in the CLIP, DINOv2+KL, and breed/car-model rows. Other criteria have no correctness borders because class labels do not assess colour/lighting, pose/view, or background relevance.

The examples also show overlap between criteria. Several car neighbours recur across model, colour/lighting, and pose/view rows, while the dog colour and pose rows can retrieve different breeds. Background alignment can change the retrieved scene without preserving object identity. These comparisons illustrate the rankings produced by each encoder, including imperfect matches, rather than implying that every retrieved image satisfies its criterion.

\subsection{Top retrievals across judge-trained models}
\label{app:model-retrieval}

\begin{figure}[t]
\centering
\includegraphics[width=\linewidth,height=0.82\textheight,keepaspectratio]{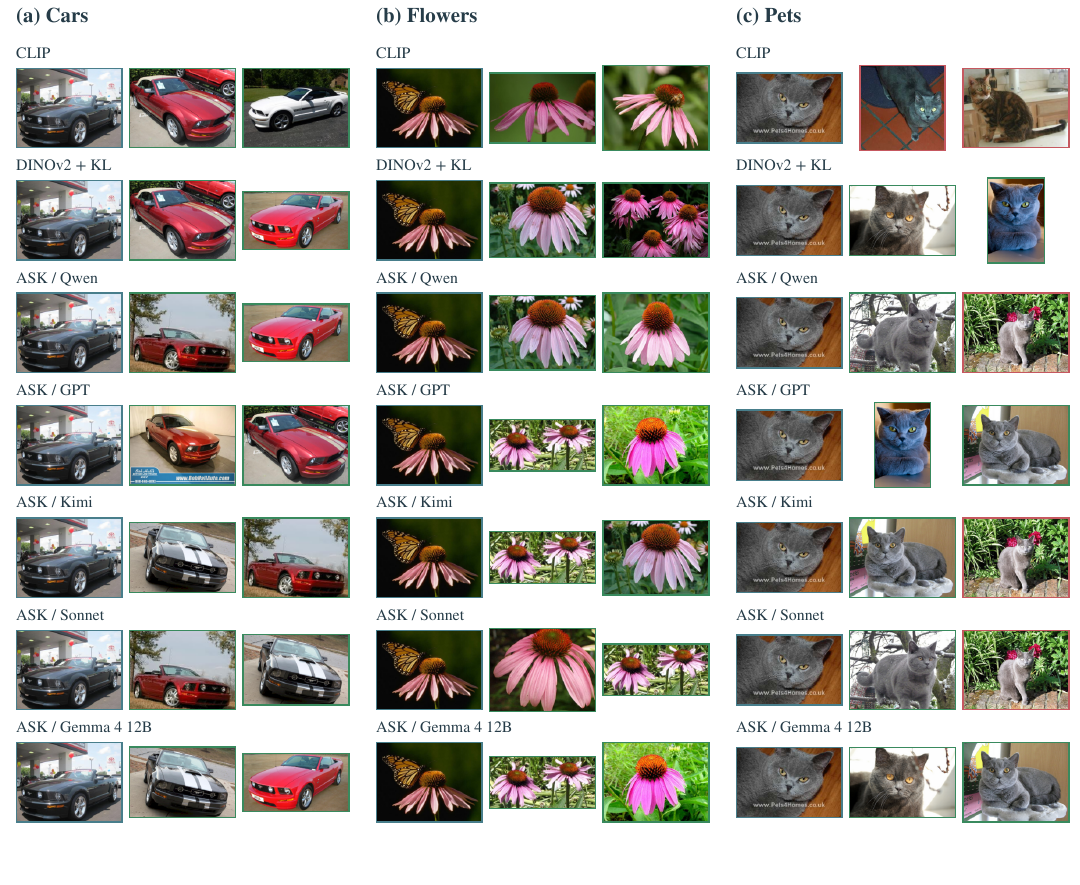}
\caption{\textbf{Different models, different nearest neighbours.} Each row shows the query (teal) and the model's top two cosine retrievals. ASK rows identify the training judge, not direct retrieval by the VLM. Green/red borders indicate matching/different dataset class labels.}
\label{fig:app-model-retrieval}
\end{figure}

Figure~\ref{fig:app-model-retrieval} compares original CLIP, dog-trained DINOv2+KL, and ASK encoders trained with Qwen3.5-397B, GPT-5.6 Sol, Kimi-K3, Claude Sonnet 5, or Gemma 4 12B. The five ASK models use the matched 31-group protocol. Each encoder ranks the same pool within Cars (500 images), Flowers (500), or Oxford-IIIT Pets (240) by cosine similarity, excluding the query. We select one readable query per domain with high disagreement among models' top-two results, without selecting for class accuracy or replacing neighbours. These selected examples illustrate different rankings rather than typical performance. All adapted encoders were trained on dogs, so the panels evaluate their behaviour on other image domains.

\clearpage

\subsection{Comparing fixed candidates across criteria}
\label{app:criterion-comparisons}

\begin{figure}[t]
\centering
\includegraphics[width=\linewidth]{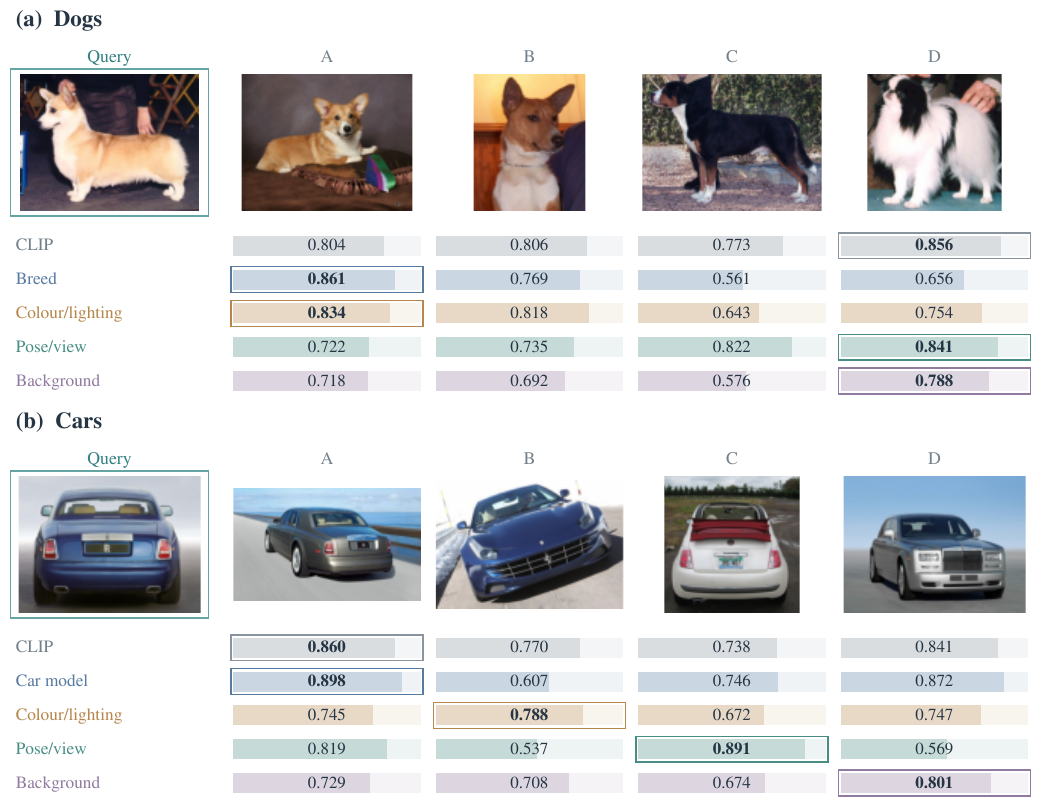}
\caption{\textbf{Same candidates, different notions of similarity.} Scores are query--candidate cosine similarities. Outlined cells mark the highest score among the four displayed candidates in each row; colours identify criteria.}
\label{fig:app-criterion-comparisons}
\end{figure}

Figure~\ref{fig:app-criterion-comparisons} holds the candidate photographs fixed while changing the similarity criterion. For cars, the highest-scoring candidate changes from A under model identity to B under colour and lighting, C under pose/view, and D under background. For dogs, breed and colour/lighting favour A, while pose and background favour D. CLIP is shown as a reference. These candidates were selected from highly ranked criterion retrievals to illustrate changes in preference, rather than to estimate retrieval accuracy. Scores are unadjusted cosine similarities; comparisons should be made within each row, since different encoders need not share a calibrated score scale.

\clearpage

\begingroup
\newcommand{\ASKpromptbox}[2]{%
  \begin{tcolorbox}[
    width=\linewidth,
    colback=black!3,
    colframe=black!50,
    colbacktitle=black!50,
    coltitle=white,
    boxrule=0.5pt,
    arc=2mm,
    outer arc=2mm,
    boxsep=0pt,
    left=10pt,right=10pt,
    top=10pt,bottom=10pt,
    toptitle=4pt,bottomtitle=4pt,
    fonttitle=\normalsize\rmfamily\bfseries,
    fontupper=\small\ttfamily,
    before upper={\raggedright\setlength{\parindent}{0pt}},
    before skip=10pt,after skip=10pt,
    title={#1}
  ]
    #2
  \end{tcolorbox}%
}

\section{Elicitation prompts}
\label{app:prompts}

Each judge receives two images and one of the following prompts. We reproduce the scoring instructions from the experiment code, with the prompt choices checked against the saved run configurations. The requested response is a single digit from 1 to 9; score extraction is described in the experimental setup. The dog primary prompt is shared across judges.

\subsection{Dog similarity criteria}

The figure labels are abbreviated descriptions of these instructions. In particular, the primary criterion includes breed, build, colour, and markings; the colour criterion covers the full image palette and lighting, rather than only the coat. Pose also includes viewpoint and framing.

\ASKpromptbox{Primary similarity (breed)}{%
Both images show dogs. Rate how similar they are on an integer scale 1-9, judging breed, build, coat colour and markings.\\
1 = clearly different breeds, nothing in common\\
3 = different breeds, loosely similar type\\
5 = related breeds or similar size/coat\\
7 = same or nearly the same breed\\
9 = same breed, near-identical individual\\
Spread your answers across the whole 1-9 range. Reply with only the single digit.
}

\ASKpromptbox{Colour and lighting}{%
Rate how similar these two images are on an integer scale 1-9, judging ONLY colour palette, lighting and overall tone. Ignore what the subject is.\\
1 = completely different palettes  5 = loosely similar tones  9 = near-identical colour and lighting\\
Spread your answers across the whole 1-9 range. Reply with only the single digit.
}

\ASKpromptbox{Pose/view}{%
Rate how similar these two images are on an integer scale 1-9, judging ONLY the subject's pose, viewing angle and framing. Ignore breed, colour and setting.\\
1 = completely different pose and viewpoint  5 = loosely similar framing  9 = near-identical pose and camera angle\\
Spread your answers across the whole 1-9 range. Reply with only the single digit.
}

\ASKpromptbox{Coat texture}{%
Rate how similar these two images are on an integer scale 1-9, judging ONLY coat texture and surface detail {-}{-} fur length, curliness, wiriness, smoothness. Ignore breed, colour, pose and setting.\\
1 = completely different coat texture  5 = loosely similar texture  9 = near-identical coat texture\\
Spread your answers across the whole 1-9 range. Reply with only the single digit.
}

\ASKpromptbox{Background}{%
Rate how similar these two images are on an integer scale 1-9, judging ONLY the background and setting {-}{-} indoors/outdoors, grass, snow, studio, street. Ignore the animal entirely.\\
1 = completely different setting  5 = loosely similar setting  9 = near-identical setting\\
Spread your answers across the whole 1-9 range. Reply with only the single digit.
}

\subsection{Primary criteria for other domains}

The five-domain experiments use the dog primary prompt above and the following four domain-specific prompts.

\ASKpromptbox{Cars}{%
Both images show cars. Rate how similar they are on an integer scale 1-9, judging body style, proportions, model and era.\\
1 = clearly different vehicles  3 = different models, similar class\\
5 = related models or same segment  7 = same or nearly the same model\\
9 = same model, near-identical vehicle\\
Spread your answers across the whole 1-9 range. Reply with only the single digit.
}

\ASKpromptbox{Flowers}{%
Both images show flowers. Rate how similar they are on an integer scale 1-9, judging petal shape, arrangement, colour and overall form.\\
1 = clearly different species, nothing in common  3 = loosely similar form\\
5 = related species or similar structure  7 = same or nearly the same species\\
9 = same species, near-identical bloom\\
Spread your answers across the whole 1-9 range. Reply with only the single digit.
}

\ASKpromptbox{Pets}{%
Both images show pets. Rate how similar they are on an integer scale 1-9, judging breed, build, coat colour and markings.\\
1 = clearly different breeds, nothing in common  3 = different breeds, loosely similar type\\
5 = related breeds or similar size and coat  7 = same or nearly the same breed\\
9 = same breed, near-identical individual\\
Spread your answers across the whole 1-9 range. Reply with only the single digit.
}

\ASKpromptbox{Textures}{%
Both images show textures or patterns. Rate how similar they are on an integer scale 1-9, judging the pattern structure, regularity, scale and surface character.\\
1 = completely different patterns  3 = loosely related structure  5 = similar family of pattern\\
7 = same kind of pattern  9 = near-identical pattern\\
Spread your answers across the whole 1-9 range. Reply with only the single digit.
}

\subsection{Car-specific criterion prompts}

The car colour prompt is identical to the colour prompt above. The pose prompt replaces ``breed'' with ``model'', and the background prompt uses vehicle-specific wording. Their complete instructions are given below.

\ASKpromptbox{Car pose/view}{%
Rate how similar these two images are on an integer scale 1-9, judging ONLY the subject's pose, viewing angle and framing. Ignore model, colour and setting.\\
1 = completely different pose and viewpoint  5 = loosely similar framing  9 = near-identical pose and camera angle\\
Spread your answers across the whole 1-9 range. Reply with only the single digit.
}

\ASKpromptbox{Car background}{%
Rate how similar these two images are on an integer scale 1-9, judging ONLY the background and setting {-}{-} indoors/outdoors, studio, street, showroom, landscape. Ignore the vehicle entirely.\\
1 = completely different setting  5 = loosely similar setting  9 = near-identical setting\\
Spread your answers across the whole 1-9 range. Reply with only the single digit.
}

\clearpage
\subsection{CUB attribute scoring prompts}
\label{app:cub-prompts}

For the CUB experiment in Appendix~\ref{app:human-attributes}, Qwen3.8-27B receives two training images and one of the following criterion-specific prompts.

\ASKpromptbox{Wing colour}{%
Rate how similar these two bird images are on an integer scale 1-9, judging ONLY the intrinsic colours of the visible wing feathers. Compare the main wing colours and their approximate proportions. Ignore bird species, bill shape, breast markings, pose, background, and differences in illumination. If a wing is partly hidden, judge only the visible evidence rather than inferring its colour from the species.\\
1 = completely different wing colours\\
5 = some shared wing colours but substantial differences\\
9 = nearly identical wing colours and proportions\\
Use the full 1-9 range. Reply with only the single digit.
}

\ASKpromptbox{Bill shape}{%
Rate how similar these two bird images are on an integer scale 1-9, judging ONLY the shape of the bill or beak. Compare its outline, curvature, taper, thickness relative to length, and tip shape. Ignore bird species, absolute image scale, bill colour, feather colours and patterns, pose, lighting, and background. Judge the visible bill rather than inferring its shape from the species.\\
1 = completely different bill shapes\\
5 = some shared shape features but substantial differences\\
9 = nearly identical bill shapes and proportions\\
Use the full 1-9 range. Reply with only the single digit.
}

\ASKpromptbox{Breast pattern}{%
Rate how similar these two bird images are on an integer scale 1-9, judging ONLY the visible breast-feather pattern: solid or unmarked, spotted, striped or streaked, or divided into multiple colour patches. Compare the kind, density, and arrangement of markings, ignoring the actual hues. Ignore bird species, wing appearance, bill shape, pose, lighting, and background. Judge the visible breast rather than inferring its pattern from the species.\\
1 = completely different breast patterns\\
5 = related patterns with noticeable differences\\
9 = nearly identical breast patterns and marking arrangement\\
Use the full 1-9 range. Reply with only the single digit.
}

\endgroup

\clearpage

\end{document}